%% file: neurips_2024.tex
\documentclass{article}

     \usepackage[preprint]{neurips_2024}

\usepackage[utf8]{inputenc} 
\usepackage[T1]{fontenc}    
\usepackage{hyperref}       
\usepackage{url}            
\usepackage{booktabs}       
\usepackage{amsfonts}       
\usepackage{nicefrac}       
\usepackage{microtype}      
\usepackage{xcolor}         

\input{math_commands.tex}

\usepackage{svg}
\usepackage{hyperref}
\usepackage{url}
\usepackage{colortbl}
\usepackage{microtype}
\usepackage{graphicx}
\usepackage{multirow}
\usepackage[inline, shortlabels]{enumitem}
\usepackage{amssymb}
\usepackage{mathtools}
\usepackage{amsthm}
\usepackage{subfigure}
\newtheorem{theorem}{Theorem}[section]

\newtheorem{lemma}[theorem]{Lemma}

\theoremstyle{definition}

\newtheorem{assumption}[theorem]{Assumption}
\theoremstyle{remark}

\usepackage{algorithm, algorithmic, bm}
\usepackage{amsthm,amsmath,amssymb}
\usepackage{pifont}
\usepackage{wrapfig,lipsum,booktabs}

\title{Covariance-Adaptive  Sequential  Black-box Optimization for  Diffusion Targeted Generation}

\author{%
 Yueming Lyu \\
 CFAR, IHPC \\
    A*STAR \\
  \texttt{yueminglyu@gmail.com} \\
   \And
  Kim Yong Tan \\
  College of Computing and Data Science \\
   Nanyang Technological University \\
   \texttt{kimyong001@e.ntu.edu.sg} \\
   \AND
   Yew Soon Ong \\
College of Computing and Data Science \\
   Nanyang Technological University \\
   \texttt{asysong@ntu.edu.sg} \\
   \And
  Ivor W. Tsang \\
  CFAR, IHPC \\
   A*STAR \\
   \texttt{ivor\_tsang@cfar.a-star.edu.sg} \\
}

\begin{document}

\maketitle

\begin{abstract}
  Diffusion models have demonstrated great potential in generating high-quality content for images, natural language, protein domains, etc. However, how to perform user-preferred targeted generation via diffusion models with only black-box target scores of users remains challenging.   To address this issue, we first formulate the fine-tuning of the targeted reserve-time stochastic differential equation (SDE) associated with a pre-trained diffusion model as a sequential black-box optimization problem. 
Furthermore, we propose a novel covariance-adaptive sequential optimization algorithm to optimize cumulative black-box scores under unknown transition dynamics.   Theoretically, we prove a $O(\frac{d^2}{\sqrt{T}})$ convergence rate for cumulative convex functions without smooth and strongly convex assumptions.  
Empirically,  experiments on both numerical test problems and target-guided 3D-molecule generation tasks show the superior performance of our method in achieving better target scores.  

\end{abstract}

\section{Introduction}

Diffusion models have shown great success in generating high-quality content in various domains, such as image generation~\citep{rombach2022high,ramesh2022hierarchical}, video generation~\citep{ho2022imagen}, speech generation~\citep{kim2022guided,kong2020diffwave},  and natural language generation~\citep{hu2023unified,diffusionbert}.   Thanks to the super-promising power of the diffusion models, 
guided sampling via diffusion models to achieve desired properties recently emerged and shown fantastic potential in many applications, e.g., text-to-image generation~\citep{kim2022diffusionclip},  image-to-image translation~\citep{tumanyan2023plug},  protein design~\citep{lee2023score,gruver2023protein}.

Despite the popularity and success of diffusion models, how to employ diffusion models to generate user-preferred content with black-box target scores while avoiding re-training from scratch is still challenging and unexplored.   One direct idea is to treat this problem as a black-box optimization problem and employ black-box optimization techniques~\citep{audet2017derivative,alarie2021two,doerr2019theory} to perform the fine-tuning of a pre-trained diffusion model with only black-box target scores.  However, naively applying black-box optimization methods to optimize diffusion model parameters faces high-dimensional optimization challenges, which are prohibitive to achieving a meaningful solution in a feasible time. 

More importantly, current black-box optimization techniques, e.g.,  Bayesian optimization techniques~\citep{gpucb,gardner2017discovering,nayebi2019framework}, Evolution strategies (ES) or stochastic zeroth-order optimization~\citep{back1991survey,hansen2006cma,NES,lyu2021black,liu2018zeroth,wang2018stochastic} and genetic algorithms~\citep{srinivas1994genetic,mirjalili2019genetic}, are designed for single objective without considering the transition dynamic nature of sequential functions. As a result, we can not directly apply them to diffusion models due to ignoring the sequential nature of the generation process of diffusion models.

In this paper,  we dig into the transition dynamic of the inference of diffusion models.  By leveraging the relationship between the inference of diffusion models and the reverse-time Stochastic Differential Equation (SDE)~\citep{song2020score}, we propose a novel targeted SDE fine-tuning framework for targeted generation.  To learn the fine-tuning parameter, we further formulate it as a sequential black-box optimization problem. 

To solve the sequential black-box optimization problem, we propose a novel covariance-adaptive sequential black-box optimization algorithm by explicitly handling the history trajectory dependency in the cumulative black-box target functions. Our method performs full covariance matrix adaptive updates that can take advantage of second-order information to deal with ill-conditioned problems.  Theoretically, we prove a $O(\frac{d^2}{\sqrt{T}})$ convergence rate for convex functions without smooth and strongly convex assumptions. Thus, our method can handle non-smooth problems. Our contributions are listed as follows:
\begin{itemize}
   
     \item We propose a novel fine-tuning framework for black-box targeted generation.  Our framework fine-tunes the targeted reverse-time SDE associated with a pre-trained diffusion model, which is general enough to guide the design of algorithms for particular downstream black-box targeted generation tasks.  Furthermore,  we formulate the learning of fine-tuning parameters as a sequential black-box optimization problem.

    \item We proposed a novel covariance-adaptive sequential black-box optimization (CASBO) algorithm. Our CASBO can perform a full covariance matrix update to exploit the second-order information.   Theoretically, we prove a $O(\frac{d^2}{\sqrt{T}})$ convergence rate for convex functions without smooth and strongly convex assumptions. Thus, our theoretical analysis can handle non-smooth problems. The convergence analysis of full covariance-adaptive black-box optimization for convex functions without the smooth and strongly convex assumptions is technically challenging. Technically,  we add a $\gamma_t$ enlargement term in the gradient update. This technique enables us to construct feasible solution sets of the adaptive update matrix during the whole algorithm running process to ensure convergence.   To the best of our knowledge, our CASBO algorithm is the first full covariance matrix adaptive black-box optimization method that achieves a provable $O(\frac{d^2}{\sqrt{T}})$ convergence rate for convex functions without smooth and strongly convex assumptions. 
    \item Empirically, we can naturally apply our optimization algorithm for diffusion black-box targeted generation.  Experiments on both numerical test problems and target-guided 3D-molecule generation tasks show the superior performance of our method in achieving better target scores.

\end{itemize}

\vspace{-8pt}
\section{Notation and Symbols}\vspace{-5pt}
Denote $\|\cdot\|_2$ and $\|\cdot\|_F$ as the spectral norm and Frobenius norm for matrices, respectively. Define $\text{tr}(\cdot)$ as the trace operation for matrix.  Notation $\|\cdot\|_2$ will also denote $l_2$-norm for vectors. Symbol $\left<\cdot, \cdot \right>$ denotes inner product under $l_2$-norm  for vectors   and inner product under Frobenius norm for matrices. For a positive semi-definite matrix $C$,  define $\|\boldsymbol{x}\|_C: = \sqrt{ \left<\boldsymbol{x},C\boldsymbol{x} \right> }$. Denote $\mathcal{S}^{+}$ and  as the set of positive semi-definite matrices.  
 Denote  $\Sigma^{\frac{1}{2}}$  as the symmetric positive semi-definite matrix such that $\Sigma = \Sigma^{\frac{1}{2}}\Sigma^{\frac{1}{2}}$ for $\Sigma \in \mathcal{S}^{+}$. Denote $\Bar{\boldsymbol{x}}_k=[\boldsymbol{x}_1^\top,\cdots,\boldsymbol{x}^\top_k]^\top \in \mathbb{R}^{kd}$ {, where $\boldsymbol{x}_i \in \mathbb{R}^d$, $d$ denotes the dimension of the data}. Denote $\boldsymbol{\Bar{\mu}}_{k} = [\boldsymbol{\mu}_1^\top,\cdots,\boldsymbol{\mu}_k^\top]^\top \in \mathbb{R}^{kd}$ and   $\boldsymbol{\Bar{\Sigma}}_{k} = diag(\boldsymbol{\Sigma}_1,\cdots,\boldsymbol{\Sigma}_k) \in \mathbb{R}^{kd \times kd}$ as the mean and  diagonal block-wise  covariance matrix for Gaussian distribution, respectively.    Denote $\Bar{\theta}_k := \{ \boldsymbol{\Bar{\mu}}_{k}, \boldsymbol{\Bar{\Sigma}}_{k}\}$ as the parameter {of the  distribution for candidate sampling in black-box} optimization and ${\theta}_k := \{ \boldsymbol{{\mu}}_{k}, \boldsymbol{{\Sigma}}_{k}\}$  as its $k^{th}$ component.  Denote $\tau$ as the time in the diffusion model.  And denote  $t$ as the iteration index in optimization.    Denote $\Bar{\theta}_k^t := \{ \boldsymbol{\Bar{\mu}}_{k}^t, \boldsymbol{\Bar{\Sigma}}_{k}^t\}$ as the parameter at $t^{th}$ iteration. 

\vspace{-8pt}
\section{Approach}
\label{Approach}
\vspace{-8pt}
\subsection{Target SDE Fine-tuning Framework}

Consider the forward SDE of the diffusion model as below
\begin{align}
    d \hat{\boldsymbol{x}}  = \hat{f}(\hat{\boldsymbol{x}},\tau) d \tau + g(\tau) d \boldsymbol{w},
\end{align}
where $\boldsymbol{w}$ is the standard Wiener process (Brownian motion), and function $\hat{f}(\cdot,\tau): \mathbb{R}^d \to \mathbb{R}^d $ maps the input $\hat{\boldsymbol{x}}$ to a $d$-dimensional vector.  
\cite{song2020score} show that the related backward SDE can be formulated as 
\begin{align}
   d \hat{\boldsymbol{x}} = [\hat{f}(\hat{\boldsymbol{x}},\tau) - g(\tau)^2 \nabla_{\hat{\boldsymbol{x}}} \log ( p_\tau ( \hat{\boldsymbol{x}} ) )    ] d\tau  + g(\tau) d \Bar{\boldsymbol{w}},
\end{align}
where $\Bar{\boldsymbol{w}}$  denotes a standard Wiener process when time flows backwards from $K$ to $0$. More details of the related background can be found in Appendix~\ref{Background}.

We propose a principle framework for fine-tuning the diffusion model SDE to optimize a target function $F(\hat{\boldsymbol{x}})$.  The proposed 
 fine-tuned SDE for target generation is given as Eq.(\ref{FineTunedSDE}).  It modifies the original reverse-time SDE by learning the  function $\mu(\cdot, \tau)$ and $\Sigma(\cdot, \tau)$  as follows:
\begin{align}
\label{FineTunedSDE}
   d \hat{\boldsymbol{x}} = [\hat{f}(\hat{\boldsymbol{x}},\tau) - g(\tau)^2 \nabla_{\hat{\boldsymbol{x}}} \log ( p_\tau ( \hat{\boldsymbol{x}} ) ) {\color{red} - \mu(\nabla F(\hat{\boldsymbol{x}}), \tau  )}    ] d\tau  + {\color{red} \Sigma( \hat{\boldsymbol{x}}, \tau)}  d \Bar{\boldsymbol{w}},
\end{align}
where function $\mu(\cdot, \tau): \mathbb{R}^d \to \mathbb{R}^d $ maps the input gradient to a target descent direction vector (for minimization), i.e.,  $\big< \mu(\nabla F(\hat{\boldsymbol{x}}), \tau  ), \nabla F(\hat{\boldsymbol{x}}) \big> \le 0 $.  And function $\Sigma(\cdot, \tau): \mathbb{R}^d \to \mathbb{R}^{d \times d}$ maps the input $\hat{\boldsymbol{x}}$ to a positive semi-definite covariance matrix. 

Given a fixed discretization schedule of time and absorbing it into index $\tau$ for notation simplicity,  we achieve the following discrete update formula for sampling: 
\begin{align}
   \hat{\boldsymbol{x}}_{\tau } =  \hat{\boldsymbol{x}}_{\tau\!+\!1} \!-\! \hat{f}(\hat{\boldsymbol{x}}_{\tau\!+\!1}) +  g(\tau \! \!+\!\!1)^2 \nabla_{\hat{\boldsymbol{x}}} \log ( p_{{\tau+1}} ( \hat{\boldsymbol{x}}_{\tau\!+\!1} ) ) + \mu_{\tau\!+\!1}(\nabla F(\hat{\boldsymbol{x}}_{\tau\!+\!1})) + \Sigma_{\tau \!+\!1}( \hat{\boldsymbol{x}}_{\tau \!+\!1} ) \boldsymbol{z}_{\tau\!+\!1},
\end{align}
where $\boldsymbol{z}_{\tau+1} \sim \mathcal{N}(0,\boldsymbol{I})$ and $\tau \in \{K-1,\cdots,0\}$. 

Let  $\tilde{\boldsymbol{x}}_k = \hat{\boldsymbol{x}}_{K-k} $, and set $\widehat{\boldsymbol{\mu}}_{\phi}( \tilde{\boldsymbol{x}}_{k-1},k-1) =  \tilde{\boldsymbol{x}}_{k-1} - \hat{f}(\tilde{\boldsymbol{x}}_{k-1}) +  g(K-k+1)^2 \nabla_{\tilde{\boldsymbol{x}}} \log ( p_{{K\!-\!k\!+\!1}} ( \tilde{\boldsymbol{x}}_{k-1} ) )  $, 
we have that   
\begin{align}
      \tilde{\boldsymbol{x}}_{k} & =  \widehat{\boldsymbol{\mu}}_{\phi}( \tilde{\boldsymbol{x}}_{k-1},k-1)  + \mu_{K\!-\!k\!+\!1}(\nabla F(\tilde{\boldsymbol{x}}_{k-1})) + \Sigma_{K\!-\!k\!+\!1}( \tilde{\boldsymbol{x}}_{k-1} ) \boldsymbol{z}_{K\!-\!k\!+\!1} 
\end{align}
for $k\in \{1,\cdots,K\}$.

The sampling rule for targeted generation can be reparameterized as follows:
\begin{align}
     \tilde{\boldsymbol{x}}_{k} =  \widehat{\boldsymbol{\mu}}_{\phi}( \tilde{\boldsymbol{x}}_{k-1},k-1)  + \mathcal{N}(\mu_{K\!-\!k\!+\!1}(\nabla F(\tilde{\boldsymbol{x}}_{k-1})),  \Sigma_{K\!-\!k\!+\!1}( \tilde{\boldsymbol{x}}_{k-1} )  )
\end{align}
for $k\in \{1,\cdots,K\}$.

In practice, we employ a pre-trained diffusion model $\widehat{\boldsymbol{\mu}}_{\phi}( \tilde{\boldsymbol{x}}_{k\!-\!1},k\!-\!1) $.  In our black-box targeted generation task,  the target function to be optimized is black-box, the query efficiency is important.   We thus employ a  simple constant function learning scheme,  i.e., $\mu_{K\!-\!k\!+\!1}(\nabla F(\tilde{\boldsymbol{x}}_{k-1})) = \tilde{\sigma}_k \boldsymbol{\mu}_{k}$ and
$\Sigma_{K\!-\!k\!+\!1}( \tilde{\boldsymbol{x}}_{k-1} ) = \tilde{\sigma}_k^2 \boldsymbol{\Sigma}_{k}$. 
The transition dynamic of the targeted diffusion inference sampling can be simplified as follows:
\begin{align}
\label{DiffusionSampling}
      \tilde{\boldsymbol{x}}_{k} =  \widehat{\boldsymbol{\mu}}_{\phi}( \tilde{\boldsymbol{x}}_{k-1},k-1)  + \tilde{\sigma}_k \mathcal{N}(\boldsymbol{\mu}_{k},  \boldsymbol{\Sigma}_{k} ),
\end{align}
where $\tilde{\sigma}_k$ denotes the SDE solver coefficient that depends on the concrete SDE slover type.
We leave more sophisticated learning of function $\mu_{k}(\cdot)$ and function $\Sigma_{k}(\cdot)$ as future work.

Let $\boldsymbol{x}_k$ be sampled from Gaussian distribution $\mathcal{N}(\boldsymbol{\mu}_{k},  \boldsymbol{\Sigma}_{k})$. From the  transition dynamic  $ \tilde{\boldsymbol{x}}_{k} =  \widehat{\boldsymbol{\mu}}_{\phi}( \tilde{\boldsymbol{x}}_{k-1},k-1)  + \tilde{\sigma}_k \boldsymbol{x}_k $,  we know that  $\tilde{\boldsymbol{x}}_k$ is a function depends on the trajectory $\Bar{\boldsymbol{x}}_k:=[\boldsymbol{x}_1,\boldsymbol{x}_2,\cdots,\boldsymbol{x}_{k}]$, i.e., $\tilde{\boldsymbol{x}}_k=\tilde{\boldsymbol{x}}_k(\Bar{\boldsymbol{x}}_k)$.  To perform guided sampling from the pre-trained diffusion model with the black-box target score function $F(\tilde{\boldsymbol{x}})$, we optimize the   cumulative target score:
\begin{align}
\label{AugObj}
    J(\Bar{\theta}_K):= \mathbb{E}_{{\boldsymbol{x}}_{1},\cdots,\boldsymbol{x}_K} \big[ \sum \nolimits_{k=1}^K F( \tilde{\boldsymbol{x}}_k) \big]  = \mathbb{E}_{\Bar{\boldsymbol{x}}_{K} \sim \mathcal{N}(\Bar{\boldsymbol{\mu}}_K,\Bar{\boldsymbol{\Sigma}}_K)} \big[ \sum \nolimits_{k=1}^K f_k( \Bar{\boldsymbol{x}}_k) \big],
\end{align}
where $\Bar{\theta}_K:= \{ \boldsymbol{\Bar{\mu}}_{K}, \boldsymbol{\Bar{\Sigma}}_{K}\} $ and $f_k( \Bar{\boldsymbol{x}}_k)= F(\tilde{\boldsymbol{x}}_k (\Bar{\boldsymbol{x}}_k))$.

\subsection{Closed-form Update Rule for Sequential Black-box Optimization}

In this section, we derive the update rule of the parameter to optimize Eq.(\ref{AugObj}).  Without loss of generality, we assume minimization in this paper. 

Given a parameter $\Bar{\theta}_K^t$ at $t^{th}$ iteration, we aim to find a new parameter $\Bar{\theta}_K^{t+1}$ by minimizing the objective difference as 
\begin{align}
\label{UpdateMinization1}
    \min \nolimits_{\Bar{\theta}_K}  J(\Bar{\theta}_K) -J(\Bar{\theta}_K^t) .
\end{align}
However, it is challenging to solve the optimization~(\ref{UpdateMinization1}) directly.  We thus optimize an approximation by first-order Taylor expansion. We add a KL-divergence regularization to ensure $q_{{\Bar{\theta}_K}} $ and $ q_{\Bar{\theta}_K^t}$ close, thus to keep the approximation accurate.  The new optimization problem is given as 
\begin{align}
    \min_{\Bar{\theta}_K}  J(\Bar{\theta}_K) -J(\Bar{\theta}_K^t) + \text{KL}(q_{{\Bar{\theta}_K}} ||  q_{\Bar{\theta}_K^t}) & =  \sum \nolimits_{k=1}^K J_k(\Bar{\theta}_k)-J_k(\Bar{\theta}_k^t) + \text{KL}(q_{{\Bar{\theta}_K}} || q_{\Bar{\theta}_K^t}) \\
    & \approx  \sum \nolimits_{k=1}^K  \beta  \left< \Bar{\theta}_k-\Bar{\theta}_k^t, \nabla_{\Bar{\theta}_k^t}J_k(\Bar{\theta}_k^t) \right> + \text{KL}(q_{{\Bar{\theta}_K}} || q_{\Bar{\theta}_K^t}), \label{UpdateRuleProblem1}
\end{align}
where $q_{{\Bar{\theta}_K}}:= \mathcal{N}(\Bar{\boldsymbol{\mu}}_K, \Bar{\boldsymbol{\Sigma}}_K)$ and $q_{{\Bar{\theta}_K^t}}:= \mathcal{N}(\Bar{\boldsymbol{\mu}}_K^t, \Bar{\boldsymbol{\Sigma}}_K^t)$. 

Note that the problem~(\ref{UpdateRuleProblem1}) is convex w.r.t. $\Bar{\theta}_K := \{ \boldsymbol{\Bar{\mu}}_{K}, \boldsymbol{\Bar{\Sigma}}_{K}\}$, by setting the derivative to zero,  we can achieve a closed-form update as 
\begin{align}\label{ExpUpdate1}
 &   \boldsymbol{\mu}_k^{t+1} =  \boldsymbol{\mu}_k^{t} - \beta \sum \nolimits_{i=k}^K \mathbb{E}_ {\mathcal{N}(\boldsymbol{\Bar{\mu}}_{i}, \boldsymbol{\Bar{\Sigma}}_{i})} [ \big(\boldsymbol{x}_k-\boldsymbol{\mu}_k^t \big)  f_i(\Bar{\boldsymbol{x}}_i) ]  \\
 & {\boldsymbol{\Sigma}_k^{t+1}}^{-1} \!=\! {\boldsymbol{\Sigma}_k^{t}}^{-1} \!+\! \beta \sum \nolimits_{i=k}^K  \mathbb{E}_ {\mathcal{N}(\boldsymbol{\Bar{\mu}}_{i}, \boldsymbol{\Bar{\Sigma}}_{i})} [ ( {\boldsymbol{\Sigma}_k^{t}}^{-1} \big( \boldsymbol{x}_k-\boldsymbol{\mu}_k^t \big) \big( \boldsymbol{x}_k-\boldsymbol{\mu}_k^t \big)^\top {\boldsymbol{\Sigma}_k^{t}}^{-1} \!-\!  {\boldsymbol{\Sigma}_k^{t}}^{-1}  )   f_i(\Bar{\boldsymbol{x}}_i) ] \label{ExpUpdate2}
\end{align}
Detailed derivation can be found in Appendix~\ref{UpdateRuleDerivation}. 

To compute the update in Eq.(\ref{ExpUpdate1}) and Eq.(\ref{ExpUpdate2}), we perform  Monte Carlo sampling by taking $N$ i.i.d. sequence $\{ \boldsymbol{x}_1^j,\cdots,\boldsymbol{x}_K^j\}$ for $j \in \{1,\cdots,N\}$,  where $\boldsymbol{x}_{k}^j \sim \mathcal{N}(\boldsymbol{\mu}_k^{t},\boldsymbol{\Sigma}_k^{t}) $ for $k \in \{1,\cdots,K\}$. This leads to unbiased estimators of the RHS of  Eq.(\ref{ExpUpdate1}) and Eq.(\ref{ExpUpdate2}) as follows: 
\begin{align}
\vspace{-10pt}
\label{UpdateMuTmp}
 &   \boldsymbol{\mu}_k^{t+1} =  \boldsymbol{\mu}_k^{t} - \frac{\beta}{N}\sum \nolimits_{i=k}^K  \sum \nolimits_{j=1}^N [ \big(\boldsymbol{x}_{k}^j-\boldsymbol{\mu}_k^t \big)  f_i(\Bar{\boldsymbol{x}}_i^j)  ]  \\
 & {\boldsymbol{\Sigma}_k^{t+1}}^{-1} \!=\! {\boldsymbol{\Sigma}_k^{t}}^{-1} \!+\!  \frac{\beta}{N}\sum \nolimits_{i=k}^K  \sum \nolimits_{j=1}^N  \Big( {\boldsymbol{\Sigma}_k^{t}}^{-1} \big( \boldsymbol{x}_{k}^j-\boldsymbol{\mu}_k^t \big) \big( \boldsymbol{x}_{k}^j-\boldsymbol{\mu}_k^t \big)^\top {\boldsymbol{\Sigma}_k^{t}}^{-1} -  {\boldsymbol{\Sigma}_k^{t}}^{-1}  \Big)   f_i(\Bar{\boldsymbol{x}}_i^j)  \label{UpdateSimgaTmp}
\end{align}
It is equivalent to the following:
\begin{align}
\vspace{-10pt}
\label{UpdateMu}
 &   \boldsymbol{\mu}_k^{t+1} =  \boldsymbol{\mu}_k^{t} - \frac{\beta}{N}  \sum \nolimits_{j=1}^N [ \big(\boldsymbol{x}_{k}^j-\boldsymbol{\mu}_k^t \big) (  \sum \nolimits_{i=k}^K f_i(\Bar{\boldsymbol{x}}_i^j)  ) ]  \\
 & {\boldsymbol{\Sigma}_k^{t+1}}^{-1} \! \!=\! {\boldsymbol{\Sigma}_k^{t}}^{-1} \!+\!  \frac{\beta}{N} \sum \nolimits_{j=1}^N  \Big( {\boldsymbol{\Sigma}_k^{t}}^{-1} \big( \boldsymbol{x}_{k}^j-\boldsymbol{\mu}_k^t \big) \big( \boldsymbol{x}_{k}^j-\boldsymbol{\mu}_k^t \big)^\top {\boldsymbol{\Sigma}_k^{t}}^{-1} \!-\!  {\boldsymbol{\Sigma}_k^{t}}^{-1}  \Big) ( \sum \nolimits_{i=k}^K  f_i(\Bar{\boldsymbol{x}}_i^j) ) \label{UpdateSimga}
\end{align}

In practice, to avoid the numerical scale problem, we normalize the cumulative score  $s_k^j= \sum_{i=k}^K  f_i(\Bar{\boldsymbol{x}}_i^j) $ 
by $h( s_k^j)=\frac{ s_k^j -s_k^{min}}{s_k^{max} -s_k^{min}}$, where $s_k^{max}$ and $s_k^{min}$ denote maximum and minimum value among of  $[s_k^1,\cdots,s_k^N]$. We thus obtain the following update rule for practical updates. 
\begin{align}
\label{UpdateMuNormalize}
 &   \boldsymbol{\mu}_k^{t+1} =  \boldsymbol{\mu}_k^{t} - \frac{\beta}{N}  \sum \nolimits_{j=1}^N [ \big(\boldsymbol{x}_{k}^j-\boldsymbol{\mu}_k^t \big) (  \frac{ s_k^j -s_k^{min}}{s_k^{max} -s_k^{min}}  ) ]   \\
 & {\boldsymbol{\Sigma}_k^{t+1}}^{-1} \!\!=\! (1-\kappa \beta  ){\boldsymbol{\Sigma}_k^{t}}^{-1} \!+\!  \frac{\beta}{N} \sum \nolimits_{j=1}^N \! \Big( {\boldsymbol{\Sigma}_k^{t}}^{-1} \big( \boldsymbol{x}_{k}^j-\boldsymbol{\mu}_k^t \big) \big( \boldsymbol{x}_{k}^j-\boldsymbol{\mu}_k^t \big)^\top {\boldsymbol{\Sigma}_k^{t}}^{-1}  \Big) ( \frac{ s_k^j \! -\!s_k^{min}}{s_k^{max} \!-\! s_k^{min}} ) \label{UpdateSimgaNormalize}
\end{align}
where $\kappa=\frac{1}{N}\sum_{j=1}^N ( \frac{ s_k^j -s_k^{min}}{s_k^{max} -s_k^{min}} ) $, and $ 0 \le \kappa \le 1$. The normalization can ensure the covariance matrix ${\boldsymbol{\Sigma}_k^{t+1}}$ to be positive semi-definite. 

Note that $\boldsymbol{x}_{k}^j= \boldsymbol{\mu}_k^t + {\boldsymbol{\Sigma}_k^{t}}^{\frac{1}{2}} \boldsymbol{z}_{k}^j $ for $\boldsymbol{z}_{k}^j \sim \mathcal{N}(\boldsymbol{0},\boldsymbol{I})$, Eq.~(\ref{UpdateSimgaNormalize})  can be rewritten as
\begin{align}
    {\boldsymbol{\Sigma}_k^{t+1}}^{-1} \!=\! {\boldsymbol{\Sigma}_k^{t}}^{-\frac{1}{2}} ( (1-\kappa \beta)\boldsymbol{I} \!+\! \beta \boldsymbol{H}_k^t  )  {\boldsymbol{\Sigma}_k^{t}}^{-\frac{1}{2}}
\end{align}
where $\boldsymbol{H}_k^t $ is constructed as Eq.~(\ref{Hkt}).
\begin{align}
    \boldsymbol{H}_k^t  = \frac{1}{N} \sum \nolimits_{j=1}^N   \boldsymbol{z}_{k}^j {\boldsymbol{z}_{k}^j}^\top ( \frac{ s_k^j -s_k^{min}}{s_k^{max} -s_k^{min}} )  \label{Hkt}
\end{align}
$\boldsymbol{H}_k^t $ serves as a pre-conditioning matrix that captures the second-order information, which  has a crucial impact on the convergence speed.  

We summarize our algorithm for black-box diffusion target generation  (BDTG)  into Algorithm~\ref{BDTG}.  In Algorithm~\ref{BDTG}, the user preference can be incorporated via the black-box target score. 
In addition, $\tilde{\sigma}_k$ is the SDE solver coefficient that depends on the concrete SDE solver used.  

\begin{algorithm}[t]
  \caption{BDTG}
  \label{BDTG}
\begin{algorithmic}[1]
    \STATE {\bf Input:} Batch Size  $N$, dimension $d$, step-size $\alpha$, a pre-trained diffusion model $\widehat{\boldsymbol{\mu}}_{\phi}(\tilde{\boldsymbol{x}}_k,k)$, number of sampling step $K$,   Number of total iteration $T$.  SDE solver coefficient  $\tilde{\sigma}_k$ for $k \in \{\1,\cdots,K\}$. 
    
    \STATE {\bf Initialization:} Initialize $\boldsymbol{\mu}_k^1=\boldsymbol{0}$,$\boldsymbol{\Sigma}_k^1= \boldsymbol{I}$ for $k \in \{1,\cdots,K\}$. 
    \FOR{{$t=1,\cdots,T$}}

  \FOR{$k=1,\cdots,K$}
  \STATE Take i.i.d. samples $\boldsymbol{x}_k^1,\cdots, \boldsymbol{x}_k^N \sim \mathcal{N}(\boldsymbol{\mu}_k^t,\boldsymbol{\Sigma}_k^t)$    
  \STATE Set $ \tilde{\boldsymbol{x}}_k^j = \widehat{\boldsymbol{\mu}}_{\phi}(\tilde{\boldsymbol{x}}^j_{k\!-\!1},{k\!-\!1})  + \tilde{\sigma}_k  \boldsymbol{x}_k^j $ for all $j \in \{1,\cdots,N\}$.
  \STATE Query black-box target function score $f_k(\Bar{\boldsymbol{x}}_k^1),\cdots,f_k(\Bar{\boldsymbol{x}}_k^N)$.
 
  \ENDFOR

  \STATE Update $\boldsymbol{\mu}_k^{t+1}$ for all $k \in \{1,\cdots,K\}$ using  Eq.~(\ref{UpdateMuNormalize}) with step-size $\beta_{}=\alpha/\sqrt{d}$.
  \STATE Update $\boldsymbol{\Sigma}_k^{t+1}$ for all $k \in \{1,\cdots,K\}$ using  Eq.~(\ref{UpdateSimgaNormalize}) with step-size  $\beta_{}=\alpha/d$.
  \ENDFOR
\end{algorithmic}
\end{algorithm}

\vspace{-8pt}
\section{Convergence Analysis}
\label{ConvergenceAnalysis}
\vspace{-4pt}

In this section, we provide the convergence analysis of a general sequential black-box optimization framework.  Algorithm~\ref{BDTG} is a special case of our framework (Alg.~\ref{MettaAlg}) with particular parameter setting and relaxation.   Without loss of generality,  we focus on minimizing the following sequential optimization problem: 
\begin{align}
\label{GerneralObj}
    \hat{F}(\Bar{\boldsymbol{x}}_K) = \sum \nolimits_{k=1}^K f_k(\Bar{\boldsymbol{x}}_k)
\end{align}
\textbf{Remark:} The black-box function  $f_k(\Bar{\boldsymbol{x}}_k)$ with $\Bar{\boldsymbol{x}}_k = [\boldsymbol{x}_1^\top,\cdots,\boldsymbol{x}_k^\top]^\top$ explicitly shows the  dependency  of the whole history trajectory $[\boldsymbol{x}_1,\cdots,\boldsymbol{x}_k]$. The sequential black-box optimization in Eq.(\ref{GerneralObj}) is general enough to include many interesting scenarios as  special cases.  One particular interesting problem is  $f_k(\Bar{\boldsymbol{x}}_k) = F(\boldsymbol{y}_k)$ where $\boldsymbol{y}_k$  is obtained by  an  unknown  transition dynamic $\boldsymbol{y}_k = Q(\boldsymbol{y}_{k-1}, \boldsymbol{x}_k )$. 

Instead of directly optimizing problem~(\ref{GerneralObj}),  we optimize an auxiliary problem~(\ref{auxiliaryOP}) as 
\begin{align}
\label{auxiliaryOP}
    J(\Bar{\boldsymbol{\mu}}_K,\Bar{\boldsymbol{\Sigma}}_K) = \sum \nolimits_{i=1}^K  J_i(\Bar{\boldsymbol{\mu}}_i,\Bar{\boldsymbol{\Sigma}}_i) =  \sum \nolimits_{i=1}^K\mathbb{E}_{\Bar{\boldsymbol{x}}_i \sim \mathcal{N}(\Bar{\boldsymbol{\mu}}_i,\Bar{\boldsymbol{\Sigma}}_i)}[ f_i(\Bar{\boldsymbol{x}}_i) ]
\end{align}
where $J_i(\Bar{\boldsymbol{\mu}}_i,\Bar{\boldsymbol{\Sigma}}_i) = \mathbb{E}_{\Bar{\boldsymbol{x}}_i \sim \mathcal{N}(\Bar{\boldsymbol{\mu}}_i,\Bar{\boldsymbol{\Sigma}}_i)}[ f_i(\Bar{\boldsymbol{x}}_i) ]$ denotes the $i^{th}$ sub-objective.

Denote gradient estimator  $\hat{{g}}_{ik}^t$ for the $i^{th}$ sub-objective w.r.t. the $k^{th}$ component ${\boldsymbol{\mu}_k}$  at $t^{th}$ iteration as 
    \begin{align}
    \label{UnbiasedEstg}
       \hat{{g}}_{ik}^t & = \frac{1}{N}  \sum \nolimits_{j=1}^N {\hat{{g}}_{ik}}^{tj} 
       =  \frac{1}{N}  \sum \nolimits_{j=1}^N
       {\boldsymbol{\Sigma}_k^t}^{-\frac{1}{2}}\boldsymbol{z}_k^j\big(f_i(\Bar{\boldsymbol{\mu}}_i^t +{\Bar{\boldsymbol{\Sigma}}_i}^{t\frac{1}{2}}\Bar{\boldsymbol{z}_i}^j)-f_i(\Bar{\boldsymbol{\mu}}_i^t)\big),
    \end{align}
where ${\hat{{g}}_{ik}}^{tj}$ is the gradient estimator using $j^{th }$  sample:
\begin{align}
    \hat{{g}}_{ik}^{tj} = {\boldsymbol{\Sigma}_k^t}^{-\frac{1}{2}}\boldsymbol{z}_k^j\big(f_i(\Bar{\boldsymbol{\mu}}_i^t +{\Bar{\boldsymbol{\Sigma}}_i}^{t\frac{1}{2}}\Bar{\boldsymbol{z}_i}^j)-f_i(\Bar{\boldsymbol{\mu}}_i^t)\big),
\end{align}
 where $\boldsymbol{z}_1^j,\cdots,\boldsymbol{z}_i^j\sim\mathcal{N}(\boldsymbol{0},\boldsymbol{I}_{d})$ and $\Bar{\boldsymbol{z}_i}^j=[\boldsymbol{z}_1^\top,\cdots,\boldsymbol{z}_i^\top]^\top$ for $i\ge k$.

We show our Covariance-Adaptive Sequential Black-box  Optimization algorithm (CASBO) in Algorithm~\ref{MettaAlg}. Our CASBO can perform full matrix updates to take advantage of second-order information.  Our Algorithm~\ref{BDTG} is a special case of our CASBO framework with a particular parameter setting and relaxes the constraints.  Our CASBO framework only requires the unbiased gradient estimator for $\boldsymbol{\mu}$, while it does not require an unbiased gradient estimator for covariance $\boldsymbol{\Sigma}$.

\begin{algorithm}[t]
  \caption{CASBO Framework}
  \label{MettaAlg}
\begin{algorithmic}[1]
    \STATE {\bf Input:} Batch-size $N$. {Parameter  $\beta_t$,  $\alpha_t$, $\gamma_t$,and $\omega_t$. Number of total iteration $T$ and the step $K$.}
    \STATE {\bf Initialization:} Initialize $\boldsymbol{\mu}_k^1=\boldsymbol{0}$,$\boldsymbol{\Sigma}_k^1= \tau \boldsymbol{I}$  for $k \in \{1,\cdots,K\}$. 
 \FOR{{$t=1,\cdots,T$}}  
  \FOR{$k=1,\cdots,K$}
  \STATE Take i.i.d. samples $\boldsymbol{z}_k^1,\cdots, \boldsymbol{z}_k^N \sim \mathcal{N}(\boldsymbol{0},\boldsymbol{I})$     
  \STATE Set $\boldsymbol{x}_k^j = \boldsymbol{\mu}_k^t + {\boldsymbol{\Sigma}_k^t}^{\frac{1}{2}} \boldsymbol{z}_k^j $ for $j \in \{1,\cdots,N\}$
  
  \STATE Query black-box objective function value $f_k(\Bar{\boldsymbol{x}}_k^1),\cdots,f_k(\Bar{\boldsymbol{x}}_k^N)$.
  \STATE Construct unbiased estimator $\hat{g}_{k1},\cdots,\hat{g}_{kk}$ as Eq.~(\ref{UnbiasedEstg}).
  
  \ENDFOR

  \FOR{$k=1,\cdots,K$}
    \STATE Compute $s_k^j= \sum_{i=k}^K  f_i(\Bar{\boldsymbol{x}}_i^j) $ for $j \in \{1,\cdots,N\}$.   
  \STATE Compute $s_k^{max}$ and $s_k^{min}$  among of  $[s_k^1,\cdots,s_k^N]$.
   \STATE Construct $ \boldsymbol{H}_k^t  =  c_1 \frac{1}{N} \sum_{j=1}^N   \boldsymbol{z}_{k}^j {\boldsymbol{z}_{k}^j}^\top ( \frac{ s_k^j -s_k^{min}}{s_k^{max} -s_k^{min}} )   +  c_2 \boldsymbol{I} $ with constants $c_1>0,c_2>0$  such that  $\boldsymbol{H}_k^t \preceq \frac{1}{\alpha_t}(\frac{\beta_{t+1}}{\beta_t} - \omega_t ) \boldsymbol{I} +   \frac{\beta_{t+1} \gamma_t}{\alpha_t} {\boldsymbol{\Sigma}_{k}^{t}} $ and $\nu \boldsymbol{I}  \preceq 
 \hat{G}_k^t=  {\boldsymbol{\Sigma}_k^t}^{-\frac{1}{2}} \boldsymbol{H}_k^t {\boldsymbol{\Sigma}_k^t}^{-\frac{1}{2}} $
  
\STATE Set $ \hat{G}_k^t=  {\boldsymbol{\Sigma}_k^t}^{-\frac{1}{2}} \boldsymbol{H}_k^t {\boldsymbol{\Sigma}_k^t}^{-\frac{1}{2}}$.
    \STATE Set $\boldsymbol{\mu}_k^{t+1} = \boldsymbol{\mu}_k^{t} - \beta_t \boldsymbol{\Sigma}_k^{t} \big( \gamma_t \boldsymbol{\mu}_k^{t} + (\sum_{i=k}^K \hat{g}_{ik} )  \big)$  
  \STATE Set $ {\boldsymbol{\Sigma}_k^{t\!+\!1}}^{-\!1} =  \omega_t {\boldsymbol{\Sigma}_k^{t}}^{-\!1} +  \alpha_t \hat{G}_k^t $. 
  \ENDFOR

   \ENDFOR
\end{algorithmic}
\end{algorithm}

We now list the assumptions employed in our convergence analysis. All the assumptions are common in the literature. The assumptions are weak because neither smooth assumptions nor strongly convex assumptions are involved.  
Thus, our algorithm can handle non-smooth cases. More importantly, we do not add any additional assumptions of the auxiliary problem~(\ref{auxiliaryOP}).  This is important for practical use because we can not check whether the auxiliary problem satisfies the assumptions given a black-box original problem.    To the best of our knowledge, our algorithm is the first full matrix adaptive black-box optimization algorithm that achieves a provable $O( \frac{d^2K^4   }{\sqrt{T}})$ convergence for convex functions without smooth and strongly convex assumptions, and any assumptions of the auxiliary problems.

\begin{assumption}\label{Assumption1}
    $f_1(\Bar{\boldsymbol{x}}_1),\cdots, f_K(\Bar{\boldsymbol{x}}_K)$ are all convex functions.
\end{assumption}

\begin{assumption}\label{Assumption2}
     $f_i(\Bar{\boldsymbol{x}}_i) $ is a $L_i$-Lipschitz continuous function for $ \forall i \in \{1,\cdots,K\}$, i.e., $|f_i(\Bar{\boldsymbol{x}}_i) - f_i(\Bar{\boldsymbol{y}}_i) | \le L_i \|\Bar{\boldsymbol{x}}_i -\Bar{\boldsymbol{y}}_i \|_2$. 
\end{assumption}

\begin{assumption}\label{Assumption4}
    The initialization $\Bar{\theta}_K^1:=\{ \Bar{\boldsymbol{\mu}}_K^1,\Bar{\boldsymbol{\Sigma}}_K^1\}$ is bounded, i.e., $ \sum_{k=1}^K \| \boldsymbol{\mu}_k^{1}  -  \boldsymbol{\mu}_k^*   \|_{ {\boldsymbol{\Sigma}_{k}^{1}}^{\!-\!1} }^2  \le R$ and $\Bar{\boldsymbol{\Sigma}}_K^1 \in \mathcal{S}^{+}$, and  $  \underline{\nu} \boldsymbol{I} \preceq  \Bar{\boldsymbol{\Sigma}}_K^1  \preceq \Bar{\nu}  \boldsymbol{I} $ for $\Bar{\nu} \ge \underline{\nu} > 0$, and $\sum_{k=1}^K \| \boldsymbol{\mu}_k^*  \|_2^2 \le B$.
\end{assumption}

\begin{theorem}
\label{MainTheorem}
    Suppose the assumptions~\ref{Assumption1}~\ref{Assumption2}~\ref{Assumption4} holds.       Set $\beta_t = t\beta$ with $\beta>0$,  $\alpha_t=  \sqrt{t+1} \alpha $ with  $ \alpha > 0$, and $\gamma_t = \frac{\alpha\nu}{\beta\sqrt{t+1}}$, and $\nu>0$, and  $\omega_t =1$.  Initialize  $\boldsymbol{\Sigma}_{k}^1 $ such that $\| \boldsymbol{\Sigma}_{k}^1 \|_2^{-1} \ge \frac{5}{3}\alpha \nu$  for $\forall k \in \{1,\cdots,K\}$.            Then, running Algorithm~\ref{MettaAlg} with  $T$-steps, we have 
    \begin{align}
    \label{TheoremConvergenceRate}
      \frac{1}{T}\sum_{t=1}^T  \sum_{k=1}^K f_k (\Bar{\boldsymbol{\mu}}_k^t) \!-\! \sum_{k=1}^Kf_k(\Bar{\boldsymbol{\mu}}_k^*) & \le \frac{\sum_{k=1}^K \| \boldsymbol{\mu}_k^{1} \! - \! \boldsymbol{\mu}_k^*   \|_{ {\boldsymbol{\Sigma}_{k}^{1}}^{\!-\!1} }^2   }{2\beta T } \!+\! \frac{ 2\sqrt{T\!+\!1}  C_1 }{T} \!+\! \frac{4(T\!+\!1)^{\frac{1}{4}}C_2}{T } \! +\! \frac{\sqrt{T\!+\!2}C_3}{T}  \nonumber \\ & \le  O( \frac{d^2K^4   }{\sqrt{T}})
    \end{align}
    where $\Bar{\boldsymbol{\mu}}_k^t=[\boldsymbol{\mu}_1^{t\top},\cdots,\boldsymbol{\mu}_k^{t\top}]^\top$ and $\Bar{\boldsymbol{\mu}}_k^*=[\boldsymbol{\mu}_1^{*\top},\cdots,\boldsymbol{\mu}_k^{*\top}]^\top$. And  $C_1 =   \frac{ 3\beta    \sum_{i=1}^K K L_i^2  (id+1)^2}{2\alpha \nu} $  and  $C_2 =  \frac{\sum_{i=1}^K \sqrt{3}idL_i}{\sqrt{ \alpha \nu}} \ $ , $C_3= \frac{\alpha\nu B}{\beta}$ .
\end{theorem}
Detailed proof can be found in Appendix~\ref{TheoremProof}. In Theorem~\ref{MainTheorem},  the error term $\frac{\sqrt{T\!+\!2}C_3}{T} $ in Eq.(\ref{TheoremConvergenceRate}) is due to the $\gamma_t$-enlargement. Error term $\frac{4(T\!+\!1)^{\frac{1}{4}}C_2}{T }$ results from the  covariance matrix term $\Bar{\boldsymbol{\Sigma}}_K$ in Gaussian-smooth relaxation of the original problem. The first two error terms result from the stochastic gradient update.

\textbf{Remark:} Note that for convex problems, the optimum $\{\Bar{\boldsymbol{\mu}}_K^*, 0\}$ of the auxiliary problem~(\ref{auxiliaryOP})  is also the optimum of the original   problem~(\ref{GerneralObj}), i.e., $J(\Bar{\boldsymbol{\mu}}_K^*, 0) = \hat{F}(\Bar{\boldsymbol{\mu}}_K^*)$.   Our algorithm achieve a $O(\frac{d^2K^4}{\sqrt{T}})$ cumulative regret of the original problem~(\ref{GerneralObj}). In addition,  our algorithm can handle non-smooth problems without expert designing of proximal operators for different types of non-smooth functions. This can be remarkably interesting when the unknown function involves compositions of lots of different types of non-smooth functions, in which case human experts can not derive the operators explicitly.

\section{Experiments}
\label{Experiments}
\vspace{-6pt}
\subsection{Baseline Methods}
\vspace{-8pt}

We compare our method with the baseline black-box optimization algorithms: CMA-ES~\citep{hansen2001completely} and TuRBO~\citep{eriksson2019scalable}. For CMA-ES,  we use the official implementation on Github~\citep{hansen2019pycma}.  We employ the default hyperparameter settings in all of our experiments.  We set the initial covariance matrix to the identity matrix, which is the same as our algorithm. For TuRBO,  we use the publicly available official implementation on GitHub~\footnote[1]{https://github.com/uber-research/TuRBO}. We employ the default hyperparameter settings in all experiments, including the number of trust regions set to $n=5$. More details of the related black-box optimization can be found in Appendix~\ref{Background}.
Both CMA-ES and TuRBO do not explicitly consider the sequential inter-dependency of the input variables ${\boldsymbol{x}_1, \cdots, \boldsymbol{x}_K}$. Thus, we directly optimize the sequential optimization problem with respect to the concatenated variables $\Bar{\boldsymbol{x}}_K = [ \boldsymbol{x}_1^\top, \cdots, \boldsymbol{x}_K^\top ] \in \mathbb{R}^{Kd}$.

\begin{table}[t]
\centering
\caption{Objective Function Definition}
\begin{tabular}{|l|l|}
\hline
Rastrigin-10& $ f(\boldsymbol{x}):=10 d+\sum_{i=1}^d(10^{\frac{i-1}{d-1}} x_i)^2-10 \cos (2 \pi 10^{\frac{i-1}{d-1}} x_i)$ \\ \hline
$L_1$-Ellipsoid & $ f({\boldsymbol{x}}):=\sum\nolimits_{i = 1}^d {10^{\frac{6(i-1)}{d-1}}|x_i|}$ \\ \hline
Levy& \scalebox{0.82}{\makecell[l]{ $f(\boldsymbol{x}):=\sin^2(\pi w_1) + \sum_{i=1}^{d-1} (w_i - 1)^2 \left[1 + 10 \sin^2(\pi w_i + 1)\right] + (w_d - 1)^2 \left[1 + \sin^2(2\pi w_d)\right]$, \\ where $w_i = 1 + \frac{x_i - 1}{4}$}} \\ \hline
\end{tabular}
\label{ToyTestFunctionDefinition}
\vspace{-10pt}
\end{table}

\begin{figure}[t]
\centering
\subfigure[\scriptsize{Cumulative-Rastrigin10}]{
\label{ToyTask0}
\includegraphics[width=0.32\linewidth]{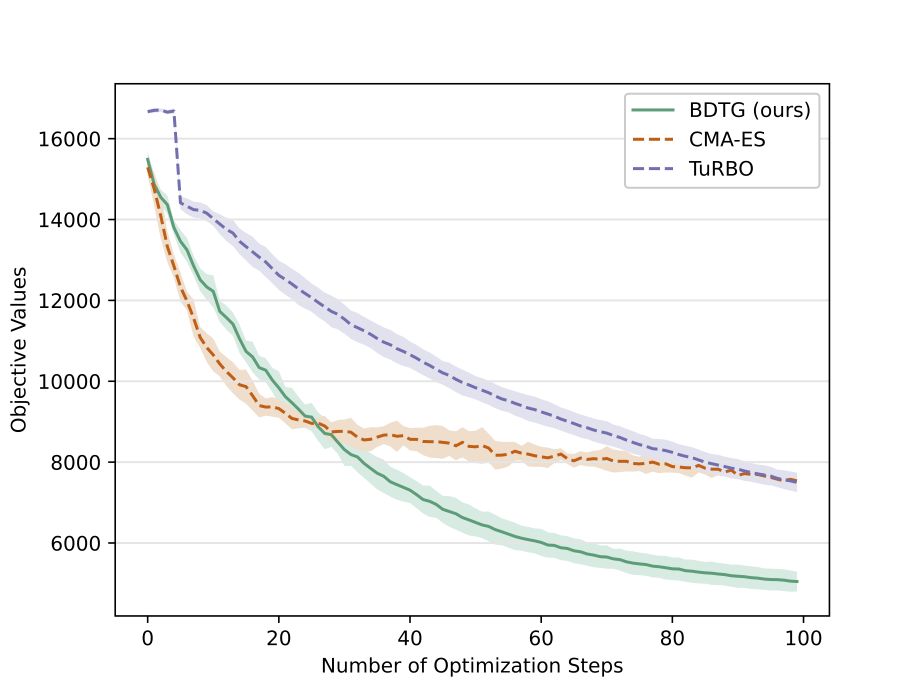}}
\subfigure[\scriptsize{Cumulative-L1-Ellipsoid}]{
\label{ToyTask1}
\includegraphics[width=0.32\linewidth]{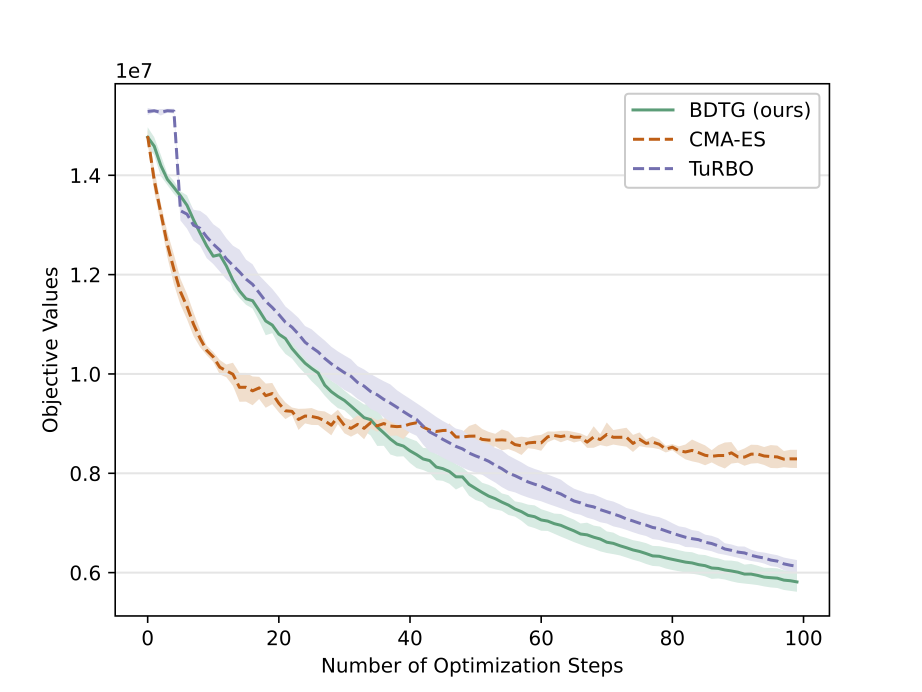}}
\subfigure[\scriptsize{Cumulative-Levy}]{
\label{ToyTask2}
\includegraphics[width=0.32\linewidth]{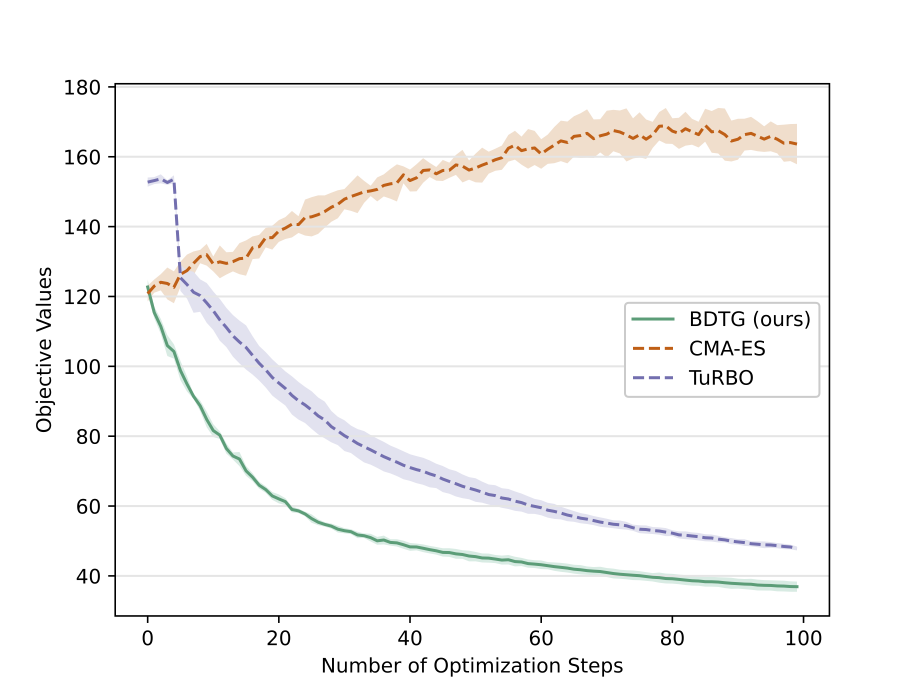}}
\caption{Objective Values (Cumulative Target Score, Lower is Better) v.s. Number of
Optimization Steps on Different Test Problems}
\label{ToyTestFunction}
\vspace{-10pt}
\end{figure} 

\vspace{-5pt}
\subsection{Empirical Study on Numerical Test Problem}
\vspace{-8pt}
We first evaluate our algorithm using the numerical sequential optimization problem $ \hat{F}(\Bar{\boldsymbol{x}}_K) = \sum_{k=1}^K f(\boldsymbol{x}_k)$, where $\boldsymbol{x}_k \in \mathbb{R}^d$. The black-box transition dynamics are given by $\boldsymbol{x}_k= \boldsymbol{Q}\boldsymbol{x}_{k-1} + \sqrt{k+1}$, where the $\boldsymbol{Q} \in \mathbb{R}^{d \times d}$ is random rotation matrix. We set the number of transitions $K=10$ and dimension $d=100$. For our algorithm, we use step size $\alpha=10$. For CMA-ES, we leave all hyper-parameters as default. For TuRBO, we set the parameters bound as $\pm 2$ and leave all other hyper-parameters as default. We test each algorithm on three different types of objective functions as defined in Table~\ref{ToyTestFunctionDefinition}. The initial values of $\boldsymbol{x}_k$ for $k\in\{1,\cdots,K\}$ are set to zeros, and optimization is performed for 100 steps, repeated over 5 independent runs. The details of compute resource are shown in Appendix \ref{compute_resources}.

The convergence curves are shown in Figure \ref{ToyTestFunction}, where the lines represent the average objective values across the 5 runs, and the shaded areas indicate the range within one standard deviation. 
\textbf{CMA-ES} converges the fastest among all algorithms on the Rastrigin10 and L1-Ellipsoid problems during the initial stages (within the first 20 steps) but shows fluctuations and plateaus afterward. Additionally, CMA-ES fails to converge on the Levy problem. This behavior could be due to the greedy optimization strategy of CMA-ES.
\textbf{TuRBO} does not converge within the first 5 steps due to the initialization period of its 5 local Gaussian Process (GP) models. Afterward, TuRBO converges consistently on all three problems.
\textbf{BDTG (ours)} algorithm converges to the lowest objective value among all algorithms on all three problems. It achieves marginally lower values on the L1-Ellipsoid problem and significantly lower objectives on the Rastrigin-10 and Levy problems. The latter problems are considered more challenging due to their high curvature and multi-mode issues, indicating our algorithm's superior performance on challenging problems. 

\subsection{Black-box Diffusion Targeted 3D-molecule Generation}
\label{molecules_generation}
\vspace{-8pt}

\textbf{Pre-trained Model: TargetDiff}. TargetDiff~\citep{guan20233d} is a diffusion model that can generate molecules in 3D space conditioned on the target protein. TargetDiff learns the 3D structure-based molecule representation, which is composed of both categorical atom types and continuous atom coordinates. The discrete atom types are modeled using the categorical distribution $\mathcal{C}$, and the continuous atom coordinates are modeled by Gaussian distribution $\mathcal{N}$. During the inference process, TargetDiff generates the atom types and atom coordinates concurrently in the single neural network. In our experiment, we use the official implementation on Github \footnote[1]{https://github.com/guanjq/targetdiff}.

\textbf{Dataset}.
The TargetDiff is pre-trained on the CrossDocked2020~\citep{francoeur2020three} dataset. It contains 22.5 million poses of ligands docked into multiple similar binding pockets across the Protein Data Bank. It is further refined by only selecting the poses with a low Root Mean Square Deviation (RMSD) ($<$1\r{A} ) and sequence identity less than 30\%. Ultimately, 100,000 molecules are used to train the TargetDiff, and 100 for testing~\citep{guan20233d,luo20213d}. It is worth noting that our fine-tuning algorithm does not require the dataset; our learning signal is based solely on the model-generated samples.

\textbf{Objective Function: Molecules Binding Affinity}.
The binding affinity quantification refers to the strength of the interaction between two bio-molecules. In the drug discovery task, the goal is to discover the drug molecule (often called ligand) that has a high binding affinity with the target protein receptor.
The binding affinity can be conveniently predicted by using the AutoDock Vina software~\citep{eberhardt2021autodock} with the Vina docking score. The Vina docking score is the predicted free energy of the binding between the drug molecule and its target protein, expressed kcal/mol. The lower the energy, the higher the binding strength. Thus, we always aimed to minimize the Vina docking score.

Recall from section \ref{Approach}, we consider a sequential optimization problem, which requires the objective function to evaluate the "noisy" data point along the diffusion sampling trajectory (ie, ${F}(\tilde{\boldsymbol{x}}_k)$ for $k < K$). However, directly evaluating the noise data using the Vina docking software can lead to unstable behavior. Therefore, for any noisy data point, we use the pre-trained model to continue the sampling process until it reaches the final step and denote the final sample as $\tilde{\boldsymbol{x}}_{k \rightarrow K}$. Note that this complementary sampling process is deterministic and not guided by the fine-tuning parameters; thus, it can be absorbed by the objective function. As a result, we can mathematically write the final objective function as $F_k(\tilde{\boldsymbol{x}}_k) = {F}(\tilde{\boldsymbol{x}}_{k\rightarrow K})$, which takes a noisy sample as input and performs the diffusion sampling internally.

\textbf{Experiment Details}. In this experiment, we fine-tune the TargetDiff model to generate optimized molecules with minimized Vina docking scores. Recall that TargetDiff concurrently generates atom types and atom coordinates. In our fine-tuning process, we keep the atom types fixed and only fine-tune the atom coordinates. To determine the atom types for fine-tuning, we first generate 100 molecules using the pre-trained TargetDiff model. We then select the molecule with the lowest Vina docking score and use its atom types for subsequent fine-tuning steps.

To keep the chosen atom types fixed throughout the diffusion sampling process, for each denoising step, the chosen atom types are fed as the model input, and the denoised atom types from the model output are simply ignored. This feature is implemented in the official TargetDiff called \texttt{pose-only} mode.

The original TargetDiff uses DDPM~\citep{ho2020denoising} sampler, which requires 1000 sampling steps. We speed up the sampling process by using DPM-Solver++ SDE sampler~\citep{lu2022dpm}, which requires only $K=10$ sampling steps. By switching to the faster solver, the molecule's binding affinity decreases. However, as we will demonstrate later, the fine-tuned TargetDiff is capable of generating higher binding affinity with the faster sampler than the original model with the slower sampler.
In practice, we fine-tune not only the diffusion trajectory but also the prior distribution. Specifically, the prior distribution is parameterized by the fine-tuning parameters. Mathematically, for the first step $k=1$ in the Equation \ref{DiffusionSampling}, we set $\widehat{\boldsymbol{\mu}}_\phi(\tilde{\boldsymbol{x}}_{0}, 0) = \boldsymbol{0}$, resulting in the prior $\tilde{\boldsymbol{x}}_{1} \sim \mathcal{N}(\boldsymbol{\mu}_1, \boldsymbol{\Sigma}_1)$. 

We fine-tune the TargetDiff model for $T=300$ optimization steps. For all the methods, we use the same batch size $N=32$. For our algorithm, we set step size $\alpha = 10$. The source code will be made publicly available upon publication.  For CMA-ES, we use the default parameter settings, including the initial covariance matrix set as identity, which is consistent with our algorithm. For TuRBO, we set the parameters bound as $\pm 4$ and use the default settings for the other hyper-parameters, including the number of trust regions set to $n=5$. The details of compute resources are shown in Appendix \ref{compute_resources}.

\begin{figure}[t]
\centering
\subfigure[\scriptsize{Receptor-0}]{
\label{task0}
\includegraphics[width=0.32\linewidth]{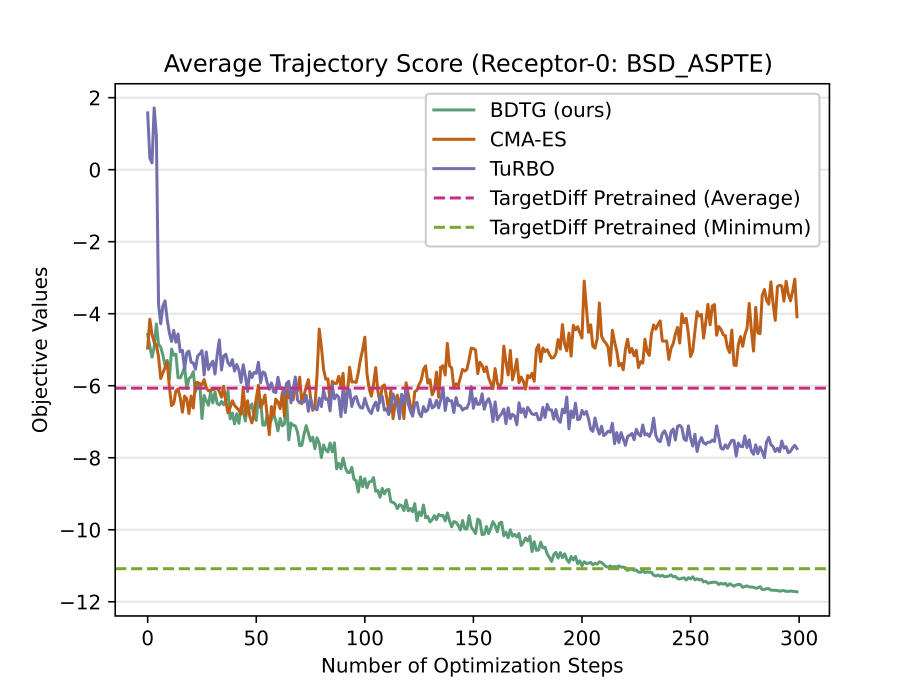}}
\subfigure[\scriptsize{Receptor-1}]{
\label{task1}
\includegraphics[width=0.32\linewidth]{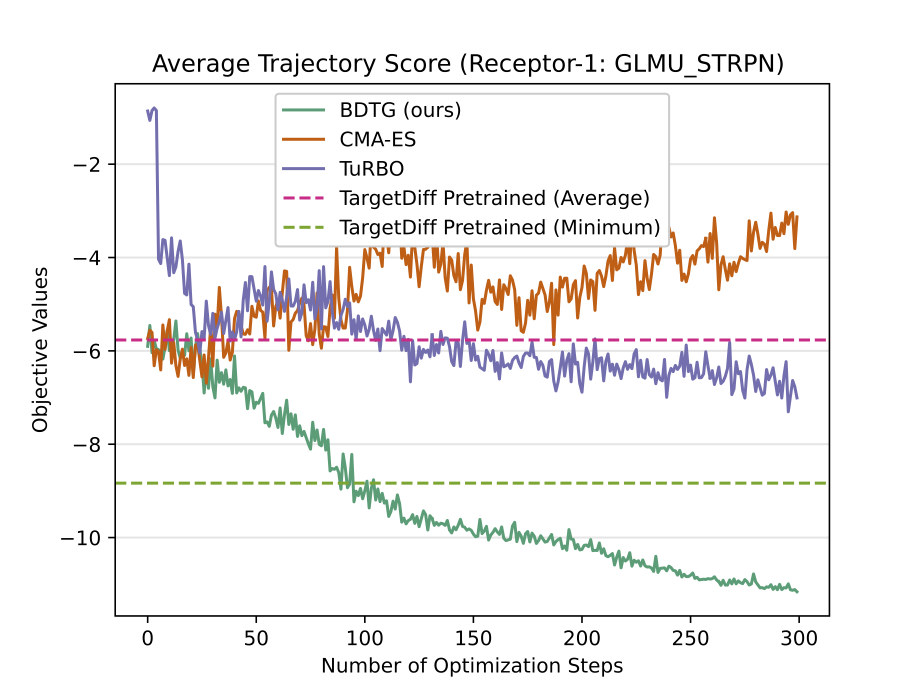}}
\subfigure[\scriptsize{Receptor-2}]{
\label{task2}
\includegraphics[width=0.32\linewidth]{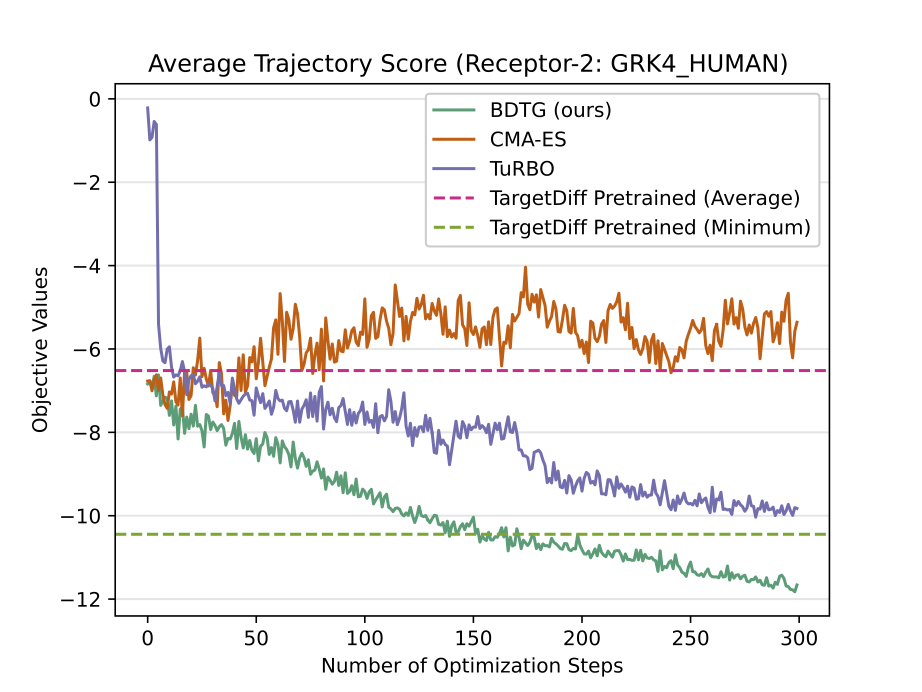}}
\subfigure[\scriptsize{Receptor-3}]{
\label{task3}
\includegraphics[width=0.32\linewidth]{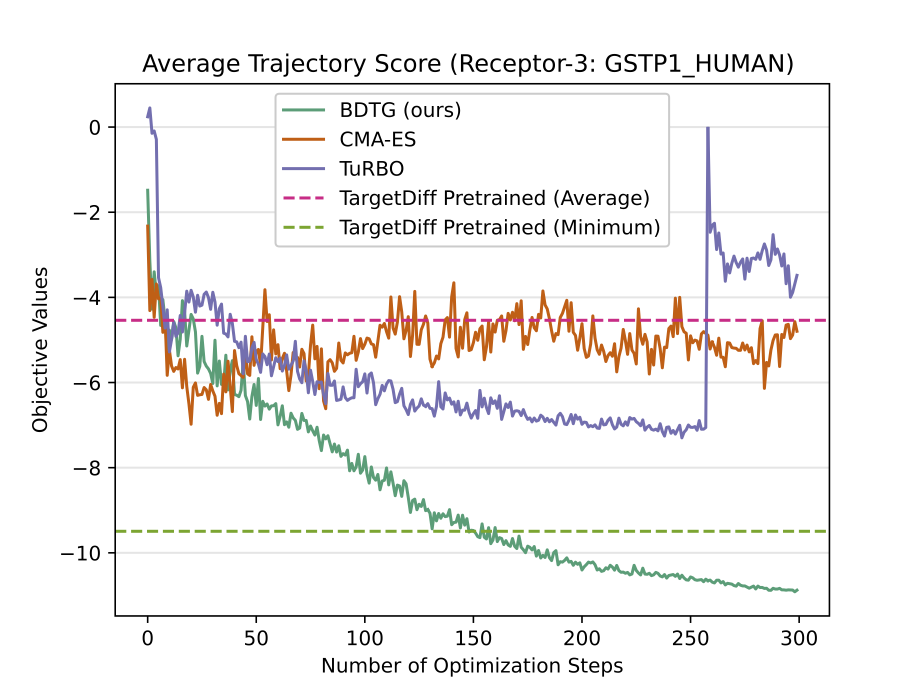}}
\subfigure[\scriptsize{Receptor-4}]{
\label{task4}
\includegraphics[width=0.32\linewidth]{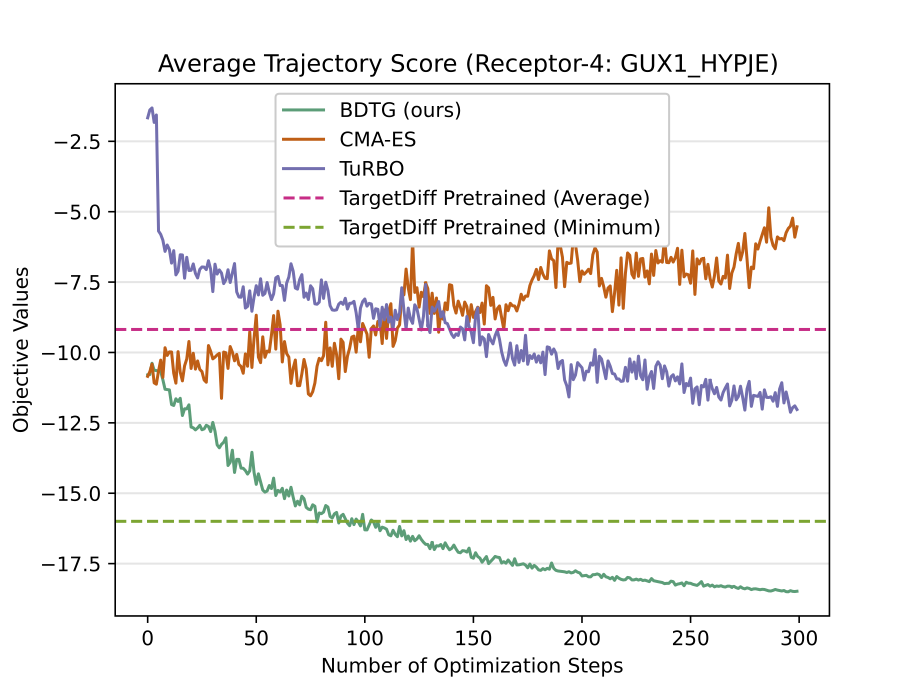}}
\subfigure[\scriptsize{Receptor-5}]{
\label{task5}
\includegraphics[width=0.32\linewidth]{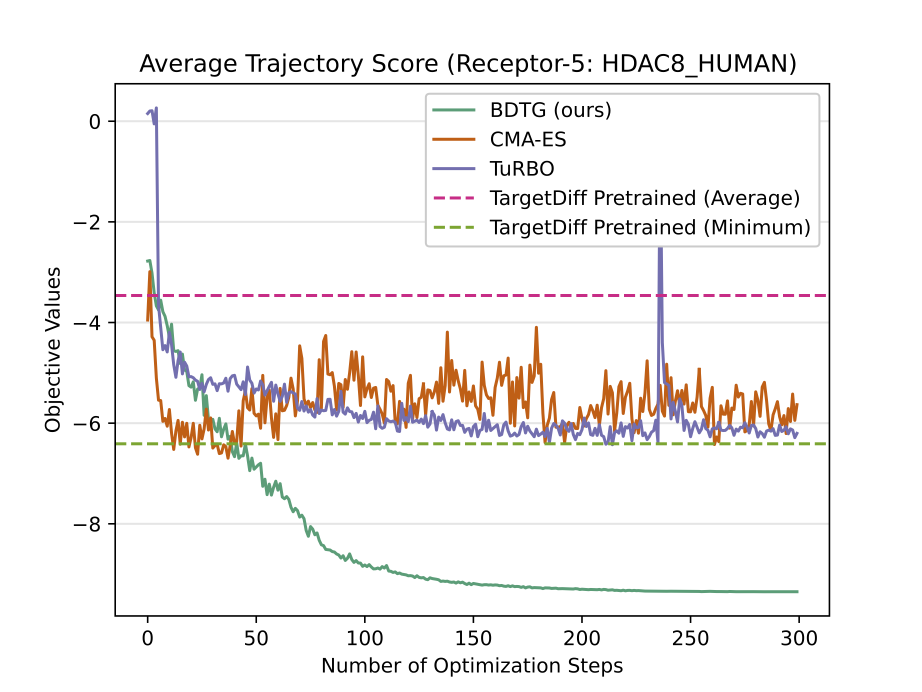}}
\caption{ Objective Values (Cumulative Vina Docking Score, Lower is Better) v.s. Number of Optimization Steps (Fine-tune Steps) for Different Receptors }
\label{TestFunction}
\vspace{-15pt}
\end{figure} 

\textbf{Experimental Results}. We compare diffusion fine-tuning using three algorithms on 6 protein receptor targets (i.e., the 0-th to 5-th from the testing set of the CrossDocked2020 dataset), respectively. We report the cumulative Vina docking score with respect to the fine-tuning steps in Figure~\ref{TestFunction}.

We compare the fine-tuned scores of each algorithm. CMA-ES can improve rapidly during the initial stages but fails to converge, exhibiting similar behavior as observed in our numerical experiments. TuRBO generally converges well, except for receptors 3 and 5, where the objective value spikes were observed due to the local GP restarts in their official code when the trust region shrinks below the preset threshold of $2^{-7}$. Our BDTG algorithm consistently converges to the lowest score compared to the other algorithms for all receptors.

As a reference, we also compare our results with the pre-trained TargetDiff model. The average and minimum (best) Vina scores of the 100 samples generated by the pre-trained TargetDiff model are shown as horizontal lines in Figure \ref{TestFunction}. Note that these samples are generated using the default DDPM sampler, which requires 1000 steps, whereas our fine-tuned samples require only 10 steps. Our fine-tuned model generates significantly better molecules than the best molecule produced by the pre-trained model. The best molecules generated during the fine-tuning process are demonstrated in Appendix \ref{demo}. Additionally, we demonstrate our algorithm in the targeted image generation task using Stable Diffusion in Appendix~\ref{TargetedImage}.

\vspace{-8pt}
\section{Conclusion and Future Work}\label{Conclusion}\vspace{-9.5pt}
In this paper, we proposed a novel targeted SDE fine-tuning framework for diffusion black-box targeted generation. Furthermore, we 
formulated the targeted fine-tuning as a sequential black-box optimization problem. We proposed a novel covariance-adaptive sequential black-box optimization(CASBO) algorithm to address this problem. Theoretically, we prove a $O(\frac{d^2K^4}{\sqrt{T}})$ convergence rate of CASBO  without smooth and strongly convex assumptions.  Thus, our theoretical results hold true for all non-smooth/smooth convex function families that are of great challenge for full covariance matrix adaptive black-box algorithms to converge.  Empirically, our method enables the fine-tuning of the diffusion model to generate targeted 3D-Molecules with a lower Vina Docking score.  The limitation of this paper is the optimization may decrease the diversity of the generation. This limitation may be because constant function learning is not sufficient to capture the diversity. This may be addressed by learning the function $\mu(\cdot)$ and $\Sigma(\cdot)$ instead of the constant function learning scheme. We leave this function learning approach as future work.


\bibliography{neurips_2024}
\bibliographystyle{plainnat}

\newpage

\section*{\centering {\Large Appendix }}

\section{Derivation of Update Rule}\label{UpdateRuleDerivation}
The  minimization can be rewritten as
\begin{align}
    &\sum_{k=1}^K \left< \Bar{\theta}_k-\Bar{\theta}_k^t, \beta \nabla_{\Bar{\theta}_k^t}J_k(\Bar{\theta}_k^t) \right> + \text{KL}(q_{{\theta}} \| q_{{\theta}^t}) =  \sum_{k=1}^K \beta \Bar{\boldsymbol{\mu}}_k^\top\nabla_{\Bar{\boldsymbol{\mu}}_k^t}\mathbb{E}_{q_{\Bar{\theta}_k^t}}[f_k(\Bar{\boldsymbol{x}}_k)] +  \nonumber\\ 
    &\sum_{k=1}^K \beta \text{tr}( \Bar{\Sigma}_k  \nabla_{\Bar{\boldsymbol{\Sigma}}_k^t}\mathbb{E}_{q_{\Bar{\theta}_k^t}}[f_k(\Bar{\boldsymbol{x}}_k )]) \!+\! \frac{1}{2}\big[\text{tr}(\Bar{\Sigma^t_K}^{\!-\!1}\Bar{\Sigma}_K)+  (\Bar{\boldsymbol{\mu}}_K \!-\!\Bar{\boldsymbol{\mu}}^t_K)^\top{\Bar{\Sigma^t_K}}^{\!-\!1}(\Bar{\boldsymbol{\mu}}_K \!-\!\Bar{\boldsymbol{\mu}}^t_K) +\log\frac{|\Bar{\Sigma}^t_K|}{|\Bar{\Sigma}_K|} \!-\!d \big],
\end{align}
where $\nabla_{\Bar{\boldsymbol{\mu}}_k^t}\mathbb{E}_{q_{\Bar{\theta}_k^t}}[f_k(\Bar{\boldsymbol{x}}_k)]$ and $\nabla_{\Bar{\boldsymbol{\Sigma}}_k^t}\mathbb{E}_{q_{\Bar{\theta}_k^t}}[f_k(\Bar{\boldsymbol{x}}_k )]$ denotes the derivative w.r.t $\Bar{\boldsymbol{\mu}}_k$ and $\Bar{\boldsymbol{\Sigma}}_k$ taking at $\Bar{\boldsymbol{\mu}}_k=\Bar{\boldsymbol{\mu}}_k^t$ and $\Bar{\boldsymbol{\Sigma}}_k=\Bar{\boldsymbol{\Sigma}}_k^k$.  The above problem is convex with respect to $\Bar{\boldsymbol{\mu}}_K=[\boldsymbol{\mu}_1^\top,\cdots,\boldsymbol{\mu}_K^\top]^\top$ and $\Bar{\boldsymbol{\Sigma}}_K = diag(\boldsymbol{\Sigma}_1,\cdots,\boldsymbol{\Sigma}_K)$. Taking the derivative w.r.t $\Bar{\boldsymbol{\mu}}_K$ and $\Bar{\boldsymbol{\Sigma}}_K$ and  setting them to zero,  for  $k^{th}$ component ,  we can obtain that 
\begin{align}
    & \sum_{i=k}^K \beta \nabla_{{\boldsymbol{\mu}}_k^t}\mathbb{E}_{q_{\Bar{\theta}_i^t}}[f_i(\Bar{\boldsymbol{x}}_i)] + {\boldsymbol{\Sigma}_k^t}^{-1}(\boldsymbol{\mu}_k-\boldsymbol{\mu}_k^t) =0 \\
    & \sum_{i=k}^K \beta \nabla_{{\boldsymbol{\Sigma}}_k^t}\mathbb{E}_{q_{\Bar{\theta}_i^t}}[f_i(\Bar{\boldsymbol{x}}_i )] + \frac{1}{2}[\boldsymbol{\Sigma}_k^{-1}-{\boldsymbol{\Sigma}^t_k}^{-1}]=0.
\end{align}
for $k \in \{1,\cdots,K \}$.

Set $\boldsymbol{\mu}_k^{t+1} = \boldsymbol{\mu}_k $ and ${\boldsymbol{\Sigma}_k^{t\!+\!1}}^{-\!1} =\boldsymbol{\Sigma}_k^{-1}$ in the above equation.  We then have the update rule as 
\begin{align}
     & \boldsymbol{\mu}_k^{t+1} = \boldsymbol{\mu}_k^t -     \sum_{i=k}^K \beta {\boldsymbol{\Sigma}_k^t}\nabla_{{\boldsymbol{\mu}}_k^t}\mathbb{E}_{q_{\Bar{\theta}_i^t}}[f_i(\Bar{\boldsymbol{x}}_i)]  \\
    &{\boldsymbol{\Sigma}_k^{t\!+\!1}}^{-\!1} =  {\boldsymbol{\Sigma}^t_k}^{-\!1} +   \sum_{i=k}^K 2\beta \nabla_{{\boldsymbol{\Sigma}}_k^t}\mathbb{E}_{q_{\Bar{\theta}_i^t}}[f_i(\Bar{\boldsymbol{x}}_i )]    .
\end{align}
for $k \in \{1,\cdots,K \}$.

In addition, note that the gradient has the following closed-form~\citep{NES}
\begin{align}
&  \nabla_{{\boldsymbol{\mu}}_k^t}\mathbb{E}_{q_{\Bar{\theta}_i^t}}[f_i(\Bar{\boldsymbol{x}}_i)] =   {\boldsymbol{\Sigma}^t_k}^{-1} \mathbb{E}_{q_{\Bar{\theta}_i^t}}[ (\boldsymbol{x}_k - \boldsymbol{\mu}_k) f_i(\Bar{\boldsymbol{x}}_i)]  \\
& \nabla_{{\boldsymbol{\Sigma}}_k^t}\mathbb{E}_{q_{\Bar{\theta}_i^t}}[f_i(\Bar{\boldsymbol{x}}_i )] = \frac{1}{2} \mathbb{E}_ { q_{\Bar{\theta}_i^t} } [ \Big(  {\boldsymbol{\Sigma}^t_k}^{-1}  \big( \boldsymbol{x}_k - \boldsymbol{\mu}_k \big) \big( \boldsymbol{x}_k - \boldsymbol{\mu}_k \big)^\top  {\boldsymbol{\Sigma}^t_k}^{-1}  -   {\boldsymbol{\Sigma}^t_k}^{-1}   \Big)  ( f_i(\Bar{\boldsymbol{x}}_i ) ) ]
\end{align}

Then, we have that
\begin{align}
& \boldsymbol{\mu}_k^{t+1} = \boldsymbol{\mu}_k^t -     \sum_{i=k}^K \beta  \mathbb{E}_{q_{\Bar{\theta}_i^t}}[ (\boldsymbol{x}_k - \boldsymbol{\mu}_k) f_i(\Bar{\boldsymbol{x}}_i)]  \\
    &{\boldsymbol{\Sigma}_k^{t\!+\!1}}^{-\!1} =  {\boldsymbol{\Sigma}^t_k}^{-\!1} +  \sum_{i=k}^K \beta   \mathbb{E}_ { q_{\Bar{\theta}_i^t} } [ \Big(  {\boldsymbol{\Sigma}^t_k}^{-1}  \big( \boldsymbol{x}_k - \boldsymbol{\mu}_k \big) \big( \boldsymbol{x}_k - \boldsymbol{\mu}_k \big)^\top  {\boldsymbol{\Sigma}^t_k}^{-1}  -   {\boldsymbol{\Sigma}^t_k}^{-1}   \Big)  ( f_i(\Bar{\boldsymbol{x}}_i ) ) ]   .
\end{align}


\section{Technical Lemmas}

In this section, we introduce the following technical lemmas for convergence analysis.

\begin{lemma}\label{lemma:SigmaNormIeq}
    Given  a positive definite matrix $\boldsymbol{\Sigma}$, we have $\|\boldsymbol{\Sigma}(\boldsymbol{x}+\boldsymbol{y}) \|_{\boldsymbol{\Sigma}^{-1}}^2 \le 2(\|\boldsymbol{\Sigma}\boldsymbol{x} \|_{\boldsymbol{\Sigma}^{-1}}^2 + \| {\boldsymbol{\Sigma}^{\frac{1}{2}}} \boldsymbol{y}  \|_2^2)$
\end{lemma}
\begin{proof}
    \begin{align}
        \|\boldsymbol{\Sigma}(\boldsymbol{x}+\boldsymbol{y}) \|_{\boldsymbol{\Sigma}^{-1}}^2 & = \left< {\boldsymbol{\Sigma}^{-1}} \boldsymbol{\Sigma}(\boldsymbol{x}+\boldsymbol{y} ) , \boldsymbol{\Sigma}(\boldsymbol{x}+\boldsymbol{y} ) \right> \\
        & =\left< (\boldsymbol{x}+\boldsymbol{y} ) , \boldsymbol{\Sigma}(\boldsymbol{x}+\boldsymbol{y} ) \right> \\
        & = \|{\boldsymbol{\Sigma}^{\frac{1}{2}}}\boldsymbol{x} + {\boldsymbol{\Sigma}^{\frac{1}{2}}} \boldsymbol{y}  \|_2^2\\
        & \le 2 (\|{\boldsymbol{\Sigma}^{\frac{1}{2}}}\boldsymbol{x}  \|_2^2+ \|{\boldsymbol{\Sigma}^{\frac{1}{2}}}\boldsymbol{y}  \|_2^2)
    \end{align}
Note that $\|{\boldsymbol{\Sigma}^{\frac{1}{2}}}\boldsymbol{x}  \|_2^2 = \big< \boldsymbol{x},\boldsymbol{\Sigma}\boldsymbol{x}\big> =  \big< {\boldsymbol{\Sigma}^{-1}}\boldsymbol{\Sigma}\boldsymbol{x},\boldsymbol{\Sigma}\boldsymbol{x}\big> = \|\boldsymbol{\Sigma}\boldsymbol{x} \|_{\boldsymbol{\Sigma}^{-1}}^2 $, we achieve that 
\begin{align}
    \|\boldsymbol{\Sigma}(\boldsymbol{x}+\boldsymbol{y}) \|_{\boldsymbol{\Sigma}^{-1}}^2 \le 2(\|\boldsymbol{\Sigma}\boldsymbol{x} \|_{\boldsymbol{\Sigma}^{-1}}^2 + \| {\boldsymbol{\Sigma}^{\frac{1}{2}}} \boldsymbol{y}  \|_2^2)
\end{align}
    
\end{proof}


\begin{lemma}\label{lemma:estimator}
  Suppose the gradient estimator  $\hat{{g}}_{ik}^t$ for the $i^{th}$ objective w.r.t. the $k^{th}$ component ${\boldsymbol{\mu}_k}$  at $t^{th}$ iteration as 
    \begin{align}\label{OneEstimator}
       \hat{{g}}_{ik}^t = {\boldsymbol{\Sigma}_k^t}^{-\frac{1}{2}}\boldsymbol{z}_k\big(f_i(\Bar{\boldsymbol{\mu}}_i^t +{\Bar{\boldsymbol{\Sigma}}_i}^{t\frac{1}{2}}\Bar{\boldsymbol{z}_i})-f_i(\Bar{\boldsymbol{\mu}}_i^t)\big),
    \end{align}
    where $\boldsymbol{z}_1,\cdots,\boldsymbol{z}_i\sim\mathcal{N}(\boldsymbol{0},\boldsymbol{I}_{d})$ and $\Bar{\boldsymbol{z}_i}=[\boldsymbol{z}_1^\top,\cdots,\boldsymbol{z}_i^\top]^\top$, $i\ge k$. Suppose assumptions \ref{Assumption2} hold, using the parameter setting in Theorem~\ref{MainTheorem}, and the  covariance matrix update  $\hat{G}_k^t= {\boldsymbol{\Sigma}_k^t}^{-\frac{1}{2}} \boldsymbol{H}_k^t {\boldsymbol{\Sigma}_k^t}^{-\frac{1}{2}} $ are positive semi-definite matrix that satisfies  $\nu \boldsymbol{I} \preceq \hat{G}_k^t$. Apply the update rule  $ {\boldsymbol{\Sigma}_k^{t\!+\!1}}^{-\!1} = \omega_t {\boldsymbol{\Sigma}_k^{t}}^{-\!1} +  \alpha_t \hat{G}_k^t $,  we have
\begin{enumerate}[label=(\alph*)]
\item $\hat{{g}}_{ik}^t$ is an unbiased estimator of $g_{ik}^t =\nabla_{\boldsymbol{\mu}_k}\mathbb{E}_{\Bar{\boldsymbol{x}}\sim \mathcal{N}(\Bar{\boldsymbol{\mu}}_i^t,{\Bar{\boldsymbol{\Sigma}}_i^t})}[f_i(\Bar{\boldsymbol{x}})]$.
\item  $\| \boldsymbol{\Sigma}_{k}^{t+1} \|_2 \le \frac{1}{ \|{\boldsymbol{\Sigma}_{k}^{t}} \|_2^{-1}  +  \sqrt{t+1}  \alpha  \nu  }  \le \cdots \le  \frac{3}{2\alpha\nu} \frac{1}{(t+1)^\frac{3}{2} + \frac{3}{2}} \le \frac{3}{2\alpha\nu} \frac{1}{(t+1)^\frac{3}{2} }   $.
\item 
$\mathbb{E} \sum_{k=1}^K \| \boldsymbol{\Sigma}_{k}^{t}(\sum_{i=k}^K \hat{g}_{ik} )   \|_{ {\boldsymbol{\Sigma}_{k}^{t}}^{\!-\!1} }^2  \le  \frac{ 3 \sum_{i=1}^K K L_i^2  (id+1)^2}{ 2 t^{\frac{3}{2}}  \alpha \nu} $
\end{enumerate}
For the average of i.i.d. sampled unbiased gradient estimators (each one has the same form as Eq.(\ref{OneEstimator})), the results (a),(b),(c) still hold true. 
\end{lemma}

\begin{proof}
    (a). We first show that $\hat{{g}}_{ik}^t$ is an unbiased estimator of $\nabla_{\boldsymbol{\mu}_k}\mathbb{E}_{\Bar{\boldsymbol{x}}\sim \mathcal{N}(\Bar{\boldsymbol{\mu}}_i^t,{\Bar{\boldsymbol{\Sigma}}_i^t})}[f_i(\Bar{\boldsymbol{x}})]$. 
\begin{align}
    \mathbb{E}_{\Bar{\boldsymbol{z}}_i}[\hat{{g}}_{ik}^t] &=\mathbb{E}_{\Bar{\boldsymbol{z}}_i}[ {\boldsymbol{\Sigma}_k^t}^{-\frac{1}{2}}\boldsymbol{z}_k f_i(\Bar{\boldsymbol{\mu}}_i^t +{\Bar{\boldsymbol{\Sigma}}_i}^{t\frac{1}{2}}\Bar{\boldsymbol{z}_i}) ]-\mathbb{E}_{\Bar{\boldsymbol{z}}_i}[ {\boldsymbol{\Sigma}_k^t}^{-\frac{1}{2}}\boldsymbol{z}_k f_i(\Bar{\boldsymbol{\mu}}_i^t ) ]\\
    & = \mathbb{E}_{\Bar{\boldsymbol{z}}_i}[ {\boldsymbol{\Sigma}_k^t}^{-\frac{1}{2}}\boldsymbol{z}_k f_i(\Bar{\boldsymbol{\mu}}_i^t +{\Bar{\boldsymbol{\Sigma}}_i}^{t\frac{1}{2}}\Bar{\boldsymbol{z}_i}) ]\\
    & = \mathbb{E}_{\Bar{\boldsymbol{x}}\sim \mathcal{N}(\Bar{\boldsymbol{\mu}}_i^t,{\Bar{\boldsymbol{\Sigma}}_i^t})} [  {\boldsymbol{\Sigma}_k^t}^{-1} (\boldsymbol{x}_k - \boldsymbol{\mu}_k^t)   f_i(\Bar{\boldsymbol{x}})] 
\end{align}
Note that  $ \Bar{\boldsymbol{\Sigma}}_i^t = diag(\boldsymbol{\Sigma}_1^t,\cdots,\boldsymbol{\Sigma}_i^t) $ is a block-wise diagonal matrix, and $\Bar{\boldsymbol{\mu}}_i = [\boldsymbol{\mu}_1^\top,\cdots,\boldsymbol{\mu}_i^\top]^\top$, $i \ge k$ , we then have that 
\begin{align}
     \mathbb{E}_{\Bar{\boldsymbol{z}}_i}[\hat{{g}}_{ik}^t] = \mathbb{E}_{\Bar{\boldsymbol{x}}\sim \mathcal{N}(\Bar{\boldsymbol{\mu}}_i^t,{\Bar{\boldsymbol{\Sigma}}_i^t})} [  {\boldsymbol{\Sigma}_k^t}^{-1} (\boldsymbol{x}_k - \boldsymbol{\mu}_k^t)   f_i(\Bar{\boldsymbol{x}})] = \nabla_{\boldsymbol{\mu}_k}\mathbb{E}_{\Bar{\boldsymbol{x}}\sim \mathcal{N}(\Bar{\boldsymbol{\mu}}_i^t,{\Bar{\boldsymbol{\Sigma}}_i^t})}[f_i(\Bar{\boldsymbol{x}})]. 
\end{align}

For $N$ i.i.d. sampled unbiased gradient estimator, the average is still  an unbiased gradient estimator. 
\end{proof}

\begin{proof}
  (b)  We now prove the decay of the spectral norm of covariance matrix. 

From the update rule of $\boldsymbol{\Sigma}_{k}^{t}$, we know that 
\begin{align}
    {\boldsymbol{\Sigma}_{k}^{t\!+\!1}}^{-\!1} &  =  \omega_t {\boldsymbol{\Sigma}_{k}^{t}}^{-\!1} + \alpha_t  \hat{G}_k^{t} 
\end{align}


 Note that $\nu \boldsymbol{I}  \preceq 
 \hat{G}_k^t$, we have that
\begin{align}
 \lambda_{\text{min}} \big(  {\boldsymbol{\Sigma}_{k}^{t\!+\!1}}^{-\!1} \big) & =    \lambda_{\text{min}} \big( \omega_t {\boldsymbol{\Sigma}_{k}^{t}}^{-\!1} + \alpha_t  \hat{G}_k^{t} \big)  \\
 & \ge  \omega_t   \lambda_{\text{min}} \big(  {\boldsymbol{\Sigma}_{k}^{t}}^{-\!1} \big)  +   \alpha_t \nu  
\end{align}
Note that $\|{\boldsymbol{\Sigma}_{k}^{t\!+\!1}} \|_2 = \frac{1}{ \lambda_{\text{min}} \big(  {\boldsymbol{\Sigma}_{k}^{t\!+\!1}}^{-\!1} \big)}$ and $\alpha_t = \sqrt{t+1} \alpha$, $\omega_t = 1$
, we have that 
\begin{align}
    \|{\boldsymbol{\Sigma}_{k}^{t\!+\!1}} \|_2 \le \frac{1}{ \omega_t \lambda_{\text{min}} \big(  {\boldsymbol{\Sigma}_{k}^{t}}^{-\!1} \big)  +   \alpha_t \nu  } = \frac{1}{   \|{\boldsymbol{\Sigma}_{k}^{t}} \|_2^{-1}  +  \sqrt{t+1} \alpha \nu  } 
\end{align}

It follows that 
\begin{align}
    \lambda_{\text{min}} \big(  {\boldsymbol{\Sigma}_{k}^{t\!+\!1}}^{-\!1} \big) & \ge   \lambda_{\text{min}} \big(  {\boldsymbol{\Sigma}_{k}^{t}}^{-\!1} \big)  + \sqrt{t+1} \alpha \nu \\ & \ge  \lambda_{\text{min}} \big(  {\boldsymbol{\Sigma}_{k}^{t\!-\!1}}^{-\!1} \big)  + \sqrt{t} \alpha \nu  + \sqrt{t+1} \alpha \nu   \\
    & \ge \lambda_{\text{min}} \big(  {\boldsymbol{\Sigma}_{k}^{1}}^{-\!1} \big) +  (\sum_{i=1}^t \sqrt{i+1}) \alpha \nu     \\
    & \ge \lambda_{\text{min}} \big(  {\boldsymbol{\Sigma}_{k}^{1}}^{-\!1} \big) + \frac{2\alpha\nu}{3} ((t+1)^{\frac{3}{2}} -1)
\end{align}

Note that the  initialization such that $\lambda_{\text{min}} \big(  {\boldsymbol{\Sigma}_{k}^{1}}^{-\!1} \big) = \|{\boldsymbol{\Sigma}_{k}^{1}} \|_2^{-1} \ge \frac{5}{3}\alpha \nu  $, we have that 
\begin{align}
     \lambda_{\text{min}} \big(  {\boldsymbol{\Sigma}_{k}^{t\!+\!1}}^{-\!1} \big) \ge \frac{2\alpha\nu}{3} (t+1)^{\frac{3}{2}} +  \alpha \nu = \frac{2\alpha\nu}{3}  \big((t+1)^{\frac{3}{2}} + \frac{3}{2}  \big) 
\end{align}

 Note that $\|{\boldsymbol{\Sigma}_{k}^{t\!+\!1}} \|_2 = \frac{1}{ \lambda_{\text{min}} \big(  {\boldsymbol{\Sigma}_{k}^{t\!+\!1}}^{-\!1} \big)}$
 , we then  have  that
\begin{align}
     \| \boldsymbol{\Sigma}_{k}^{t+1} \|_2 \le  \frac{3}{2\alpha\nu} \frac{1}{(t+1)^\frac{3}{2} + \frac{3}{2}}
\end{align}



\end{proof}

\begin{proof}
    (c).  We now prove the upper bound of $\mathbb{E} \sum_{k=1}^K \| \boldsymbol{\Sigma}_{k}^{t}(\sum_{i=k}^K \hat{g}_{ik} )   \|_{ {\boldsymbol{\Sigma}_{k}^{t}}^{\!-\!1} }^2   $.

Note that $ \hat{{g}}_{ik}^t = {\boldsymbol{\Sigma}_k^t}^{-\frac{1}{2}}\boldsymbol{z}_k\big(f_i(\Bar{\boldsymbol{\mu}}_i^t +{\Bar{\boldsymbol{\Sigma}}_i}^{t\frac{1}{2}}\Bar{\boldsymbol{z}_i})-f_i(\Bar{\boldsymbol{\mu}}_i^t)\big)$, we have that 
\begin{align}
    \| \boldsymbol{\Sigma}_{k}^{t}(\sum_{i=k}^K \hat{g}_{ik} )   \|_{ {\boldsymbol{\Sigma}_{k}^{t}}^{\!-\!1} }^2 & = \| \boldsymbol{\Sigma}_{k}^{t} {\boldsymbol{\Sigma}_k^t}^{-\frac{1}{2}}\boldsymbol{z}_k  (\sum_{i=k}^K\big(f_i(\Bar{\boldsymbol{\mu}}_i^t +{\Bar{\boldsymbol{\Sigma}}_i}^{t\frac{1}{2}}\Bar{\boldsymbol{z}_i})-f_i(\Bar{\boldsymbol{\mu}}_i^t)\big) )   \|_{ {\boldsymbol{\Sigma}_{k}^{t}}^{\!-\!1} }^2 \\ 
    & = \| {\boldsymbol{\Sigma}_k^t}^{\frac{1}{2}}\boldsymbol{z}_k  (\sum_{i=k}^K\big(f_i(\Bar{\boldsymbol{\mu}}_i^t +{\Bar{\boldsymbol{\Sigma}}_i}^{t\frac{1}{2}}\Bar{\boldsymbol{z}_i})-f_i(\Bar{\boldsymbol{\mu}}_i^t)\big) )   \|_{ {\boldsymbol{\Sigma}_{k}^{t}}^{\!-\!1} }^2 \\
    & = \| \boldsymbol{z}_k  (\sum_{i=k}^K\big(f_i(\Bar{\boldsymbol{\mu}}_i^t +{\Bar{\boldsymbol{\Sigma}}_i}^{t\frac{1}{2}}\Bar{\boldsymbol{z}_i})-f_i(\Bar{\boldsymbol{\mu}}_i^t)\big) )   \|_2^2 \\
    & = \big( \sum_{i=k}^K\big(f_i(\Bar{\boldsymbol{\mu}}_i^t +{\Bar{\boldsymbol{\Sigma}}_i}^{t\frac{1}{2}}\Bar{\boldsymbol{z}_i})-f_i(\Bar{\boldsymbol{\mu}}_i^t)\big) \big)^2 \|\boldsymbol{z}_k \|_2^2 \label{SigmaOne}
\end{align}

Note that $f_i(\Bar{x})$ is $L_i$-Lipschitz continuous function, we then have that
\begin{align}
    \big( \sum_{i=k}^K\big(f_i(\Bar{\boldsymbol{\mu}}_i^t +{\Bar{\boldsymbol{\Sigma}}_i}^{t\frac{1}{2}}\Bar{\boldsymbol{z}_i})-f_i(\Bar{\boldsymbol{\mu}}_i^t)\big) \big)^2 & \le (K-k+1) (\sum_{i=k}^K\big(f_i(\Bar{\boldsymbol{\mu}}_i^t +{\Bar{\boldsymbol{\Sigma}}_i}^{t\frac{1}{2}}\Bar{\boldsymbol{z}_i})-f_i(\Bar{\boldsymbol{\mu}}_i^t)\big)^2) \\
    & \le (K\!-\!k\!+\!1) (\sum_{i=k}^K L_i^2 \| \Bar{\boldsymbol{\mu}}_i^t +{\Bar{\boldsymbol{\Sigma}}_i}^{t\frac{1}{2}}\Bar{\boldsymbol{z}_i} - \Bar{\boldsymbol{\mu}}_i^t \|_2^2 )  \\
    & \le (K\!-\!k\!+\!1) (\sum_{i=k}^K L_i^2 \|{\Bar{\boldsymbol{\Sigma}}_i}^{t\frac{1}{2}} \|_2^2 \|\Bar{\boldsymbol{z}_i} \|_2^2 ) \\
    & = (K\!-\!k\!+\!1) (\sum_{i=k}^K L_i^2 \|{\Bar{\boldsymbol{\Sigma}}_i}^t \|_2 \|\Bar{\boldsymbol{z}_i} \|_2^2 ) \label{SingleSigmaIeq}
\end{align}
Plug Eq.(\ref{SingleSigmaIeq}) into Eq.(\ref{SigmaOne}), we have that 
\begin{align}
     \| \boldsymbol{\Sigma}_{k}^{t}(\sum_{i=k}^K \hat{g}_{ik} )   \|_{ {\boldsymbol{\Sigma}_{k}^{t}}^{\!-\!1} }^2 \le (K\!-\!k\!+\!1)  \|\boldsymbol{z}_k \|_2^2 (\sum_{i=k}^K L_i^2 \|{\Bar{\boldsymbol{\Sigma}}_i^t} \|_2 \|\Bar{\boldsymbol{z}_i} \|_2^2 )
\end{align}

It follows that
\begin{align}
    \sum_{k=1}^K  \| \boldsymbol{\Sigma}_{k}^{t}(\sum_{i=k}^K \hat{g}_{ik} )   \|_{ {\boldsymbol{\Sigma}_{k}^{t}}^{\!-\!1} }^2 & \le \sum_{k=1}^K (K\!-\!k\!+\!1)  \|\boldsymbol{z}_k \|_2^2 (\sum_{i=k}^K L_i^2 \|{\Bar{\boldsymbol{\Sigma}}_i^t} \|_2 \|\Bar{\boldsymbol{z}_i} \|_2^2 ) \\
    & = \sum_{i=1}^K \sum_{k=1}^i \left( (K\!-\!k\!+\!1)  \|\boldsymbol{z}_k \|_2^2 L_i^2 \|{\Bar{\boldsymbol{\Sigma}}_i^t} \|_2 \|\Bar{\boldsymbol{z}_i} \|_2^2 \right) \\
    & = \sum_{i=1}^K  L_i^2 \|{\Bar{\boldsymbol{\Sigma}}_i^t} \|_2 \|\Bar{\boldsymbol{z}_i} \|_2^2  \Big( \sum_{k=1}^i (K\!-\!k\!+\!1)  \|\boldsymbol{z}_k \|_2^2  \Big) \\
    & \le \sum_{i=1}^K  L_i^2 \|{\Bar{\boldsymbol{\Sigma}}_i^t} \|_2 \|\Bar{\boldsymbol{z}_i} \|_2^2 K \| \Bar{\boldsymbol{z}_i} \|_2^2 \\
    & = \sum_{i=1}^K K L_i^2 \|{\Bar{\boldsymbol{\Sigma}}_i^t}  \|_2 \|\Bar{\boldsymbol{z}_i} \|_2^4 
\end{align}

In addition, note that for $z \sim \mathcal{N}(0,\sigma^2)$, we have $\mathbb{E} [z^4] = 3\sigma^4$. It follows that 
\begin{align}
    \mathbb{E} \|\Bar{\boldsymbol{z}_i} \|_2^4 = \sum_{j=1}^{id} \mathbb{E}[z_j^4] + \sum_{j_1=1}^{id} \sum_{j_2 \ne j_1}^{id} \mathbb{E}[ z^2_{j_1}z^2_{j_2}] = 3id + id(id-1) = i^2d^2+2id
\end{align}

We then have that
\begin{align}
 \mathbb{E}   \sum_{k=1}^K  \| \boldsymbol{\Sigma}_{k}^{t}(\sum_{i=k}^K \hat{g}_{ik} )   \|_{ {\boldsymbol{\Sigma}_{k}^{t}}^{\!-\!1} }^2  & \le \sum_{i=1}^K K L_i^2 \|{\Bar{\boldsymbol{\Sigma}}_i^t}  \|_2  \mathbb{E} \|\Bar{\boldsymbol{z}_i} \|_2^4 \\
    & \le \sum_{i=1}^K K L_i^2 \|{\Bar{\boldsymbol{\Sigma}}_i^t}  \|_2  (id+1)^2
\end{align}

Note that ${\Bar{\boldsymbol{\Sigma}}_i^t}=diag(\boldsymbol{\Sigma}_1^t,\cdots,\boldsymbol{\Sigma}_i^t)$ is a block-wise diagonal matrix, we have $\|{\Bar{\boldsymbol{\Sigma}}_i^t} \|_2 \le \max _{k \in \{1,\cdots,i\}} \| \boldsymbol{\Sigma}_{k}^{t} \|_2 $. Then we know that
\begin{align}
     \mathbb{E}   \sum_{k=1}^K  \| \boldsymbol{\Sigma}_{k}^{t}(\sum_{i=k}^K \hat{g}_{ik} )   \|_{ {\boldsymbol{\Sigma}_{k}^{t}}^{\!-\!1} }^2  & \le  \sum_{i=1}^K K L_i^2 \|{\Bar{\boldsymbol{\Sigma}}_i^t}  \|_2  (id+1)^2 \\
     & \le \max _{k \in \{1,\cdots,K\}} \| \boldsymbol{\Sigma}_{k}^{t} \|_2 \sum_{i=1}^K K L_i^2  (id+1)^2
\end{align}

From Lemma~\ref{lemma:estimator} (b), we know that  $\| \boldsymbol{\Sigma}_{k}^{t} \|_2 \le \frac{3}{2\alpha\nu} \frac{1}{ t^{\frac{3}{2}} }  $ for $\forall k \in \{1,\cdots,K\}$.    Then, we have 
\begin{align}
     \mathbb{E}   \sum_{k=1}^K  \| \boldsymbol{\Sigma}_{k}^{t}(\sum_{i=k}^K \hat{g}_{ik} )   \|_{ {\boldsymbol{\Sigma}_{k}^{t}}^{\!-\!1} }^2  & \le \max _{k \in \{1,\cdots,K\}} \| \boldsymbol{\Sigma}_{k}^{t} \|_2 \sum_{i=1}^K K L_i^2  (id+1)^2 \\
     & \le   \frac{ 3 \sum_{i=1}^K K L_i^2  (id+1)^2}{ 2 t^{\frac{3}{2}} \alpha \nu}
\end{align}

Note that the square norm $\|\cdot \|_{ {\boldsymbol{\Sigma}_{k}^{t}}^{\!-\!1} }^2$ is a convex function, then for the average of $N$ i.i.d. sampled gradient estimator $\hat{g}_{ik}^j, j \in \{1,\cdots,N\}$,  we have 
\begin{align}
     \mathbb{E}   \sum_{k=1}^K  \| \boldsymbol{\Sigma}_{k}^{t}(\sum_{i=k}^K \frac{1}{N}\sum_{j=1}^N \hat{g}_{ik}^j )   \|_{ {\boldsymbol{\Sigma}_{k}^{t}}^{\!-\!1} }^2 & \le \frac{1}{N }\sum_{j=1}^N \mathbb{E}   \sum_{k=1}^K  \| \boldsymbol{\Sigma}_{k}^{t}(\sum_{i=k}^K \hat{g}_{ik}^j )   \|_{ {\boldsymbol{\Sigma}_{k}^{t}}^{\!-\!1} }^2 \\
     & = \frac{N}{N} \mathbb{E}   \sum_{k=1}^K  \| \boldsymbol{\Sigma}_{k}^{t}(\sum_{i=k}^K \hat{g}_{ik} )   \|_{ {\boldsymbol{\Sigma}_{k}^{t}}^{\!-\!1} }^2 \\
      & \le   \frac{ 3 \sum_{i=1}^K K L_i^2  (id+1)^2}{ 2 t^{\frac{3}{2}} \alpha \nu}
\end{align}
    
\end{proof}

\begin{lemma}\label{lemma:trGsigma}
 Denote $ G_i^t    = \nabla_{\Bar{\boldsymbol{\Sigma}}_i = \Bar{\boldsymbol{\Sigma}}_i^{t} }  J_i(\Bar{\boldsymbol{\mu}}_i^t,\Bar{\boldsymbol{\Sigma}}_i^t)$.    Suppose assumption~\ref{Assumption2} holds,  using the parameter setting in Theorem~\ref{MainTheorem}, and the  covariance matrix update  $\hat{G}_k^t= {\boldsymbol{\Sigma}_k^t}^{-\frac{1}{2}} \boldsymbol{H}_k^t {\boldsymbol{\Sigma}_k^t}^{-\frac{1}{2}} $ are positive semi-definite matrix that satisfies  $\nu \boldsymbol{I} \preceq \hat{G}_k^t$. Apply the update rule  $ {\boldsymbol{\Sigma}_k^{t\!+\!1}}^{-\!1} = \omega_t {\boldsymbol{\Sigma}_k^{t}}^{-\!1} +  \alpha_t \hat{G}_k^t $,   for $k \in \{1,\cdots,K\}$.    Then we have
    \begin{align}
   \text{tr}(G_i^t\Bar{\boldsymbol{\Sigma}}_i^t) \le   | \text{tr}(G_i^t\Bar{\boldsymbol{\Sigma}}_i^t) | \le \frac{L_iid}{2t^{\frac{3}{4}}} \sqrt{\frac{3}{\alpha \nu}}
\end{align}
\end{lemma}

\begin{proof}
    \begin{align}
        \text{tr}(G_i^t\Bar{\boldsymbol{\Sigma}}_i^t) & = \text{tr}(\Bar{\boldsymbol{\Sigma}}_i^{t\frac{1}{2}}G_i^t \Bar{\boldsymbol{\Sigma}}_i^{t\frac{1}{2}}) \\
        & =  \frac{1}{2}\text{tr}\left( \Bar{\boldsymbol{\Sigma}}_i^{t\frac{1}{2}}\mathbb{E}_{\mathcal{N}(\Bar{\boldsymbol{\mu}}_i^t, \Bar{\boldsymbol{\Sigma}}_i^t)} \big[ ( {\Bar{\boldsymbol{\Sigma}_i^t} }^{ -\!1} (\Bar{\boldsymbol{x}_i} - \Bar{\boldsymbol{\mu}}_i^t)(\Bar{\boldsymbol{x}_i} - \Bar{\boldsymbol{\mu}}_i^t)^\top  {\Bar{\boldsymbol{\Sigma}_i^t} }^{ -\!1} - {\Bar{\boldsymbol{\Sigma}_i^t} }^{ -\!1} ) f_i(\Bar{\boldsymbol{x}_i}) \big]\Bar{\boldsymbol{\Sigma}}_i^{t\frac{1}{2}}\right) \\
        & = \frac{1}{2}\text{tr}\left( \mathbb{E}_{\mathcal{N}(\Bar{\boldsymbol{\mu}}_i^t, \Bar{\boldsymbol{\Sigma}}_i^t)} \big[ ( {\Bar{\boldsymbol{\Sigma}_i^t} }^{ -\!\frac{1}{2}} (\Bar{\boldsymbol{x}_i} - \Bar{\boldsymbol{\mu}}_i^t)(\Bar{\boldsymbol{x}_i} - \Bar{\boldsymbol{\mu}}_i^t)^\top  {\Bar{\boldsymbol{\Sigma}_i^t} }^{ -\!\frac{1}{2}} - \boldsymbol{I} ) f_i(\Bar{\boldsymbol{x}_i}) \big]\right) \\
        & = \frac{1}{2}\text{tr}\left(  \mathbb{E}_{ \Bar{\boldsymbol{z}} \sim \mathcal{N}(\boldsymbol{0}, \boldsymbol{I})} \big[ ( \Bar{\boldsymbol{z}}\Bar{\boldsymbol{z}}^\top - \boldsymbol{I} ) f_i( \Bar{\boldsymbol{\mu}}_i^t +    \Bar{\boldsymbol{\Sigma}}_i^{t\frac{1}{2}}\Bar{\boldsymbol{z}}) \big]\right) \\
        & = \frac{1}{2}\text{tr}\left(  \mathbb{E}_{ \Bar{\boldsymbol{z}} \sim \mathcal{N}(\boldsymbol{0}, \boldsymbol{I})} \big[ ( \Bar{\boldsymbol{z}}\Bar{\boldsymbol{z}}^\top - \boldsymbol{I} ) (f_i( \Bar{\boldsymbol{\mu}}_i^t +    \Bar{\boldsymbol{\Sigma}}_i^{t\frac{1}{2}}\Bar{\boldsymbol{z}}) - f_i( \Bar{\boldsymbol{\mu}}_i^t)  )\big]\right)  \\
        & = \frac{1}{2} \mathbb{E}_{ \Bar{\boldsymbol{z}} \sim \mathcal{N}(\boldsymbol{0}, \boldsymbol{I})} \big[ (\sum_{j=1}^{id} (z_j^2-1)) (f_i( \Bar{\boldsymbol{\mu}}_i^t +    \Bar{\boldsymbol{\Sigma}}_i^{t\frac{1}{2}}\Bar{\boldsymbol{z}}) - f_i( \Bar{\boldsymbol{\mu}}_i^t)  )   \big] 
    \end{align}
    where $z_j$ denotes the $j^{th}$ element in $\Bar{\boldsymbol{z}}$. 

    From Cauchy–Schwarz inequality $|\mathbb{E}[XY]| \le \sqrt{\mathbb{E}[X^2]\mathbb{E}[Y^2]} $, we know that 
    \begin{align}
        | \text{tr}(G_i^t\Bar{\boldsymbol{\Sigma}}_i^t) |& = \frac{1}{2} | \mathbb{E}_{ \Bar{\boldsymbol{z}} \sim \mathcal{N}(\boldsymbol{0}, \boldsymbol{I})} \big[ (\sum_{j=1}^{id} (z_j^2-1)) (f_i( \Bar{\boldsymbol{\mu}}_i^t +    \Bar{\boldsymbol{\Sigma}}_i^{t\frac{1}{2}}\Bar{\boldsymbol{z}}) - f_i( \Bar{\boldsymbol{\mu}}_i^t)  )   \big] |  \\
       & \le \frac{1}{2} \sqrt{ \mathbb{E}_{ \Bar{\boldsymbol{z}} \sim \mathcal{N}(\boldsymbol{0}, \boldsymbol{I})} \big[ (\sum_{j=1}^{id} (z_j^2-1))^2 \big] \mathbb{E}_{ \Bar{\boldsymbol{z}} \sim \mathcal{N}(\boldsymbol{0}, \boldsymbol{I})} \big[ (f_i( \Bar{\boldsymbol{\mu}}_i^t +    \Bar{\boldsymbol{\Sigma}}_i^{t\frac{1}{2}}\Bar{\boldsymbol{z}}) - f_i( \Bar{\boldsymbol{\mu}}_i^t)  )^2 \big]   } \label{trBound}
    \end{align}

We first check the term $\mathbb{E}_{ \Bar{\boldsymbol{z}} \sim \mathcal{N}(\boldsymbol{0}, \boldsymbol{I})} \big[ (\sum_{j=1}^{id} (z_j^2-1))^2 \big]$.
\begin{align}
    \mathbb{E}_{ \Bar{\boldsymbol{z}} \sim \mathcal{N}(\boldsymbol{0}, \boldsymbol{I})} \big[ (\sum_{j=1}^{id} (z_j^2-1))^2 \big] & = \sum_{j=1}^{id} \mathbb{E}(z_j^2-1)^2 + \sum_{j_1=1}^{id}\sum_{j_2 \ne j_1}^{id} \mathbb{E}(z_{j_1}^2-1)(z_{j_2}^2-1) \\
    & = \sum_{j=1}^{id} \mathbb{E}(z_j^4-2z_j^2 +1) + \sum_{j_1=1}^{id}\sum_{j_2 \ne j_1}^{id} \mathbb{E}(z_{j_1}^2-1)\mathbb{E}(z_{j_2}^2-1) \\
    & = \sum_{j=1}^{id} \mathbb{E}(z_j^4-2z_j^2 +1)  = \sum_{j=1}^{id} [3 - 2 + 1] = 2id \label{Sqrt1}
\end{align}

We now check the term $\mathbb{E}_{ \Bar{\boldsymbol{z}} \sim \mathcal{N}(\boldsymbol{0}, \boldsymbol{I})} \big[ (f_i( \Bar{\boldsymbol{\mu}}_i^t +    \Bar{\boldsymbol{\Sigma}}_i^{t\frac{1}{2}}\Bar{\boldsymbol{z}}) - f_i( \Bar{\boldsymbol{\mu}}_i^t)  )^2  \big]$. Note that $f_i(\boldsymbol{x})$ is  $L_i$-Lipschitz continuous function, we then have that
\begin{align}
    \mathbb{E}_{ \Bar{\boldsymbol{z}} \sim \mathcal{N}(\boldsymbol{0}, \boldsymbol{I})} \big[ (f_i( \Bar{\boldsymbol{\mu}}_i^t +    \Bar{\boldsymbol{\Sigma}}_i^{t\frac{1}{2}}\Bar{\boldsymbol{z}}) - f_i( \Bar{\boldsymbol{\mu}}_i^t)  )^2  \big] & \le L_i^2   \mathbb{E}_{ \Bar{\boldsymbol{z}} \sim \mathcal{N}(\boldsymbol{0}, \boldsymbol{I})} \big[  \| \Bar{\boldsymbol{\Sigma}}_i^{t\frac{1}{2}}\Bar{\boldsymbol{z}} \|_2^2 \big] \\
    & \le L_i^2  \| \Bar{\boldsymbol{\Sigma}}_i^{t\frac{1}{2}} \|_2^2  \mathbb{E}_{ \Bar{\boldsymbol{z}} \sim \mathcal{N}(\boldsymbol{0}, \boldsymbol{I})}  \| \Bar{\boldsymbol{z}} \|_2^2 \\
    & =  L_i^2\| \Bar{\boldsymbol{\Sigma}}_i^{t} \|_2 id  \label{DeltaFpre1}
\end{align}
 From Lemma~\ref{lemma:estimator} (b)   we know that 
\begin{align}
    \| \Bar{\boldsymbol{\Sigma}}_i^{t} \|_2 & = \max_{k \in \{1,\cdots, i\}} \| {\boldsymbol{\Sigma}}_k^{t} \|_2  
     \le \frac{3}{2\alpha\nu} \frac{1}{t^\frac{3}{2} }  \label{DeltaFpre2}
\end{align}
Together with Eq.(\ref{DeltaFpre1}) and Eq.(\ref{DeltaFpre2} ), we know that 
\begin{align}
    \mathbb{E}_{ \Bar{\boldsymbol{z}} \sim \mathcal{N}(\boldsymbol{0}, \boldsymbol{I})} \big[ (f_i( \Bar{\boldsymbol{\mu}}_i^t +    \Bar{\boldsymbol{\Sigma}}_i^{t\frac{1}{2}}\Bar{\boldsymbol{z}}) - f_i( \Bar{\boldsymbol{\mu}}_i^t)  )^2  \big] \le    \frac{3L_i^2 id }{  2 t^{\frac{3}{2}} \alpha \nu }   \label{SigmaPre1}
\end{align}
Plug Eq.(\ref{SigmaPre1}) and Eq.(\ref{Sqrt1}) into Eq.(\ref{trBound}),   we have that 
\begin{align}
     | \text{tr}(G_i^t\Bar{\boldsymbol{\Sigma}}_i^t) | & \le  \frac{L_iid}{2t^{\frac{3}{4}}} \sqrt{\frac{3}{\alpha \nu}}
\end{align}

\end{proof}

\begin{lemma}\label{lemma:convexRelax}
    Given a convex function $f(x)$,  for Gaussian distribution with parameters  ${\boldsymbol{\theta} } := \{{\boldsymbol{\mu}}, \Sigma^{\frac{1}{2}}\} $,   let $\bar{J}({\boldsymbol{\theta}}):= \mathbb{E}_{p({\boldsymbol{x}};{\boldsymbol{\theta}})}[f({\boldsymbol{x}})]$. Then $\bar{J}({\boldsymbol{\theta}})$ is a convex function with respect to ${\boldsymbol{\theta}}$.
\end{lemma}
\begin{proof}
    For $\lambda \in [0,1]$, we have 
\begin{align}
    \lambda \bar{J}({\boldsymbol{\theta}}_1) + (1-\lambda)\bar{J}({\boldsymbol{\theta}}_2) & = \lambda \mathbb{E}_{\boldsymbol{z}\sim\mathcal{N}(\boldsymbol{0},\boldsymbol{I})}[f({\boldsymbol{\mu}_1} + \Sigma_1^{\frac{1}{2}}{\boldsymbol{z}}) ]   + (1-\lambda) \mathbb{E}_{\boldsymbol{z}\sim\mathcal{N}(\boldsymbol{0},\boldsymbol{I})}[f({\boldsymbol{\mu}_2} + \Sigma_2^{\frac{1}{2}}{\boldsymbol{z}}) ] \\
    & = \mathbb{E}[ \lambda f({\boldsymbol{\mu}_1} + \Sigma_1^{\frac{1}{2}}{\boldsymbol{z}}) + (1-\lambda) f({\boldsymbol{\mu}_2} +  \Sigma_2^{\frac{1}{2}}{\boldsymbol{z}}) ]  \\
    & \ge \mathbb{E}[  f\left(\lambda{\boldsymbol{\mu}_1} + (1-\lambda){\boldsymbol{\mu}_2} + (\lambda\Sigma_1^{\frac{1}{2}}+ (1-\lambda)\Sigma_2^{\frac{1}{2}}){\boldsymbol{z}} \right) ] \\
    & = \bar{J}(\lambda{\boldsymbol{\theta}}_1 + (1-\lambda){\boldsymbol{\theta}}_2)
\end{align}
\end{proof}

\begin{lemma}\label{lemma:convexEq}
    Given a convex function $f(x)$, let $J(\boldsymbol{\mu},\boldsymbol{\Sigma}):= \mathbb{E}_{\boldsymbol{x}\sim \mathcal{N}(\boldsymbol{\mu},\boldsymbol{\Sigma})}[f(\boldsymbol{x})]$. Then,  we have
    \begin{align}
        f(\boldsymbol{\mu}) - f(\boldsymbol{\mu}^*) \le J(\boldsymbol{\mu},\boldsymbol{\Sigma}) - J(\boldsymbol{\mu}^*,\boldsymbol{0})
    \end{align}
\end{lemma}
\begin{proof}
    From the definition of $J(\boldsymbol{\mu},\boldsymbol{\Sigma})$, we know that $f(\boldsymbol{\mu}^*)= J(\boldsymbol{\mu}^*,\boldsymbol{0})$.

    Note that $f(\boldsymbol{x})$ is a convex function, we have that
    \begin{align}
         f(\boldsymbol{\mu}) = f(\mathbb{E}_{\boldsymbol{x}\sim \mathcal{N}(\boldsymbol{\mu},\boldsymbol{\Sigma})}[\boldsymbol{x}]) \le \mathbb{E}_{\boldsymbol{x}\sim \mathcal{N}(\boldsymbol{\mu},\boldsymbol{\Sigma})}[ f(\boldsymbol{x})] = J(\boldsymbol{\mu},\boldsymbol{\Sigma})
    \end{align}
    It follows that 
    \begin{align}
        f(\boldsymbol{\mu}) - f(\boldsymbol{\mu}^*) \le J(\boldsymbol{\mu},\boldsymbol{\Sigma}) - J(\boldsymbol{\mu}^*,\boldsymbol{0})
    \end{align}
\end{proof}


\section{Proof of Theorem~\ref{MainTheorem}}\label{TheoremProof}

In this section, we prove our main Theorem~\ref{MainTheorem}. We decompose the proof into two parts. The proof of Theorem~\ref{PreTheorem1} and the proof of Theorem~\ref{PreTheorem2}.  Together with Theorem~\ref{PreTheorem1} and the  Theorem~\ref{PreTheorem2}, we achieve our main Theorem~\ref{MainTheorem}. 

\begin{theorem}\label{PreTheorem1}
    Suppose the assumptions~\ref{Assumption1}~\ref{Assumption2}~\ref{Assumption4} holds.       Set $\beta_t = t\beta$ with $\beta>0$,  $\alpha_t=  \sqrt{t+1} \alpha $ with  $ \alpha > 0$, and $\gamma_t = \frac{\alpha\nu}{\beta\sqrt{t+1}}$, and $\nu>0$, and  $\omega_t =1$.  Initialize  $\boldsymbol{\Sigma}_{k}^1 $ such that $\| \boldsymbol{\Sigma}_{k}^1 \|_2^{-1} \ge \frac{5}{3}\alpha \nu$  for $\forall k \in \{1,\cdots,K\}$.   Suppose   the constraints $ 
  \boldsymbol{H}_k^t \preceq \frac{1}{\alpha_t}(\frac{\beta_{t+1}}{\beta_t} - \omega_t) \boldsymbol{I} +   \frac{\beta_{t+1} \gamma_t}{\alpha_t} {\boldsymbol{\Sigma}_{k}^{t}} $ and $\nu \boldsymbol{I}  \preceq 
 \hat{G}_k^t=  {\boldsymbol{\Sigma}_k^t}^{-\frac{1}{2}} \boldsymbol{H}_k^t {\boldsymbol{\Sigma}_k^t}^{-\frac{1}{2}} $ always have feasible solutions.        Then, running Algorithm~\ref{MettaAlg} with  $T$-steps, we have 
    \begin{align}
      \frac{1}{T}\sum_{t=1}^T  \sum_{k=1}^K f_k (\Bar{\boldsymbol{\mu}}_k^t) - \sum_{k=1}^Kf_k(\Bar{\boldsymbol{\mu}}_k^*) & \le \frac{\sum_{k=1}^K \| \boldsymbol{\mu}_k^{1}  \!-\!  \boldsymbol{\mu}_k^*   \|_{ {\boldsymbol{\Sigma}_{k}^{1}}^{\!-\!1} }^2   }{2\beta T } \!+\! \frac{ 2\sqrt{T\!+\!1}  C_1 }{T} \!+\! \frac{4(T\!+\!1)^{\frac{1}{4}}C_2}{T } \! +\! \frac{\sqrt{T\!+\!2}C_3}{T}  \\ & \le  O( \frac{d^2K^4   }{\sqrt{T}})
    \end{align}
    where $\Bar{\boldsymbol{\mu}}_k^t=[\boldsymbol{\mu}_1^{t\top},\cdots,\boldsymbol{\mu}_k^{t\top}]^\top$ and $\Bar{\boldsymbol{\mu}}_k^*=[\boldsymbol{\mu}_1^{*\top},\cdots,\boldsymbol{\mu}_k^{*\top}]^\top$. And  $C_1 =   \frac{ 3\beta    \sum_{i=1}^K K L_i^2  (id+1)^2}{2\alpha \nu} $  and  $C_2 =  \frac{\sum_{i=1}^K \sqrt{3}idL_i}{\sqrt{ \alpha \nu}} \ $ , $C_3= \frac{\alpha\nu B}{\beta}$ 
\end{theorem}

\begin{proof}

For $\forall k \in \{1,\cdots,K\}$, we have 
\begin{align}
  &  \| \boldsymbol{\mu}_k^{t+1} - \boldsymbol{\mu}_k^*   \|_{ {\boldsymbol{\Sigma}_{k}^{t}}^{\!-\!1} }^2  \nonumber \\ & =  \| \boldsymbol{\mu}_k^{t} - \beta_t \ \boldsymbol{\Sigma}_{k}^{t}((\sum_{i=k}^K \hat{g}_{ik}^t) \!+\! \gamma_t \boldsymbol{\mu}_k^t   )  -  \boldsymbol{\mu}_k^*   \|_{ {\boldsymbol{\Sigma}_{k}^{t}}^{\!-\!1} }^2 \\
    & = \| \boldsymbol{\mu}_k^{t}  \!-\!  \boldsymbol{\mu}_k^*   \|_{ {\boldsymbol{\Sigma}_{k}^{t}}^{\!-\!1} }^2 - 2 \beta_t \left< \ \boldsymbol{\Sigma}_{k}^{t}((\sum_{i=k}^K \hat{g}_{ik}^t) \!+\! \gamma_t \boldsymbol{\mu}_k^t  )  ,\boldsymbol{\mu}_k^{t} \! -\!  \boldsymbol{\mu}_k^*\right>_{ {\boldsymbol{\Sigma}_{k}^{t}}^{\!-\!1} } + \beta_t^2  \|  \boldsymbol{\Sigma}_{k}^{t}((\sum_{i=k}^K \hat{g}_{ik}^t) \!+\! \gamma_t \boldsymbol{\mu}_k^t  )   \|_{ {\boldsymbol{\Sigma}_{k}^{t}}^{\!-\!1} }^2 \\
    & = \| \boldsymbol{\mu}_k^{t}  \!- \! \boldsymbol{\mu}_k^*   \|_{ {\boldsymbol{\Sigma}_{k}^{t}}^{\!-\!1} }^2 - 2 \beta_t \left<\gamma_t \boldsymbol{\mu}_k^t \!+\!\!  \sum_{i=k}^K \hat{g}_{ik}^t   ,\boldsymbol{\mu}_k^{t}  -  \boldsymbol{\mu}_k^*\right> + \beta_t^2  \| \boldsymbol{\Sigma}_{k}^{t}((\sum_{i=k}^K \hat{g}_{ik}^t) \!+\! \gamma_t \boldsymbol{\mu}_k^t  )   \|_{ {\boldsymbol{\Sigma}_{k}^{t}}^{\!-\!1} }^2 \label{IniNormEq}
\end{align}
Note that 
\begin{align}
    \gamma_t\big<\boldsymbol{\mu}_k^t, \boldsymbol{\mu}_k^{t}  -  \boldsymbol{\mu}_k^* \big> = \frac{\gamma_t}{2}\|\boldsymbol{\mu}_k^{t}  -  \boldsymbol{\mu}_k^*  \|_2^2 - \frac{\gamma_t}{2}\| \boldsymbol{\mu}_k^* \|_2^2 +  \frac{\gamma_t}{2}\| \boldsymbol{\mu}_k^{t} \|_2^2 \label{MuNormEq}
\end{align}
Plug Eq.(\ref{MuNormEq}) into Eq.(\ref{IniNormEq}), we have that
\begin{align}
   &  \| \boldsymbol{\mu}_k^{t+1} - \boldsymbol{\mu}_k^*   \|_{ {\boldsymbol{\Sigma}_{k}^{t}}^{\!-\!1} }^2 \nonumber \\
   & = \| \boldsymbol{\mu}_k^{t}  -  \boldsymbol{\mu}_k^*   \|_{ {\boldsymbol{\Sigma}_{k}^{t}}^{\!-\!1} }^2 
  - \!2 \beta_t \big< \sum_{i=k}^K \hat{g}_{ik}^t   ,\boldsymbol{\mu}_k^{t}  -  \boldsymbol{\mu}_k^*\big> 
  - \beta_t \gamma_t ( \|\boldsymbol{\mu}_k^{t}  \!-\!  \boldsymbol{\mu}_k^*  \|_2^2 \!-\! \| \boldsymbol{\mu}_k^* \|_2^2 \!+\! \| \boldsymbol{\mu}_k^{t} \|_2^2) + \beta_t^2  \| \boldsymbol{\Sigma}_{k}^{t}((\sum_{i=k}^K \hat{g}_{ik}^t) \!+\! \gamma_t \boldsymbol{\mu}_k^t  )   \|_{ {\boldsymbol{\Sigma}_{k}^{t}}^{\!-\!1} }^2
\end{align}
From Lemma~\ref{lemma:SigmaNormIeq}, we then have that 
\begin{align}
     \| \boldsymbol{\mu}_k^{t+1} - \boldsymbol{\mu}_k^*   \|_{ {\boldsymbol{\Sigma}_{k}^{t}}^{\!-\!1} }^2 &\le \| \boldsymbol{\mu}_k^{t}  -  \boldsymbol{\mu}_k^*   \|_{ {\boldsymbol{\Sigma}_{k}^{t}}^{\!-\!1} }^2 
  - \!2 \beta_t \big< \sum_{i=k}^K \hat{g}_{ik}^t   ,\boldsymbol{\mu}_k^{t}  -  \boldsymbol{\mu}_k^*\big> 
  - \beta_t \gamma_t ( \|\boldsymbol{\mu}_k^{t}  \!-\!  \boldsymbol{\mu}_k^*  \|_2^2 \!-\! \| \boldsymbol{\mu}_k^* \|_2^2 \!+\! \| \boldsymbol{\mu}_k^{t} \|_2^2) \nonumber \\ &  + 2\beta_t^2  \| \boldsymbol{\Sigma}_{k}^{t}((\sum_{i=k}^K \hat{g}_{ik}^t)   \|_{ {\boldsymbol{\Sigma}_{k}^{t}}^{\!-\!1} }^2 + 2\beta_t^2 \| \gamma_t{\boldsymbol{\Sigma}_{k}^{t}}^{\frac{1}{2}}  \boldsymbol{\mu}_k^{t} \|_2^2 \label{PreExgammat}
\end{align}
From Lemma~\ref{lemma:estimator} (b), we know that $\| {\boldsymbol{\Sigma}_{k}^{t}} \|_2 \le \frac{3}{2\alpha\nu} \frac{1}{t\sqrt{t}+3/2}$ , together with the setting $\beta_t = t \beta$ and $\gamma_t = \frac{\alpha\nu}{\beta\sqrt{t+1}}$, we know that
\begin{align}
    -\gamma_t \| \boldsymbol{\mu}_k^{t} \|_2^2+ 2 \beta_t \| \gamma_t{\boldsymbol{\Sigma}_{k}^{t}}^{\frac{1}{2}}  \boldsymbol{\mu}_k^{t} \|_2^2 &  =   -\gamma_t \| \boldsymbol{\mu}_k^{t} \|_2^2 + 2 \beta_t \gamma_t^2 \| {\boldsymbol{\Sigma}_{k}^{t}}^{\frac{1}{2}}  \boldsymbol{\mu}_k^{t} \|_2^2  \\
    & \le -\gamma_t \| \boldsymbol{\mu}_k^{t} \|_2^2 + 2 \beta_t \gamma_t^2 \|{\boldsymbol{\Sigma}_{k}^{t}}^{\frac{1}{2}}  \|_2^2 \|  \boldsymbol{\mu}_k^{t} \|_2^2  \\
    & = -\gamma_t \| \boldsymbol{\mu}_k^{t} \|_2^2 + 2 \beta_t \gamma_t^2 \|{\boldsymbol{\Sigma}_{k}^{t}}  \|_2 \|  \boldsymbol{\mu}_k^{t} \|_2^2 \\
    & = \gamma_t \| \boldsymbol{\mu}_k^{t} \|_2^2 ( -1 + 2 \beta_t \gamma_t \|{\boldsymbol{\Sigma}_{k}^{t}}  \|_2 ) \\
    & \le \gamma_t \| \boldsymbol{\mu}_k^{t} \|_2^2 (-1 + 2 t \beta  \frac{\alpha \nu}{\beta \sqrt{t+1}} \frac{3}{2\alpha\nu} \frac{1}{3/2+t\sqrt{t}}) \\
    & = \gamma_t \| \boldsymbol{\mu}_k^{t} \|_2^2 (-1 + \frac{3t}{\frac{3}{2} \sqrt{(t+1)}  + t \sqrt{t (t+1) }   }) 
\end{align}
 We now check the term $(-1 + \frac{3t}{\frac{3}{2} \sqrt{(t+1)}  + t \sqrt{t (t+1) }   })$. 
For $t=1$ and $t=2$, it is easy to see the term $(-1 + \frac{3t}{\frac{3}{2} \sqrt{(t+1)}  + t \sqrt{t (t+1) }   })  \le 0$.  For $t\ge 3$, we have that
\begin{align}
    \frac{3}{2} \sqrt{(t+1)}  + t \sqrt{t (t+1) }  -3t \ge 
    \frac{3}{2} \sqrt{(t+1)} + t^2 -3t \ge 0
\end{align}
It follows that $(-1 + \frac{3t}{\frac{3}{2} \sqrt{(t+1)}  + t \sqrt{t (t+1) }   })  \le 0$. Thus, we have that
\begin{align}
      -\gamma_t \| \boldsymbol{\mu}_k^{t} \|_2^2+ 2 \beta_t \| \gamma_t{\boldsymbol{\Sigma}_{k}^{t}}^{\frac{1}{2}}  \boldsymbol{\mu}_k^{t} \|_2^2 \le 0 \label{SrongIeq2}
\end{align}

Plug the inequality~(\ref{SrongIeq2}) into inequality~(\ref{PreExgammat}), we know that
\begin{align}
     \| \boldsymbol{\mu}_k^{t+1} - \boldsymbol{\mu}_k^*   \|_{ {\boldsymbol{\Sigma}_{k}^{t}}^{\!-\!1} }^2 & \le 
     \| \boldsymbol{\mu}_k^{t}  -  \boldsymbol{\mu}_k^*   \|_{ {\boldsymbol{\Sigma}_{k}^{t}}^{\!-\!1} }^2 
  - \!2 \beta_t \big< \sum_{i=k}^K \hat{g}_{ik}^t   ,\boldsymbol{\mu}_k^{t}  -  \boldsymbol{\mu}_k^*\big> 
  - \beta_t \gamma_t ( \|\boldsymbol{\mu}_k^{t}  \!-\!  \boldsymbol{\mu}_k^*  \|_2^2 \!-\! \| \boldsymbol{\mu}_k^* \|_2^2 ) \nonumber \\ & + 2\beta_t^2  \| \boldsymbol{\Sigma}_{k}^{t}((\sum_{i=k}^K \hat{g}_{ik}^t)   \|_{ {\boldsymbol{\Sigma}_{k}^{t}}^{\!-\!1} }^2
\end{align}

It follows that
\begin{align}
    \sum_{k=1}^K  \| \boldsymbol{\mu}_k^{t+1} - \boldsymbol{\mu}_k^*   \|_{ {\boldsymbol{\Sigma}_{k}^{t}}^{\!-\!1} }^2 & \le  \sum_{k=1}^K\| \boldsymbol{\mu}_k^{t}  -  \boldsymbol{\mu}_k^*   \|_{ {\boldsymbol{\Sigma}_{k}^{t}}^{\!-\!1} }^2 - 2 \beta_t  \sum_{k=1}^K \left< \ \sum_{i=k}^K \hat{g}_{ik}^t   ,\boldsymbol{\mu}_k^{t}  -  \boldsymbol{\mu}_k^*\right> + 2 \beta_t^2  \sum_{k=1}^K \| \boldsymbol{\Sigma}_{k}^{t}(\sum_{i=k}^K \hat{g}_{ik} )   \|_{ {\boldsymbol{\Sigma}_{k}^{t}}^{\!-\!1} }^2 \nonumber \\ &  -  \sum_{k=1}^K   \beta_t \gamma_t ( \|\boldsymbol{\mu}_k^{t}  \!-\!  \boldsymbol{\mu}_k^*  \|_2^2 \!-\! \| \boldsymbol{\mu}_k^* \|_2^2 )
\end{align}
Then, we have that
\begin{align}
  \mathbb{E}   \sum_{k=1}^K  \| \boldsymbol{\mu}_k^{t+1} - \boldsymbol{\mu}_k^*   \|_{ {\boldsymbol{\Sigma}_{k}^{t}}^{\!-\!1} }^2  & \le  \sum_{k=1}^K \mathbb{E} \| \boldsymbol{\mu}_k^{t}  -  \boldsymbol{\mu}_k^*   \|_{ {\boldsymbol{\Sigma}_{k}^{t}}^{\!-\!1} }^2 - 2 \beta_t  \mathbb{E}  \sum_{k=1}^K \left< \ \sum_{i=k}^K \hat{g}_{ik}^t   ,\boldsymbol{\mu}_k^{t}  -  \boldsymbol{\mu}_k^*\right> + 2\beta_t^2  \mathbb{E}  \sum_{k=1}^K \| \boldsymbol{\Sigma}_{k}^{t}(\sum_{i=k}^K \hat{g}_{ik} )   \|_{ {\boldsymbol{\Sigma}_{k}^{t}}^{\!-\!1} }^2   \nonumber \\ &  -  \sum_{k=1}^K   \beta_t \gamma_t ( \mathbb{E}\|\boldsymbol{\mu}_k^{t}  \!-\!  \boldsymbol{\mu}_k^*  \|_2^2 \!-\! \| \boldsymbol{\mu}_k^* \|_2^2 )   \label{Esum}
\end{align}

Note that $\Bar{\boldsymbol{\mu}}_k^t=[\boldsymbol{\mu}_1^{t\top},\cdots,\boldsymbol{\mu}_k^{t\top}]^\top$ and $\Bar{\boldsymbol{\mu}}_k^*=[\boldsymbol{\mu}_1^{*\top},\cdots,\boldsymbol{\mu}_k^{*\top}]^\top$, together with Lemma~\ref{lemma:estimator} (a),  we have that
\begin{align}
   \mathbb{E}   \sum_{k=1}^K \left< \ \sum_{i=k}^K \hat{g}_{ik}^t   ,\boldsymbol{\mu}_k^{t}  -  \boldsymbol{\mu}_k^*\right> & =  \mathbb{E} 
 \sum_{i=1}^K \left< \ \sum_{k=1}^i \hat{g}_{ik}^t   ,\boldsymbol{\mu}_k^{t}  -  \boldsymbol{\mu}_k^*\right> = \sum_{i=1}^K \big< g_i^t, \Bar{\boldsymbol{\mu}}_i^t - \Bar{\boldsymbol{\mu}}_i^*  \big> \label{InnerEq}
\end{align}
where $g_i^t = \nabla_{\Bar{\boldsymbol{\mu}}_i^t}\mathbb{E}_{\boldsymbol{x} \sim \mathcal{N}(\Bar{\boldsymbol{\mu}}_i^t, \Bar{\boldsymbol{\Sigma}}_i^t)}[f_i(\boldsymbol{x})] = \nabla_{\Bar{\boldsymbol{\mu}}_i^t} J_i(\Bar{\boldsymbol{\mu}}_i^t, \Bar{\boldsymbol{\Sigma}}_i^t)$

From Eq.~(\ref{InnerEq}) and Eq.~(\ref{Esum}), we have that  
\begin{align}
    \sum_{i=1}^K \big< g_i^t, \Bar{\boldsymbol{\mu}}_i^t - \Bar{\boldsymbol{\mu}}_i^*  \big>  & \le  \frac{\sum_{k=1}^K\mathbb{E}\| \boldsymbol{\mu}_k^{t}  -  \boldsymbol{\mu}_k^*   \|_{ {\boldsymbol{\Sigma}_{k}^{t}}^{\!-\!1} }^2  - \sum_{k=1}^K  \mathbb{E}   \| \boldsymbol{\mu}_k^{t+1} - \boldsymbol{\mu}_k^*   \|_{ {\boldsymbol{\Sigma}_{k}^{t}}^{\!-\!1} }^2    }{2\beta_t} + \beta_t \sum_{k=1}^K\mathbb{E} \| \boldsymbol{\Sigma}_{k}^{t}(\sum_{i=k}^K \hat{g}_{ik} )   \|_{ {\boldsymbol{\Sigma}_{k}^{t}}^{\!-\!1} }^2  \nonumber \\ &  -  \sum_{k=1}^K  \frac{\gamma_t}{2}   ( \mathbb{E}\|\boldsymbol{\mu}_k^{t}  \!-\!  \boldsymbol{\mu}_k^*  \|_2^2 \!-\! \| \boldsymbol{\mu}_k^* \|_2^2 )  
    \label{KeyInnerProduct}
\end{align}

From Lemma~\ref{lemma:convexRelax}, we know that for $ \forall i \in \{1,\cdots,K\}$,  $J_i(\Bar{\boldsymbol{\mu}}_i,\Bar{\boldsymbol{\Sigma}}_i)$    is  convex function w.r.t. $\Bar{\boldsymbol{\mu}}_i$ and $\Bar{\boldsymbol{\Sigma}}_i^{\frac{1}{2}}$. Then, we have that   
\begin{align}
    J_i(\Bar{\boldsymbol{\mu}}_i^t,\Bar{\boldsymbol{\Sigma}}_i^t) - J_i(\Bar{\boldsymbol{\mu}}_i^*, 0) \le \big< g_i^t, \Bar{\boldsymbol{\mu}}_i^t - \Bar{\boldsymbol{\mu}}_i^*  \big>   +  \big< \nabla_{\Bar{\boldsymbol{\Sigma}}_i^{\frac{1}{2}} = \Bar{\boldsymbol{\Sigma}}_i^{t\frac{1}{2}} }  J_i(\Bar{\boldsymbol{\mu}}_i^t,\Bar{\boldsymbol{\Sigma}}_i^t), \Bar{\boldsymbol{\Sigma}}_i^{t\frac{1}{2}}  -0   \big> \label{ConvexExpand}
\end{align}

Denote  $ G_i^t    = \nabla_{\Bar{\boldsymbol{\Sigma}}_i = \Bar{\boldsymbol{\Sigma}}_i^{t} }  J_i(\Bar{\boldsymbol{\mu}}_i^t,\Bar{\boldsymbol{\Sigma}}_i^t)$. Note that  $\nabla _ {\Bar{\boldsymbol{\Sigma}}_i^{t\frac{1}{2}} } J_i = \Bar{\boldsymbol{\Sigma}}_i^{t\frac{1}{2}}  \nabla _ {\Bar{\boldsymbol{\Sigma}}_i^t } {J_i} +   \nabla _ {\Bar{\boldsymbol{\Sigma}}_i^t } {J_i} \Bar{\boldsymbol{\Sigma}}_i^{t\frac{1}{2}} $, and $G_i^t$,  $\nabla _ {\Bar{\boldsymbol{\Sigma}}_i^t } {J_i}$ and $\Bar{\boldsymbol{\Sigma}}_i^{t\frac{1}{2}} $ are symmetric matrix, it follows that
\begin{align}
    \big< \nabla_{\Bar{\boldsymbol{\Sigma}}_i^{\frac{1}{2}} = \Bar{\boldsymbol{\Sigma}}_i^{t\frac{1}{2}} }  J_i(\Bar{\boldsymbol{\mu}}_i^t,\Bar{\boldsymbol{\Sigma}}_i^t), \Bar{\boldsymbol{\Sigma}}_i^{t\frac{1}{2}}  -0   \big>  & = \big< \Bar{\boldsymbol{\Sigma}}_i^{t\frac{1}{2}} G_i^t + G_i^t \Bar{\boldsymbol{\Sigma}}_i^{t\frac{1}{2}}, \Bar{\boldsymbol{\Sigma}}_i^{t\frac{1}{2}} \big> \\
    & = 2 \big< G_i^t , \Bar{\boldsymbol{\Sigma}}_i^{t } \big> = 
    2\text{tr}(G_i^t \Bar{\boldsymbol{\Sigma}}_i^{t} ) \label{G12tr}
\end{align}


Plug Eq.(\ref{G12tr}) into Eq.~(\ref{ConvexExpand}), we have that
\begin{align}
    J_i(\Bar{\boldsymbol{\mu}}_i^t,\Bar{\boldsymbol{\Sigma}}_i^t) - J_i(\Bar{\boldsymbol{\mu}}_i^*, 0) \le \big< g_i^t, \Bar{\boldsymbol{\mu}}_i^t - \Bar{\boldsymbol{\mu}}_i^*  \big>   + 2\text{tr}(G_i^t \Bar{\boldsymbol{\Sigma}}_i^{t} ) 
\end{align}

It follows that 
\begin{align}
 \sum_{i=1}^K J_i(\Bar{\boldsymbol{\mu}}_i^t,\Bar{\boldsymbol{\Sigma}}_i^t) -  \sum_{i=1}^K J_i(\Bar{\boldsymbol{\mu}}_i^*,0)  \le     \sum_{i=1}^K \big< g_i^t, \Bar{\boldsymbol{\mu}}_i^t - \Bar{\boldsymbol{\mu}}_i^*  \big>  + 2\sum_{i=1}^K\text{tr}(G_i^t \Bar{\boldsymbol{\Sigma}}_i^{t} ) \label{KeyExpand}
\end{align}
Plug Eq.(\ref{KeyInnerProduct}) into Eq.(\ref{KeyExpand}), we have that
\begin{align}
  &   \sum_{i=1}^K J_i(\Bar{\boldsymbol{\mu}}_i^t,\Bar{\boldsymbol{\Sigma}}_i^t) -  \sum_{i=1}^K J_i(\Bar{\boldsymbol{\mu}}_i^*,0) \nonumber \\
  & \le \frac{\sum_{k=1}^K\mathbb{E}\| \boldsymbol{\mu}_k^{t}  -  \boldsymbol{\mu}_k^*   \|_{ {\boldsymbol{\Sigma}_{k}^{t}}^{\!-\!1} }^2  - \sum_{k=1}^K  \mathbb{E}   \| \boldsymbol{\mu}_k^{t+1} - \boldsymbol{\mu}_k^*   \|_{ {\boldsymbol{\Sigma}_{k}^{t}}^{\!-\!1} }^2  - \beta_t\gamma_t \sum_{i=1}^K  \mathbb{E} \| {\boldsymbol{\mu}}_i^t - {\boldsymbol{\mu}}_i^*\|_2^2  }{2\beta_t} \nonumber \\ & +  \beta_t \sum_{k=1}^K\mathbb{E} \| \boldsymbol{\Sigma}_{k}^{t}(\sum_{i=k}^K \hat{g}_{ik} )   \|_{ {\boldsymbol{\Sigma}_{k}^{t}}^{\!-\!1} }^2  + 2\sum_{i=1}^K\text{tr}(G_i^t \Bar{\boldsymbol{\Sigma}}_i^{t} ) + \frac{\gamma_t}{2} \sum_{k=1}^K \| \boldsymbol{\mu}_k^* \|_2^2
  \label{SingleJ}
\end{align}

In addition,  we have that
\begin{align}
 & \frac{1}{\beta_{t+1}}  \mathbb{E}   \| \boldsymbol{\mu}_k^{t+1} - \boldsymbol{\mu}_k^*   \|_{ {\boldsymbol{\Sigma}_{k}^{t\!+\!1}}^{\!-\!1} }^2 - \frac{1}{\beta_{t}}\mathbb{E}   \| \boldsymbol{\mu}_k^{t+1} - \boldsymbol{\mu}_k^*   \|_{ {\boldsymbol{\Sigma}_{k}^{t}}^{\!-\!1} }^2   -  \gamma_t   \mathbb{E} \| {\boldsymbol{\mu}}_i^t - {\boldsymbol{\mu}}_i^*\|_2^2 \nonumber \\
    & = \frac{1}{\beta_{t+1}} \mathbb{E} \big< {\boldsymbol{\Sigma}_{k}^{t\!+\!1}}^{\!-\!1}(\boldsymbol{\mu}_k^{t+1} - \boldsymbol{\mu}_k^*),  \boldsymbol{\mu}_k^{t+1} - \boldsymbol{\mu}_k^*\big> -  \frac{1}{\beta_{t}} \mathbb{E} \big< {\boldsymbol{\Sigma}_{k}^{t}}^{\!-\!1}(\boldsymbol{\mu}_k^{t+1} - \boldsymbol{\mu}_k^*),  \boldsymbol{\mu}_k^{t+1} - \boldsymbol{\mu}_k^*\big> - \gamma_t   \mathbb{E} \| {\boldsymbol{\mu}}_i^t - {\boldsymbol{\mu}}_i^*\|_2^2 \\
    & = \mathbb{E}  \big<   \big( \frac{1}{\beta_{t+1}}  {\boldsymbol{\Sigma}_{k}^{t\!+\!1}}^{\!-\!1}  - \frac{1}{\beta_{t}}  {\boldsymbol{\Sigma}_{k}^{t}}^{\!-\!1}  - \gamma_t \boldsymbol{I}  \big) (\boldsymbol{\mu}_k^{t+1} - \boldsymbol{\mu}_k^*),  \boldsymbol{\mu}_k^{t+1} - \boldsymbol{\mu}_k^*\big>  \label{DifferenSigmaTerm}  
\end{align}
 Note that ${\boldsymbol{\Sigma}_{k}^{t\!+\!1}}^{\!-\!1} =  \omega_t {\boldsymbol{\Sigma}_{k}^{t}}^{\!-\!1} + \alpha_t \hat{G}_k^t = {\boldsymbol{\Sigma}_{k}^{t}}^{-\frac{1}{2}}( \omega_t  \boldsymbol{I} + \alpha_t \boldsymbol{H}_k^t){\boldsymbol{\Sigma}_{k}^{t}}^{-\frac{1}{2}}$, we have that 
\begin{align}
    \frac{1}{\beta_{t+1}}  {\boldsymbol{\Sigma}_{k}^{t\!+\!1}}^{\!-\!1}  - \frac{1}{\beta_{t}}  {\boldsymbol{\Sigma}_{k}^{t}}^{\!-\!1}   - \gamma_t \boldsymbol{I} & = \frac{1}{\beta_{t+1}}{\boldsymbol{\Sigma}_{k}^{t}}^{-\frac{1}{2}}( \omega_t  \boldsymbol{I} + \alpha_t \boldsymbol{H}_k^t){\boldsymbol{\Sigma}_{k}^{t}}^{-\frac{1}{2}} - \frac{1}{\beta_{t}}  {\boldsymbol{\Sigma}_{k}^{t}}^{\!-\!1}  - \gamma_t \boldsymbol{I} \\
    & = {\boldsymbol{\Sigma}_{k}^{t}}^{-\frac{1}{2}} (\frac{\omega_t }{\beta_{t+1}}\boldsymbol{I} + \frac{\alpha_t}{\beta_{t+1}}\boldsymbol{H}_k^t - \frac{1}{\beta_t} \boldsymbol{I}  - \gamma_t {\boldsymbol{\Sigma}_{k}^{t}} )  {\boldsymbol{\Sigma}_{k}^{t}}^{-\frac{1}{2}}
\end{align}
Because of $\boldsymbol{H}_k^t \preceq \frac{1}{\alpha_t}(\frac{\beta_{t+1}}{\beta_t} - \omega_t ) \boldsymbol{I} +   \frac{\beta_{t+1} \gamma_t}{\alpha_t} {\boldsymbol{\Sigma}_{k}^{t}} $ in algorithm~\ref{MettaAlg}, we have that
\begin{align}
    \frac{1}{\beta_{t+1}}  {\boldsymbol{\Sigma}_{k}^{t\!+\!1}}^{\!-\!1}  - \frac{1 }{\beta_{t}}  {\boldsymbol{\Sigma}_{k}^{t}}^{\!-\!1}  - \gamma_t \boldsymbol{I}  & =  {\boldsymbol{\Sigma}_{k}^{t}}^{-\frac{1}{2}} (\frac{\omega_t }{\beta_{t+1}}\boldsymbol{I} + \frac{\alpha_t}{\beta_{t+1}}\boldsymbol{H}_k^t - \frac{1}{\beta_t} \boldsymbol{I}  - \gamma_t {\boldsymbol{\Sigma}_{k}^{t}}   )  {\boldsymbol{\Sigma}_{k}^{t}}^{-\frac{1}{2}} \\
    & \preceq  {\boldsymbol{\Sigma}_{k}^{t}}^{-\frac{1}{2}} (\frac{\omega_t }{\beta_{t+1}}\boldsymbol{I} + \frac{1}{\beta_{t+1}}(\frac{\beta_{t+1}}{\beta_t} - \omega_t ) \boldsymbol{I}  + \gamma_t {\boldsymbol{\Sigma}_{k}^{t}}   - \frac{1}{\beta_t} \boldsymbol{I}  - \gamma_t {\boldsymbol{\Sigma}_{k}^{t}}   )  {\boldsymbol{\Sigma}_{k}^{t}}^{-\frac{1}{2}}  \\ & 
     \preceq \boldsymbol{0} \label{NegSigma}
\end{align}
Plug Eq.(\ref{NegSigma}) into Eq.(\ref{DifferenSigmaTerm}), we know that
\begin{align}
    \frac{1}{\beta_{t+1}}  \mathbb{E}   \| \boldsymbol{\mu}_k^{t+1} - \boldsymbol{\mu}_k^*   \|_{ {\boldsymbol{\Sigma}_{k}^{t\!+\!1}}^{\!-\!1} }^2 - \frac{1}{\beta_{t}}\mathbb{E}   \| \boldsymbol{\mu}_k^{t+1} - \boldsymbol{\mu}_k^*   \|_{ {\boldsymbol{\Sigma}_{k}^{t}}^{\!-\!1} }^2  - \gamma_t   \mathbb{E} \| {\boldsymbol{\mu}}_i^t - {\boldsymbol{\mu}}_i^*\|_2^2
    & \le 0 
\end{align}


  Telescope  with Eq.(\ref{SingleJ}), we have that
\begin{align}
    &  \sum_{t=1}^T \sum_{i=1}^K J_i(\Bar{\boldsymbol{\mu}}_i^t,\Bar{\boldsymbol{\Sigma}}_i^t) - \sum_{t=1}^T \sum_{i=1}^K J_i(\Bar{\boldsymbol{\mu}}_i^*,0) \nonumber \\
  & \le \frac{\sum_{k=1}^K\mathbb{E}\| \boldsymbol{\mu}_k^{1}  -  \boldsymbol{\mu}_k^*   \|_{ {\boldsymbol{\Sigma}_{k}^{1}}^{\!-\!1} }^2      }{2\beta_1} - \frac{ \sum_{k=1}^K  \mathbb{E}   \| \boldsymbol{\mu}_k^{T+1} - \boldsymbol{\mu}_k^*   \|_{ {\boldsymbol{\Sigma}_{k}^{T}}^{\!-\!1} }^2 }{2\beta_T} +  \sum_{t=1}^T  \beta_t \sum_{k=1}^K\mathbb{E} \| \boldsymbol{\Sigma}_{k}^{t}(\sum_{i=k}^K \hat{g}_{ik} )   \|_{ {\boldsymbol{\Sigma}_{k}^{t}}^{\!-\!1} }^2  \nonumber \\ & + 2 \sum_{t=1}^T \sum_{i=1}^K\text{tr}(G_i^t \Bar{\boldsymbol{\Sigma}}_i^{t} )  + \frac{1}{2} \sum_{t=1}^T \gamma_t \sum_{k=1}^K \| \boldsymbol{\mu}_k^* \|_2^2 \label{PreConverge}
\end{align}

We now show the upper bound of term $ \sum_{t=1}^T  \beta_t \sum_{k=1}^K\mathbb{E} \| \boldsymbol{\Sigma}_{k}^{t}(\sum_{i=k}^K \hat{g}_{ik} )   \|_{ {\boldsymbol{\Sigma}_{k}^{t}}^{\!-\!1} }^2 $. 

Note that  $\beta_t = t\beta$, together with  Lemma~\ref{lemma:estimator} (c), we know that
\begin{align}
    \sum_{t=1}^T \beta_t  \sum_{k=1}^K\mathbb{E} \| \boldsymbol{\Sigma}_{k}^{t}(\sum_{i=k}^K \hat{g}_{ik} )   \|_{ {\boldsymbol{\Sigma}_{k}^{t}}^{\!-\!1} }^2  & \le \sum_{t=1}^T t \frac{C_1}{ t^{\frac{3}{2}} } \le   2 \sqrt{T+1} C_1  \label{ErrorTerm2}
\end{align}
where $C_1 =   \frac{ 3\beta    \sum_{i=1}^K K L_i^2  (id+1)^2}{2\alpha \nu} $.

We now show the upper bound of term $2 \sum_{t=1}^T \sum_{i=1}^K\text{tr}(G_i^t \Bar{\boldsymbol{\Sigma}}_i^{t} )$. 

From  Lemma~\ref{lemma:trGsigma}, we know that
\begin{align}
    2 \sum_{t=1}^T \sum_{i=1}^K\text{tr}(G_i^t \Bar{\boldsymbol{\Sigma}}_i^{t} ) & \le  2 \sum_{t=1}^T \sum_{i=1}^K \frac{L_iid}{2t^{\frac{3}{4}}} \sqrt{\frac{3}{\alpha \nu}}\\
    & = \sum_{t=1}^T C_2 \frac{ 1 }{ t^{\frac{3}{4}} } \\ 
    & \le   4 (T+1)^{\frac{1}{4}} C_2\label{BoundTerm2}
\end{align}
where $C_2 =  \frac{\sum_{i=1}^K \sqrt{3}idL_i}{\sqrt{ \alpha \nu}} \ $.

We now show the upper bound of the term $\frac{1}{2} \sum_{t=1}^T \gamma_t \sum_{k=1}^K \| \boldsymbol{\mu}_k^* \|_2^2$.

Note that the optimal solution is bounded from Assumption~\ref{Assumption4}, i.e., $\sum_{k=1}^K \| \boldsymbol{\mu}_k^* \|_2^2 \le B$. Together with the setting $\gamma_t = \frac{\alpha\nu}{\beta\sqrt{t+1}}$,  we can achieve that
\begin{align}
    \frac{1}{2} \sum_{t=1}^T \gamma_t \sum_{k=1}^K \| \boldsymbol{\mu}_k^* \|_2^2 \le \frac{B}{2 }\sum_{t=1}^T \gamma_t \le \frac{\alpha\nu B}{2\beta} 2\sqrt{T+2} = \sqrt{T+2} C_3
\end{align}
where $C_3= \frac{\alpha\nu B}{\beta}$

Plug Eq.(\ref{ErrorTerm2}) and Eq.(\ref{BoundTerm2}) into Eq.(\ref{PreConverge}), we have that 
\begin{align}
    &  \sum_{t=1}^T \sum_{i=1}^K J_i(\Bar{\boldsymbol{\mu}}_i^t,\Bar{\boldsymbol{\Sigma}}_i^t) - \sum_{t=1}^T \sum_{i=1}^K J_i(\Bar{\boldsymbol{\mu}}_i^*,0) \nonumber \\
  & \le \frac{\sum_{k=1}^K\mathbb{E}\| \boldsymbol{\mu}_k^{1}  -  \boldsymbol{\mu}_k^*   \|_{ {\boldsymbol{\Sigma}_{k}^{1}}^{\!-\!1} }^2      }{2\beta_1} - \frac{ \sum_{k=1}^K  \mathbb{E}   \| \boldsymbol{\mu}_k^{T+1} - \boldsymbol{\mu}_k^*   \|_{ {\boldsymbol{\Sigma}_{k}^{T}}^{\!-\!1} }^2 }{2\beta_T}  +  2\sqrt{T+1} C_1  + 4(T+1)^{\frac{1}{4}}C_2 + \sqrt{T+2}C_3
\end{align}

From Lemma~\ref{lemma:convexEq}, we then have that
\begin{align}
 \sum_{t=1}^T \left( \sum_{k=1}^K( f_k (\Bar{\boldsymbol{\mu}}_k^t)  - f_k(\Bar{\boldsymbol{\mu}}_k^*) )\right) & \le   \sum_{t=1}^T \sum_{i=1}^K J_i(\Bar{\boldsymbol{\mu}}_i^t,\Bar{\boldsymbol{\Sigma}}_i^t) - \sum_{t=1}^T \sum_{i=1}^K J_i(\Bar{\boldsymbol{\mu}}_i^*,0)  \\
 & \le \frac{\sum_{k=1}^K\mathbb{E}\| \boldsymbol{\mu}_k^{1}  -  \boldsymbol{\mu}_k^*   \|_{ {\boldsymbol{\Sigma}_{k}^{1}}^{\!-\!1} }^2      }{2\beta_1} - \frac{ \sum_{k=1}^K  \mathbb{E}   \| \boldsymbol{\mu}_k^{T+1} - \boldsymbol{\mu}_k^*   \|_{ {\boldsymbol{\Sigma}_{k}^{T}}^{\!-\!1} }^2 }{2\beta_T} \nonumber \\ &  +  +  2\sqrt{T+1} C_1  + 4(T+1)^{\frac{1}{4}}C_2 + \sqrt{T+2}C_3 \label{SumTBound}
\end{align}

Finally, divide $T$ on both sides of Eq.(\ref{SumTBound}),  we have that
\begin{align}
    \frac{1}{T}\sum_{t=1}^T  \sum_{k=1}^K f_k (\Bar{\boldsymbol{\mu}}_k^t) - \sum_{k=1}^Kf_k(\Bar{\boldsymbol{\mu}}_k^*) \le \frac{\sum_{k=1}^K \| \boldsymbol{\mu}_k^{1}  -  \boldsymbol{\mu}_k^*   \|_{ {\boldsymbol{\Sigma}_{k}^{1}}^{\!-\!1} }^2   }{2\beta T } + \frac{ 2\sqrt{T+1}  C_1 }{T} + \frac{4(T+1)^{\frac{1}{4}}C_2}{T }  + \frac{\sqrt{T+2}C_3}{T}
\end{align}

\end{proof}

\begin{theorem}\label{PreTheorem2}
   Set $\beta_t = t\beta$ with $\beta>0$,  $\alpha_t=  \sqrt{t+1} \alpha $ with  $ \alpha > 0$, and $\gamma_t = \frac{\alpha\nu}{\beta\sqrt{t+1}}$, and $\nu>0$, and  $\omega_t =1$.         Then, during the  running process of Algorithm~\ref{MettaAlg}, the constraints $ 
  \boldsymbol{H}_k^t \preceq \frac{1}{\alpha_t}(\frac{\beta_{t+1}}{\beta_t} - \omega_t) \boldsymbol{I} +   \frac{\beta_{t+1} \gamma_t}{\alpha_t} {\boldsymbol{\Sigma}_{k}^{t}} $ and $\nu \boldsymbol{I}  \preceq 
 \hat{G}_k^t=  {\boldsymbol{\Sigma}_k^t}^{-\frac{1}{2}} \boldsymbol{H}_k^t {\boldsymbol{\Sigma}_k^t}^{-\frac{1}{2}} $ always have feasible solutions. 
\end{theorem}

\begin{proof}
    To ensure the constraints $ 
  \boldsymbol{H}_k^t \preceq \frac{1}{\alpha_t}(\frac{\beta_{t+1}}{\beta_t} - \omega_t) \boldsymbol{I} + \frac{\beta_{t+1} \gamma_t}{\alpha_t} {\boldsymbol{\Sigma}_{k}^{t}}$ and $\nu \boldsymbol{I}  \preceq 
 \hat{G}_k^t=  {\boldsymbol{\Sigma}_k^t}^{-\frac{1}{2}} \boldsymbol{H}_k^t {\boldsymbol{\Sigma}_k^t}^{-\frac{1}{2}} $ always feasible during the algorithm, it is equivalent to show the constraints Eq.(\ref{Constrains}) always has feasible solutions $\boldsymbol{H}_k^t$. 
 \begin{align}
 \label{Constrains}
     \nu {\boldsymbol{\Sigma}_k^t} \preceq \boldsymbol{H}_k^t \preceq \frac{1}{\alpha_t}(\frac{\beta_{t+1}}{\beta_t} - \omega_t ) \boldsymbol{I} +   \frac{\beta_{t+1} \gamma_t}{\alpha_t} {\boldsymbol{\Sigma}_{k}^{t}}
 \end{align}
It is equivalent to show that the inequality~(\ref{feasibleEq}) always holds true. 
\begin{align}
     \nu {\boldsymbol{\Sigma}_k^t} \preceq \frac{1}{\alpha_t}(\frac{\beta_{t+1}}{\beta_t} - \omega_t ) \boldsymbol{I} +   \frac{\beta_{t+1} \gamma_t}{\alpha_t} {\boldsymbol{\Sigma}_{k}^{t}} \label{feasibleEq}
\end{align}
Then, it is equivalent to show that the inequality~(\ref{feasibleEq2}) always holds true. 
\begin{align}
    (\nu -\frac{\beta_{t+1} \gamma_t}{\alpha_t} )  {\boldsymbol{\Sigma}_{k}^{t}} \preceq  \frac{1}{\alpha_t}(\frac{\beta_{t+1}}{\beta_t} - \omega_t ) \boldsymbol{I} \label{feasibleEq2}
\end{align}

We first check the left hand side of the inequality~(\ref{feasibleEq2}). 

Note that   the setting 
 $\beta_t = t\beta$ with $\beta>0$,  $\alpha_t=  \sqrt{t+1} \alpha $ with  $ \alpha > 0$, and $\gamma_t = \frac{\alpha\nu}{\beta\sqrt{t+1}}$, and $\nu>0$, and  $\omega_t =1$. We can achieve that
 \begin{align}
       (\nu -\frac{\beta_{t+1} \gamma_t}{\alpha_t} )  {\boldsymbol{\Sigma}_{k}^{t}}  & = (\nu - \frac{(t+1)\beta}{\sqrt{t+1}\alpha} \frac{\alpha\nu}{\beta\sqrt{t+1}} )  {\boldsymbol{\Sigma}_{k}^{t}}  \\
       & = (\nu -  \nu)  {\boldsymbol{\Sigma}_{k}^{t}}  = 0      
 \end{align}

We now check the right hand side of the inequality~(\ref{feasibleEq2}). 

\begin{align}
    \frac{1}{\alpha_t}(\frac{\beta_{t+1}}{\beta_t} - \omega_t ) & = \frac{1}{\alpha_t} ( \frac{(t+1)\beta}{t\beta} -1 )  = \frac{1}{\alpha_t t } > 0
\end{align}
Thus, the   inequality~(\ref{feasibleEq2}) always hold true. As a result, the constraints set always have feasible solutions.  
 
\end{proof}


\section{Related Background}
\label{Background}

\subsection{Diffusion Model Sampling with SDE solver}

The sampling phase of the diffusion model from noise to image can be implemented by solving the  stochastic differential equation (SDE)~\citep{song2020score,kingma2021variational} as in Eq.(\ref{SDE}):
\begin{align}
\label{SDE}
    \text{d}  \hat{\boldsymbol{x}}_\tau = [\hat{f}(\tau) \hat{\boldsymbol{x}}_\tau + \frac{g(\tau)^2}{\sigma_\tau}\boldsymbol{\epsilon}_{\phi}(\hat{\boldsymbol{x}}_\tau,\tau)   ] \text{d}\tau + g(\tau) \text{d}\bar{\boldsymbol{w}}_\tau
\end{align}
where  $\bar{\boldsymbol{w}}_\tau$ is the reverse-time Wiener process,  $\tau$ denotes time  changing from $K$ to 0, $\boldsymbol{\epsilon}_{\phi}(\tilde{\boldsymbol{x}}_\tau,\tau)$  denotes the diffusion model noise prediction with input $\hat{\boldsymbol{x}}_\tau$ and time $\tau$. And $\sigma_\tau$ denotes the standard deviation of the diffusion noise scheme at time $\tau$.  And $\hat{f}(\tau):=\frac{\text{d} \log \alpha_\tau}{\text{d} \tau}$, where $\alpha_\tau$ denotes the scaling parameter scheme in the diffusion model.  And $g(\tau)^2:=\frac{\text{d} \sigma_\tau^2}{\text{d} \tau} -2\hat{f}(\tau) \sigma_\tau^2$~
\citep{kingma2021variational}.

Recently, \cite{lu2022dpm,lu2022dpm2} proposed a DPM solver for solving the diffusion SDE with a small number of samples. 
The first-order SDE DPM solver is given in Eq.(\ref{DPM1}). 
\begin{align}
   \hat{\boldsymbol{x}}_\tau = \frac{\alpha_\tau}{\alpha_{\tau'}} \hat{\boldsymbol{x}}_{\tau'} -2 \sigma_\tau (e^{h}-1) \boldsymbol{\epsilon}_\phi(\hat{\boldsymbol{x}}_{\tau'},{\tau'}) + \sigma_\tau \sqrt{e^{2h}-1} \boldsymbol{z}
    \label{DPM1}
\end{align}
where  $\boldsymbol{z} \!\sim\! \mathcal{N}(\boldsymbol{0},\boldsymbol{I})$,  $h\!=\!\lambda_\tau \!-\! \lambda_{\tau'}$, and  $\lambda_\tau=\log(\alpha_\tau / \sigma_\tau)$ and $\lambda_{\tau'}=\log(\alpha_{\tau'} / \sigma_{\tau'})$. And   $\alpha_\tau$, $\alpha_{\tau'}$ denote the scaling parameter at step $\tau$ and $\tau'$ in diffusion model, respectively.


The second-order DPM solver is provided in Eq.(\ref{DPM2}):
\begin{align}
    \hat{\boldsymbol{x}}_\tau = \frac{\alpha_\tau}{\alpha_{\tau'}}  \hat{\boldsymbol{x}}_{\tau'} -2 \sigma_\tau (e^{h}\!-\!1) \boldsymbol{\epsilon}_\phi( \hat{\boldsymbol{x}}_{\tau'},{\tau'}) - \sigma_\tau (e^{h}\!-\!1) \frac{\boldsymbol{\epsilon}_\phi(\hat{\boldsymbol{x}}_r,r)-\boldsymbol{\epsilon}_\phi(\hat{\boldsymbol{x}}_{\tau'},{\tau'})}{r_1} + \sigma_\tau \sqrt{e^{2h}-1} \boldsymbol{z}
    \label{DPM2}
\end{align}
where $r_1=\frac{\lambda_r-\lambda_{\tau'}}{h}$  and $\lambda_r,\lambda_{\tau'}$ denotes the parameter represents the log signal-to-noise-ratio.  

Diffusion models can be applied to perform guided sampling for a wide range of applications~\citep{dhariwal2021diffusion,rombach2022high,ramesh2022hierarchical,tumanyan2023plug}.   Typically, guided sampling relies on a conditional diffusion model $\hat{\boldsymbol{\epsilon}}_\phi(\boldsymbol{x}_\tau,\tau,c)$, and can be categorized as classifier guidance sampling and classifier-free sampling model based on whether they require a classifier model. However, these methods are not designed for black-box targeted generation and usually rely on joint training for the conditional model from scratch, which may be prohibitively expensive due to the re-training for each query batch update.


\subsection{Black-box optimization}

Given a proper function $f({\boldsymbol{x}}): \mathbb{R}^d \rightarrow \mathbb{R}$ such that $f({\boldsymbol{x}}) > -\infty$, black-box optimization is to minimize $f({\boldsymbol{x}})$ by using function queries only.  Instead of optimizing the original problem directly, 
ES or stochastic zeroth-order optimization methods optimize  a relaxation of the problem $J({{\theta}}):=\mathbb{E}_{p({\boldsymbol{x}};{{\theta}})}[f({\boldsymbol{x}})]$ w.r.t. the parameter $\theta$ of the sampling distribution of the relaxed problem.


Evolution strategies~\citep{rechenberg1970optimierung,nesterov2017random,liu2018zeroth} employ a Gaussian distribution $\mathcal{N}(\boldsymbol{\mu},\sigma^2\boldsymbol{I})$ with a fixed variance  for candidate sampling. The approximate gradient descent update is given as 
\begin{align}
    \boldsymbol{\mu}_{t+1} = \boldsymbol{\mu}_{t} -  \frac{\beta}{N\sigma}\sum  _{i=1}^N \boldsymbol{\epsilon}_i f( \boldsymbol{\mu}_t + \sigma \boldsymbol{\epsilon}_i ) , 
\end{align}
where $ \boldsymbol{\epsilon}_i \sim \mathcal{N}(\boldsymbol{0},\boldsymbol{I})$ and $\beta$ denotes the step-size, and $\boldsymbol{\mu}_{t}$ denotes the mean  parameter of the  Gaussian distribution for candidate sampling at $t^{th}$ black-box optimization iteration.

The ES methods only perform the first-order approximate gradient update, the convergence speed is limited.  \cite{NES} proposed the natural evolution strategies (NES), which perform the approximate natural gradient update, in which A Gaussian distribution $\mathcal{N}(\boldsymbol{\mu},\boldsymbol{\Sigma})$ is employed for sampling. Besides the updating of parameter $\boldsymbol{\mu}$, the covariance matrix $\boldsymbol{\Sigma})$ is also updated. \cite{lyu2021black} proposed an implicit natural gradient optimizer (INGO) for black-box optimization, which provides an alternative way to compute the natural gradient update. In INGO update rule, the inverse covariance matrix $\boldsymbol{\Sigma}^{-1}$ is updated instead of the  covariance matrix $\boldsymbol{\Sigma}$. Moreover,  the Covariance Matrix Adaptation Evolution Strategy (CMA-ES)~\citep{hansen2001completely,hansen2006cma}  optimizes the objective function by adaptively fitting a multivariate Gaussian distribution, guiding the search for the optimum based on the covariance matrix.
CMAES~\citep{hansen2006cma} provides a more sophisticated update rule and performs well on a wide range of black-box optimization problems.  The update scheme combines several update rules, including heuristic updates, that are challenging to analyze the convergence theoretically. \cite{lyu2023fast} proposed to employ local GP with closed-form fast rank-1 lattice targeted sampling to accelerate the convergence of INGO. \cite{feiyang2023adaptive} study the black-box multi-objective optimization by Gaussian sampling with the diagonal covariance matrix assumption.   
Despite the success of these methods, all these methods ignore the dynamic transition of the target function.

Bayesian Optimization (BO) usually builds a global Gaussian Process (GP) model as a surrogate and provides queries by optimizing some acquisition functions~\citep{Nips2012practical}. Although BO achieves good query efficiency for low-dimensional problems, it often fails to handle high-dimensional problems with large sample budgets~\citep{eriksson2019scalable}. The computation of GP with a large number of samples itself is expensive, and the internal optimization of the acquisition functions is challenging.  The Trust Region Bayesian Optimization (TuRBO)~\citep{eriksson2019scalable} extends traditional Bayesian optimization  to solve global optimization problems by simultaneously running multiple independent local Gaussian processes. 
Recently,  \cite{muller2021local,nguyen2022local} builds a GP model for both the function value and the gradient and performs local Bayesian optimization. Although these methods improve the scalability of global BO, they usually cannot scale up to five hundred dimensional complex problems.  This may be because the learned gradient heavily depends on the accuracy of the GP model.  However, achieving an accurate GP model is challenging for high-dimensional problems. A slightly misspecified GP model may lead to a wrong estimated gradient due to the highly nonlinear acquisition functions. 

The recent work~\citep{krishnamoorthy2023diffusion} introduces Denoising Diffusion Optimization Models (DDOM) for solving offline black-box optimization tasks using diffusion models. This method can also be naturally extended to black-box targeted generation tasks. The DDOM relies on an offline conditional model trained with reweighted data sampling.  The generation is performed conditioned on a high target score. The pre-collected data set has a crucial influence on DDOM generation. 

\section{Experiments Compute Resources}
\label{compute_resources}
All experiments were conducted on a computer equipped with an AMD Ryzen Threadripper 3960X (24-Core) CPU, 260GB RAM, and 4 Nvidia RTX A6000 GPUs. For the diffusion fine-tuning experiment, we use AWS Lambda to parallelize the batch computation of the Vina docking score. The memory allocation for AWS Lambda was set at 1536 MB on the x86\_64 architecture\footnote{https://docs.aws.amazon.com/lambda/latest/operatorguide/computing-power.html}, and each function evaluation took approximately 800ms. The numerical experiments required less than 30 minutes per run, whereas the fine-tuning experiments (300 steps) required about 300 minutes per run.

\section{Broader Impacts}
\label{Broader_Impacts}

The proposed Covariance-adaptive sequential black-box optimization algorithm is a foundational research focused on black-box optimization and theoretical convergence analysis.  This part may not have a direct social impact.  The proposed targeted SDE fine-tuning framework for diffusion black-box targeted generation is one of the downstream real-world tasks that can be optimized by our optimization algorithm.  This part may have a potential social impact.  The proposed diffusion black-box targeted generation method may be used in 3D-molecule targeted generation for drug discovery,  new material design for superconducting,   etc.


\section{Demonstration of the Generated 3D-molecule}
\label{demo}
\begin{figure}[h]
    \centering
    \subfigure[\scriptsize{Pre-train Best (Score = -11.082)}]{
    \label{pretrainBest0}
    \includegraphics[width=0.32\linewidth]{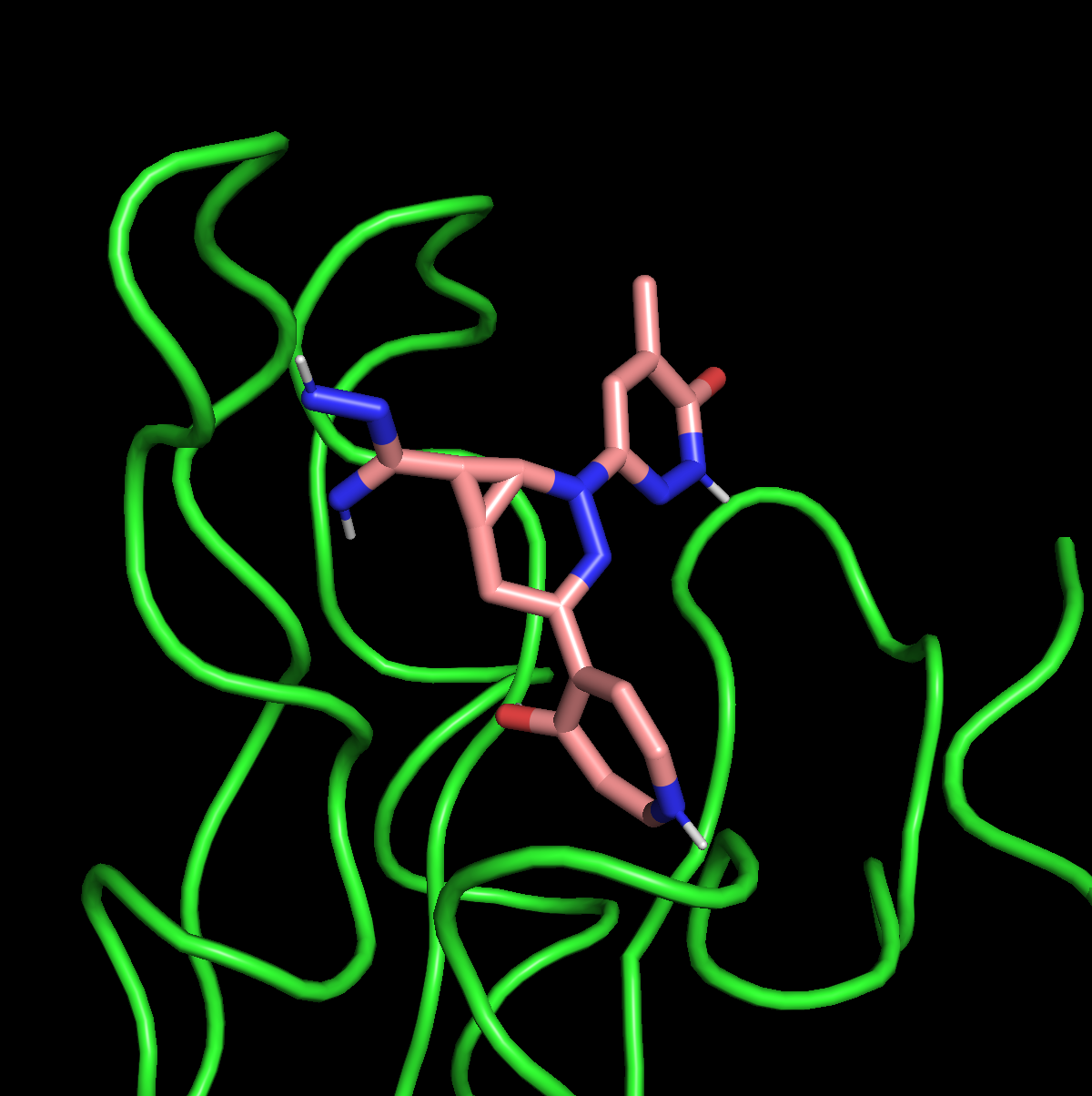}}
    \subfigure[\scriptsize{Before finetune (Score = -8.092)}]{
    \label{noFinetune0}
    \includegraphics[width=0.32\linewidth]{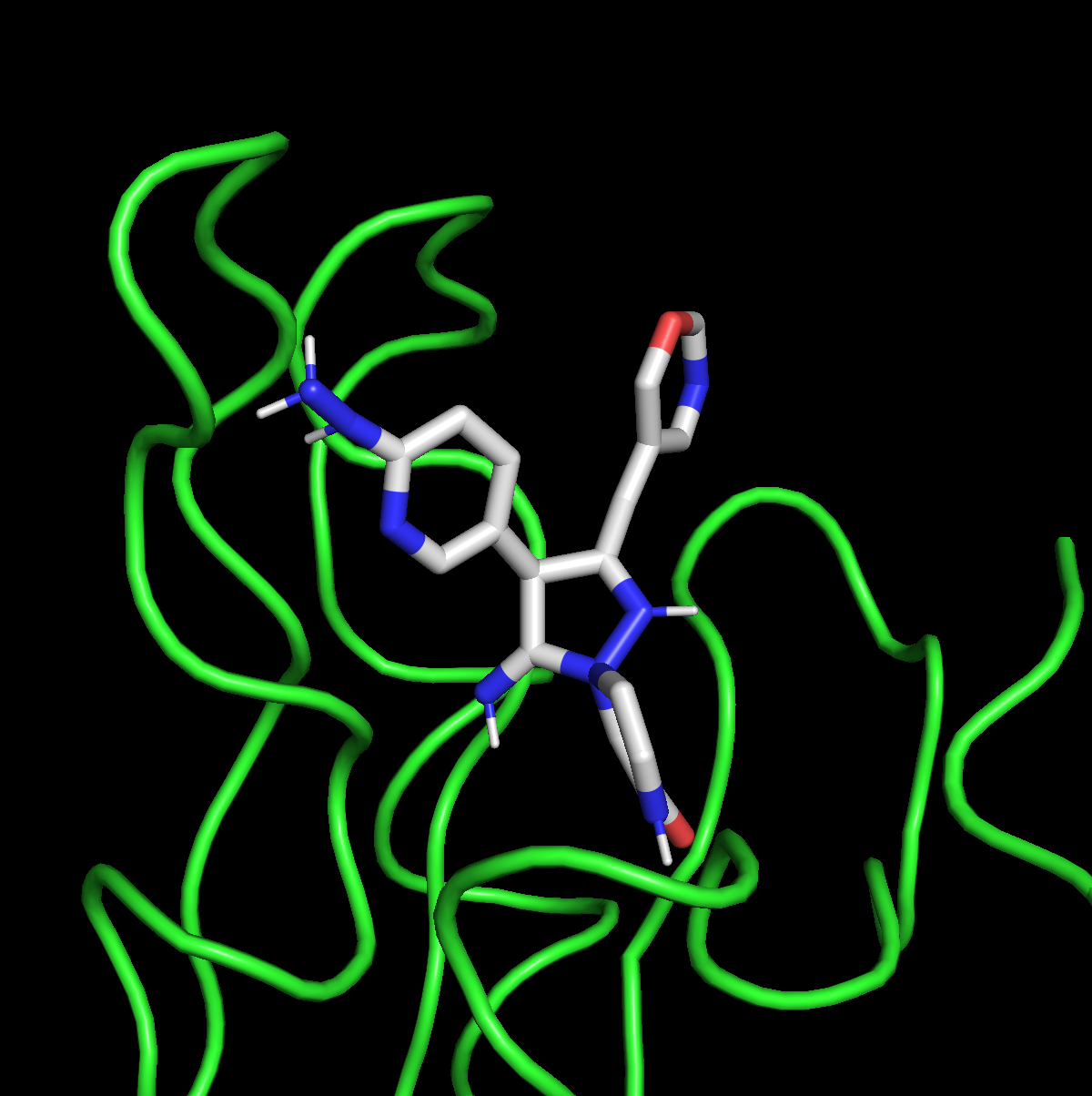}}
    \\
    \subfigure[\scriptsize{BDTG (ours) (Score = -12.431)}]{
    \label{ours0}
    \includegraphics[width=0.32\linewidth]{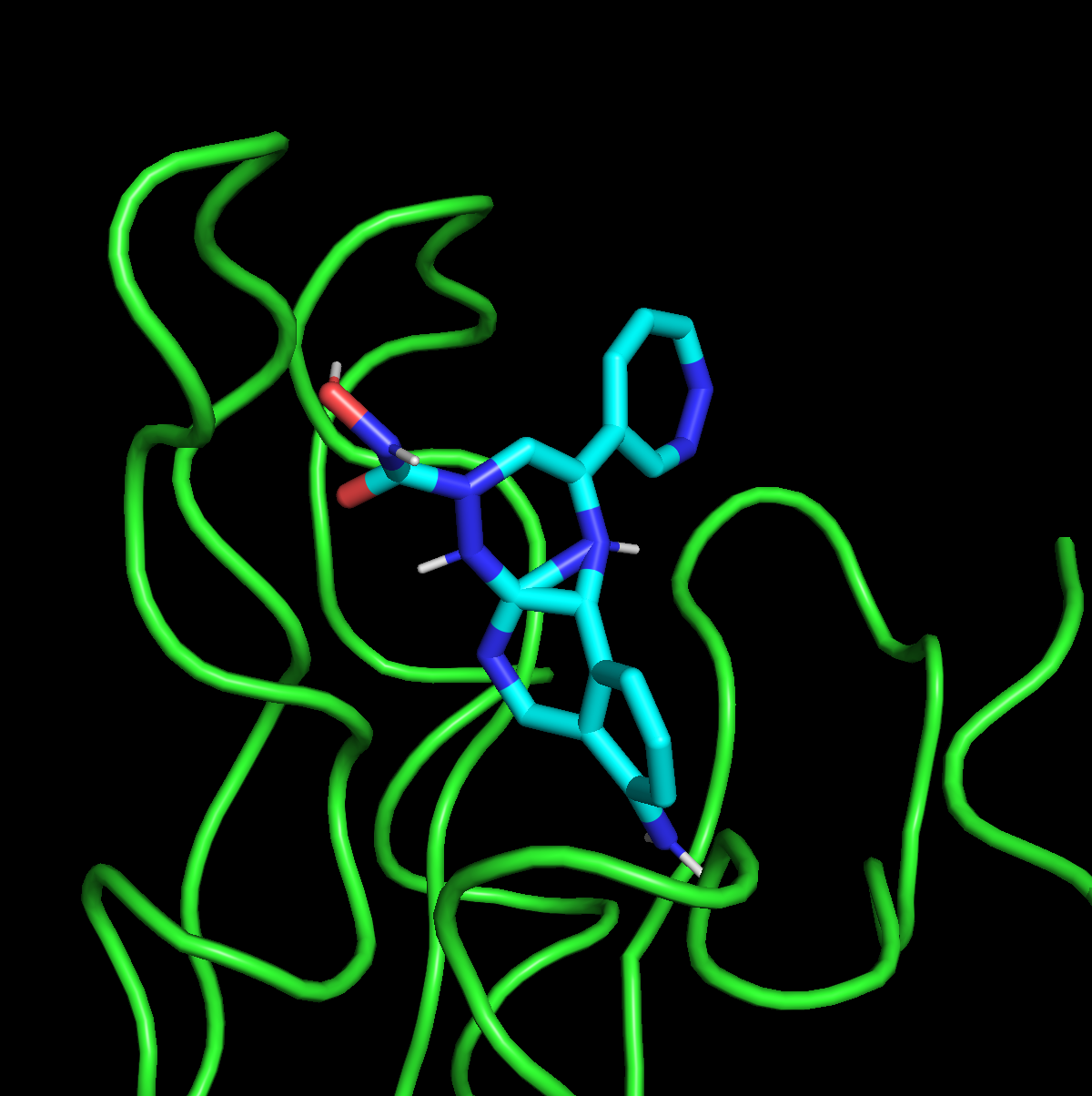}}
    \subfigure[\scriptsize{TuRBO (Score = -10.653)}]{
    \label{bo0}
    \includegraphics[width=0.32\linewidth]{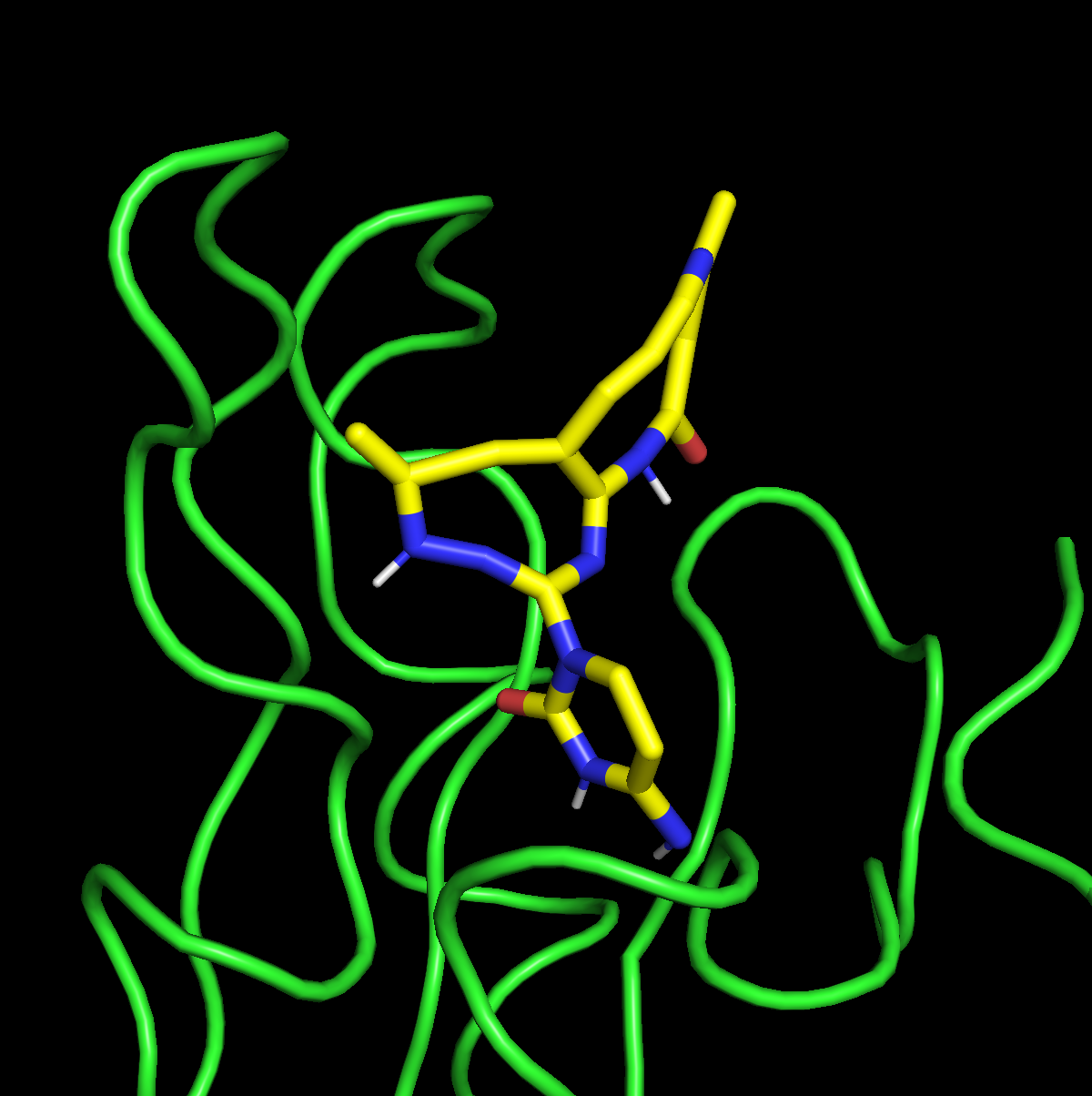}}
    \subfigure[\scriptsize{CMAES (Score = -10.210)}]{
    \label{cma0}
    \includegraphics[width=0.32\linewidth]{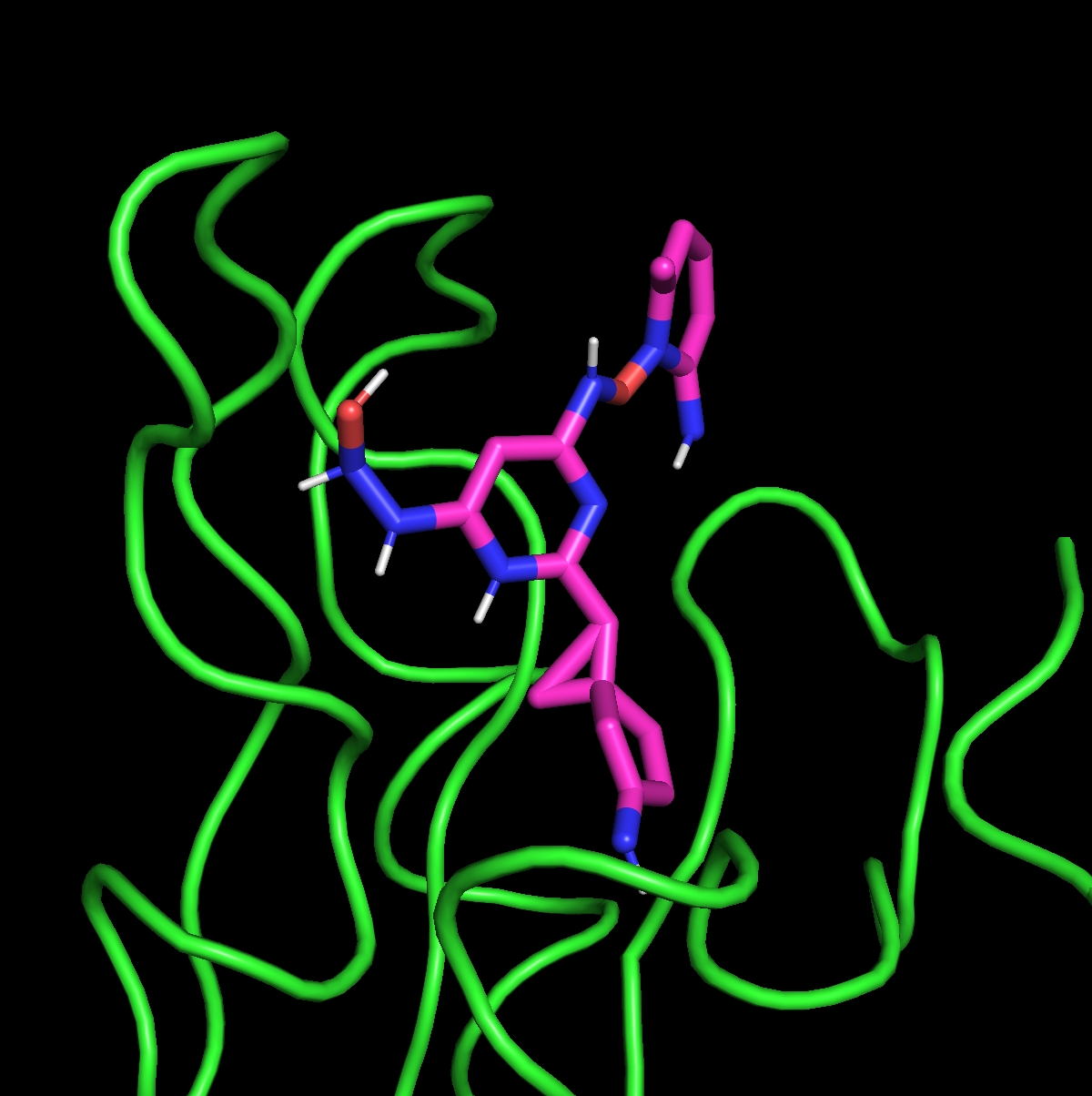}}
    \caption{Demonstration of the Generated 3D-molecule on Receptor-0}
    \label{Demo0}
\end{figure}

\begin{figure}[h]
    \centering
    \subfigure[\scriptsize{Pre-train Best (Score = -8.834)}]{
    \label{pretrainBest1}
    \includegraphics[width=0.32\linewidth]{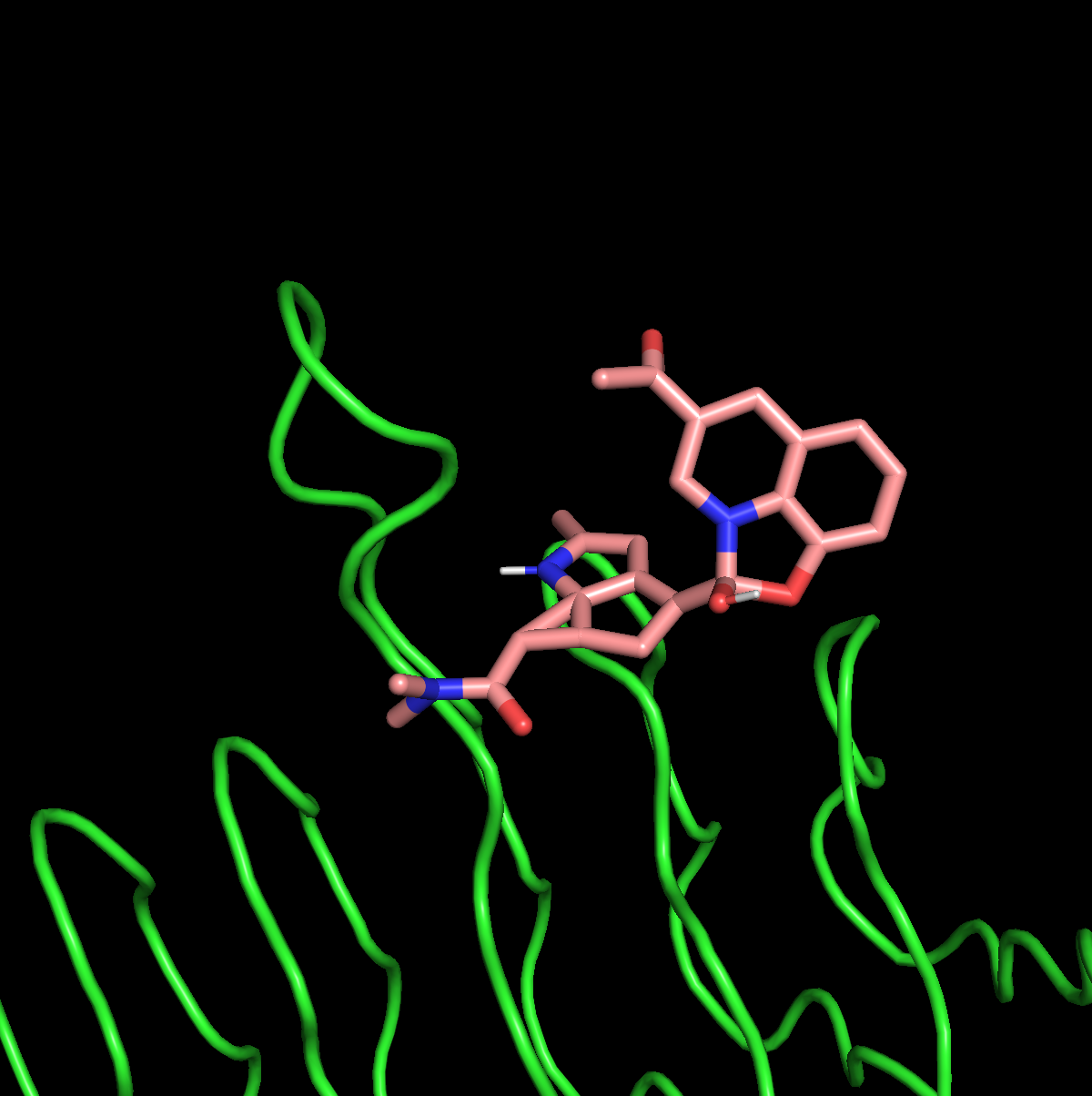}}
    \subfigure[\scriptsize{Before finetune (Score = -10.230)}]{
    \label{noFinetune1}
    \includegraphics[width=0.32\linewidth]{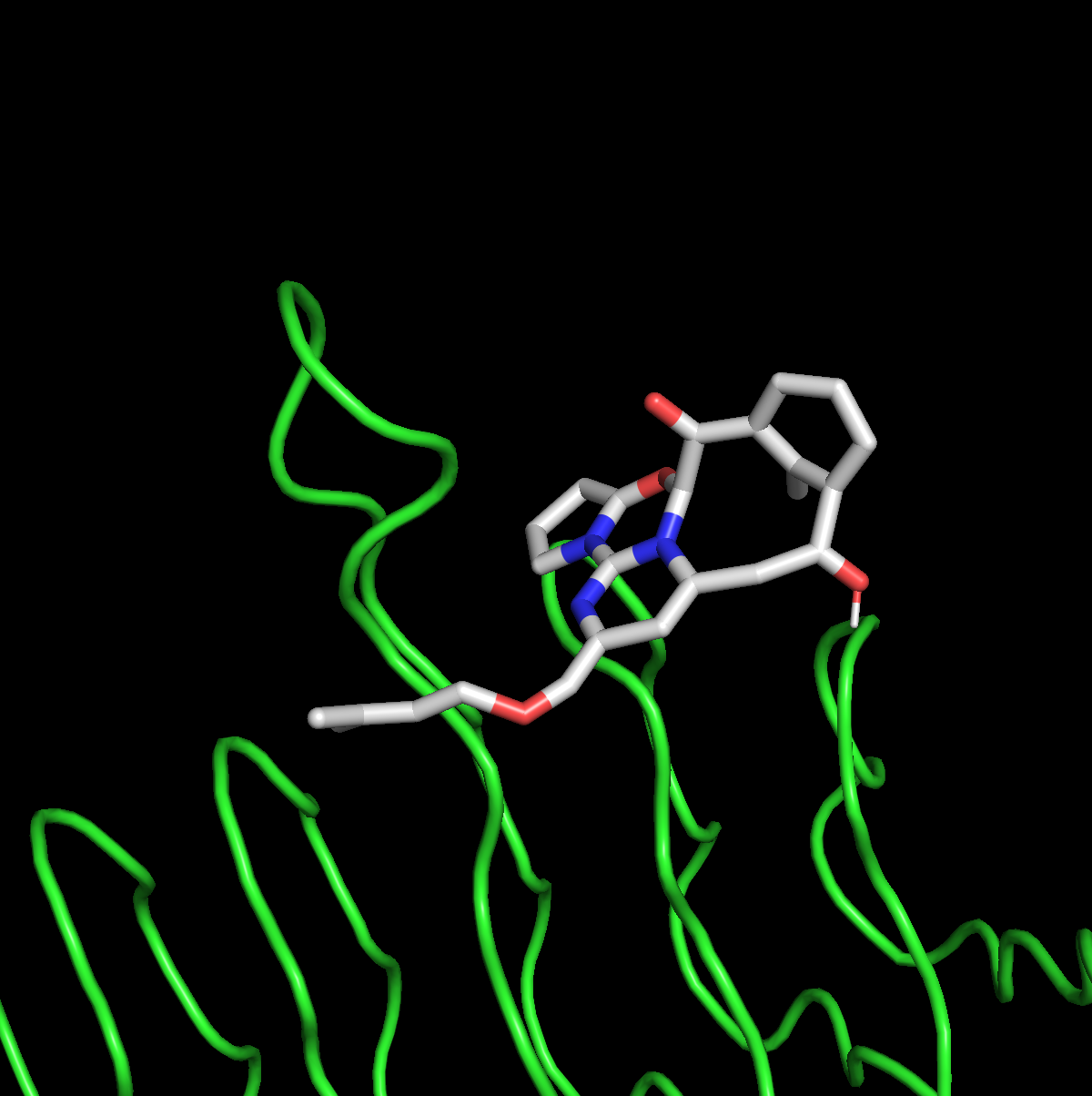}}
    \\
    \subfigure[\scriptsize{BDTG (ours) (Score = -12.740)}]{
    \label{ours1}
    \includegraphics[width=0.32\linewidth]{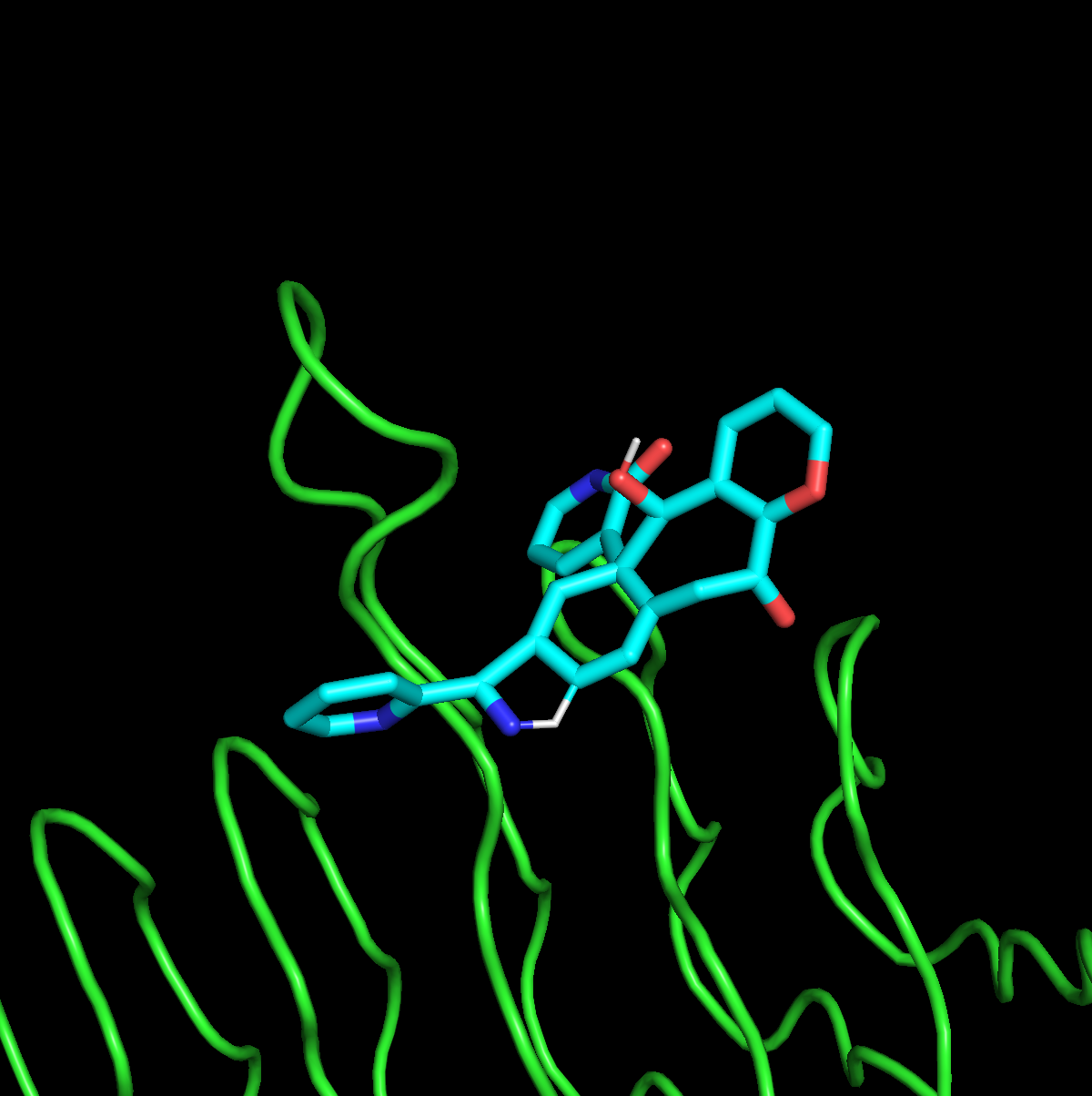}}
    \subfigure[\scriptsize{TuRBO (Score = -11.203)}]{
    \label{bo1}
    \includegraphics[width=0.32\linewidth]{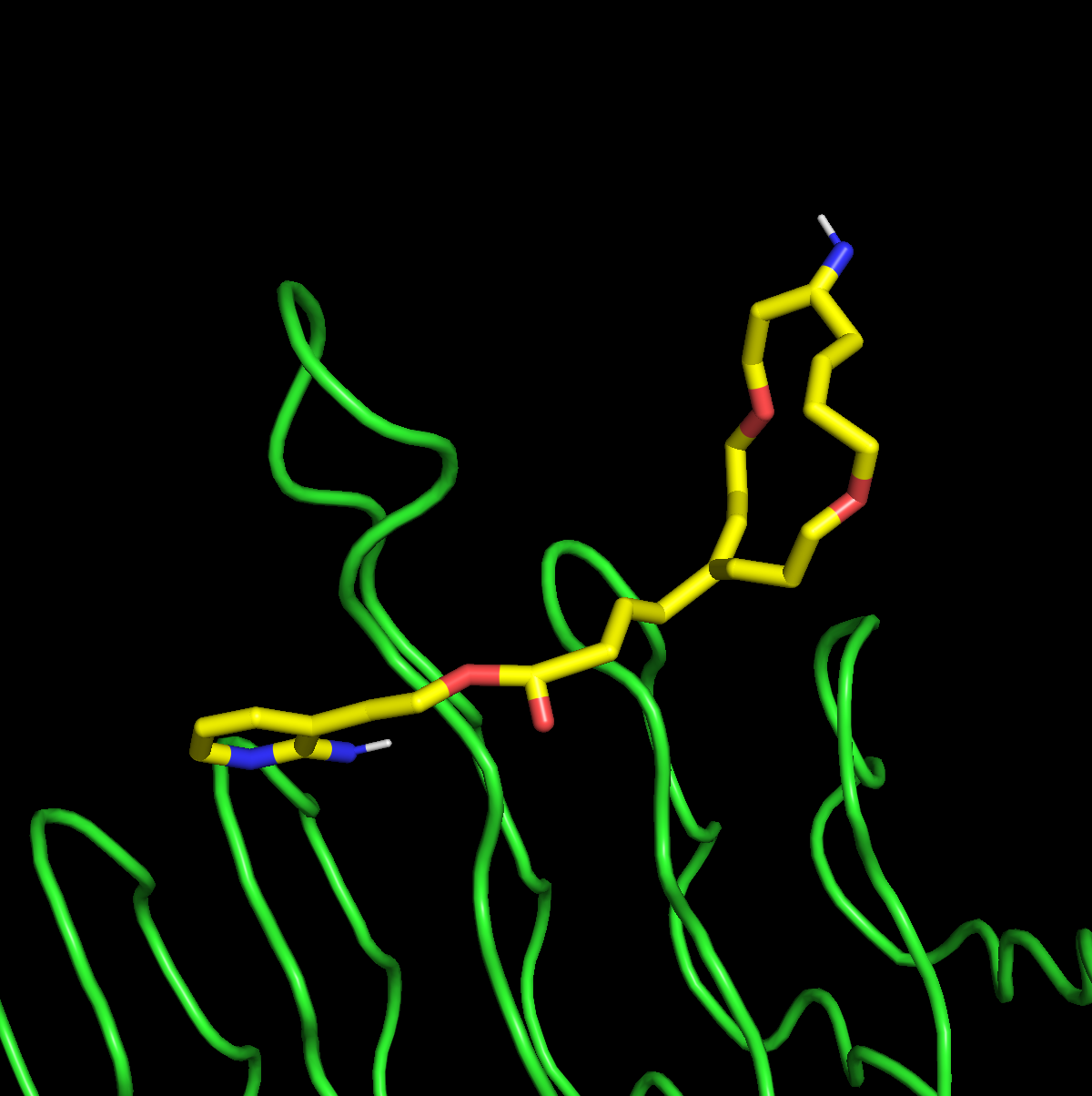}}
    \subfigure[\scriptsize{CMAES (Score = -10.192)}]{
    \label{cma1}
    \includegraphics[width=0.32\linewidth]{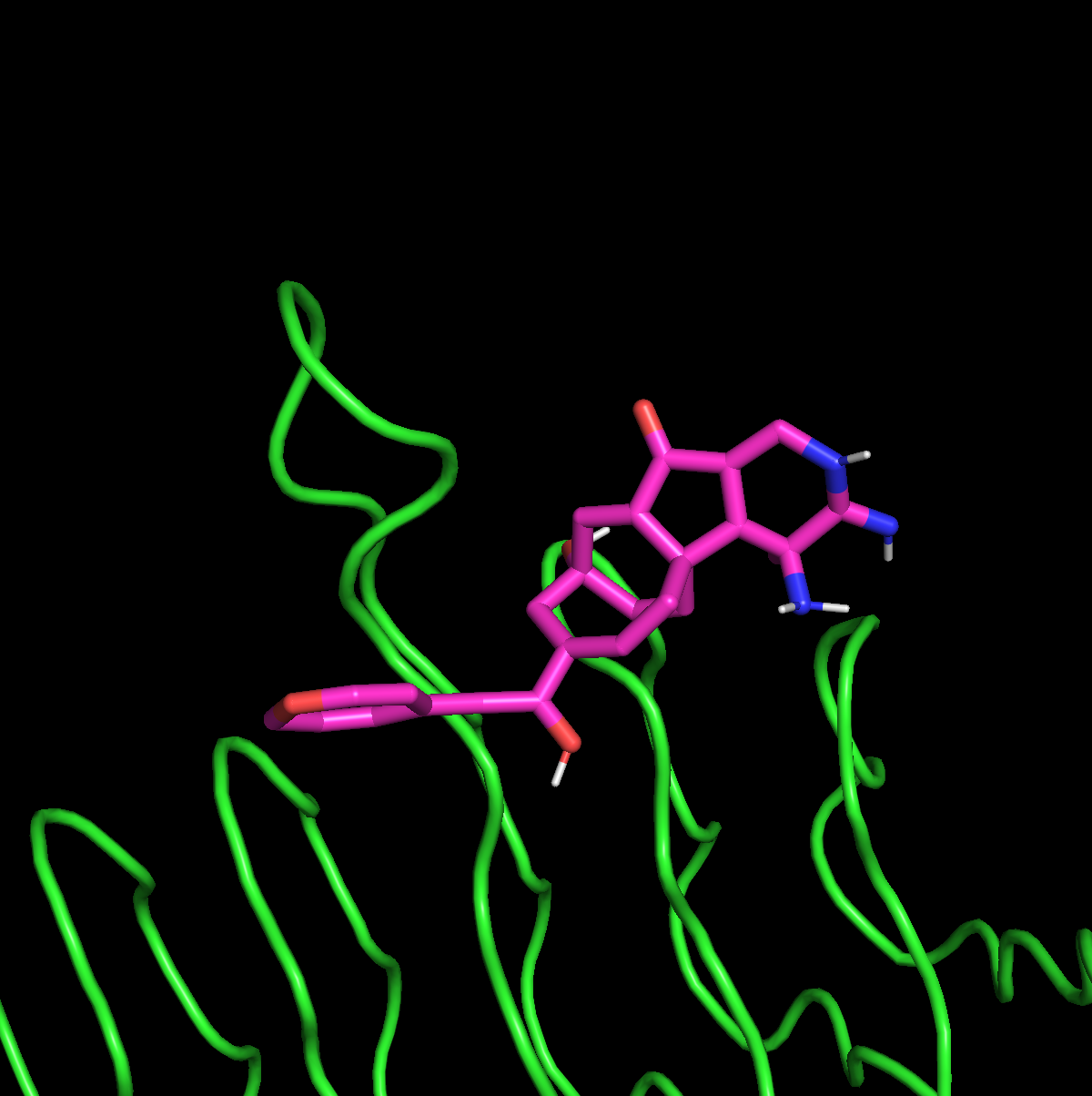}}
    \caption{Demonstration of the Generated 3D-molecule on Receptor-1}
    \label{Demo1}
\end{figure}

\begin{figure}[h]
    \centering
    \subfigure[\scriptsize{Pre-train Best (Score = -10.445)}]{
    \label{pretrainBest2}
    \includegraphics[width=0.32\linewidth]{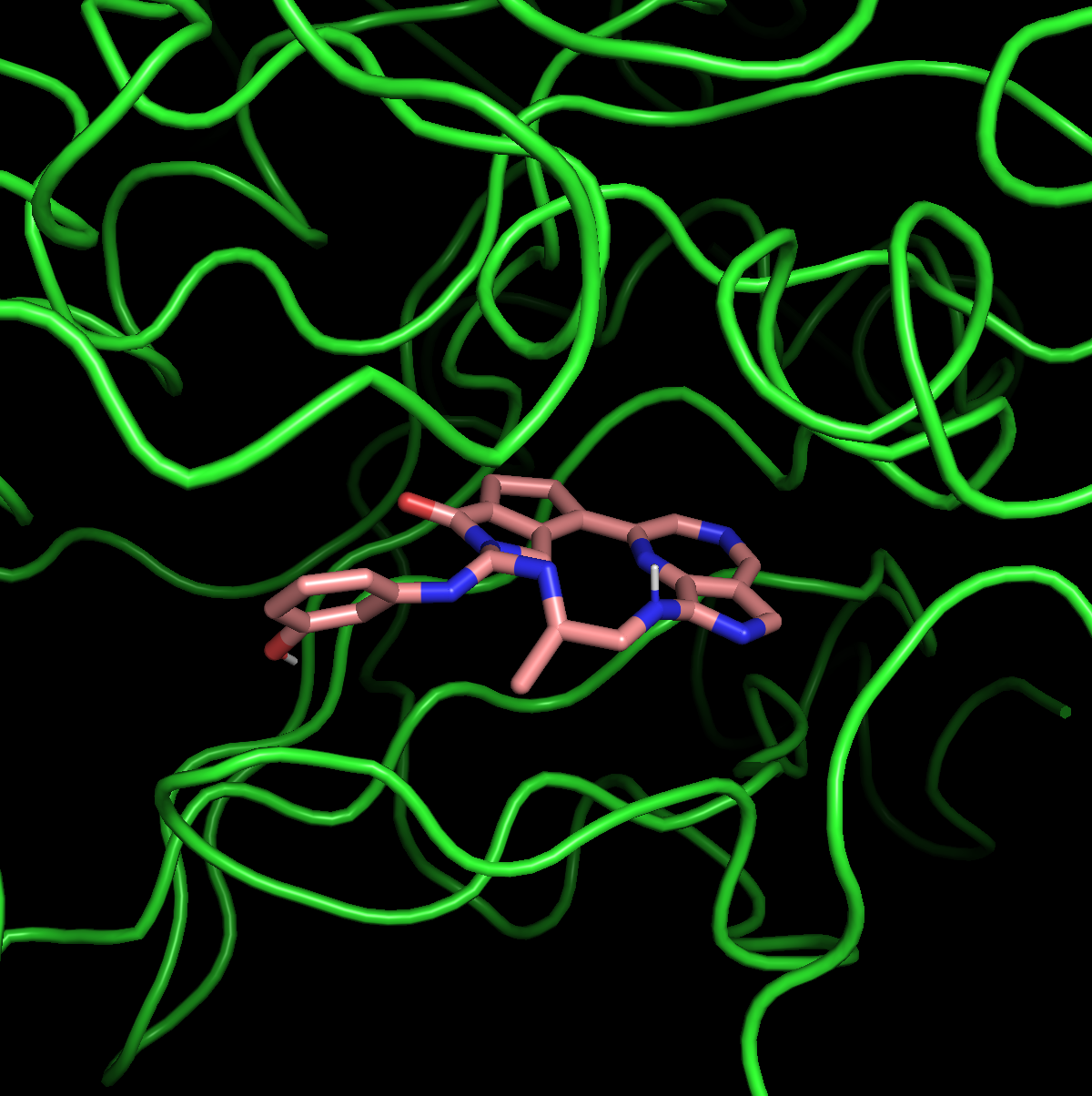}}
    \subfigure[\scriptsize{Before finetune (Score = -11.080)}]{
    \label{noFinetune2}
    \includegraphics[width=0.32\linewidth]{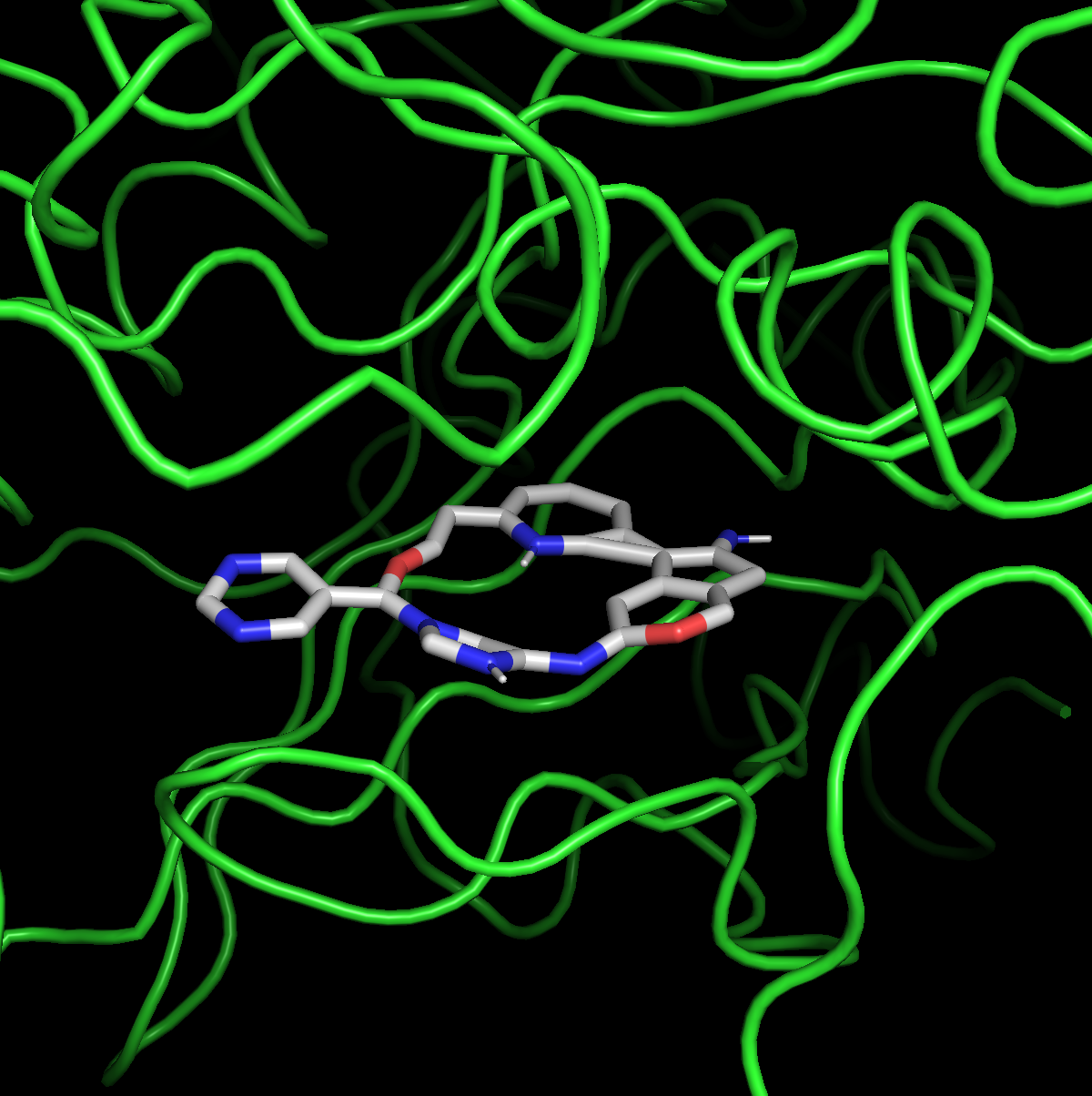}}
    \\
    \subfigure[\scriptsize{BDTG (ours) (Score = -12.999)}]{
    \label{ours2}
    \includegraphics[width=0.32\linewidth]{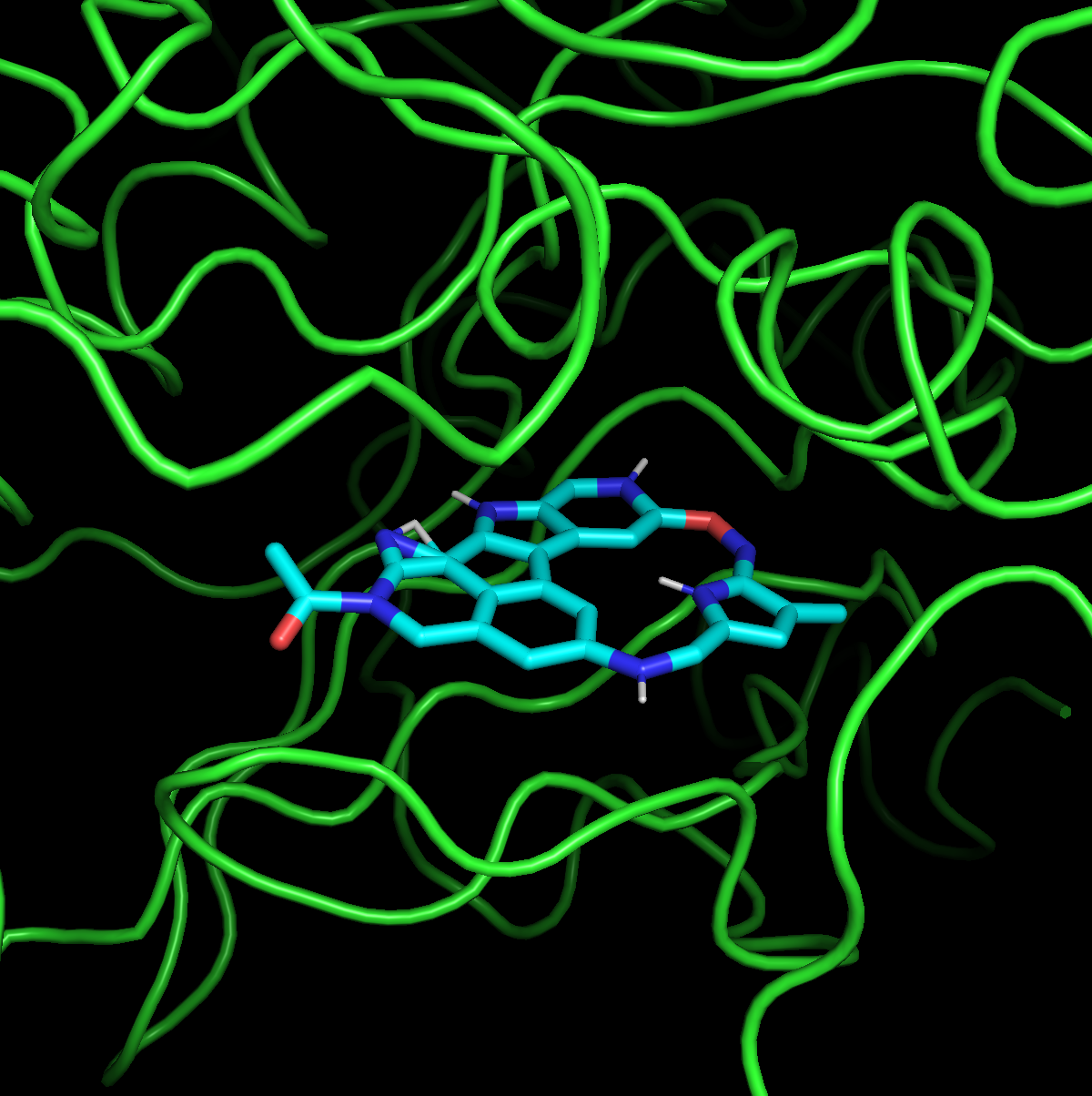}}
    \subfigure[\scriptsize{TuRBO (Score = -12.688)}]{
    \label{bo2}
    \includegraphics[width=0.32\linewidth]{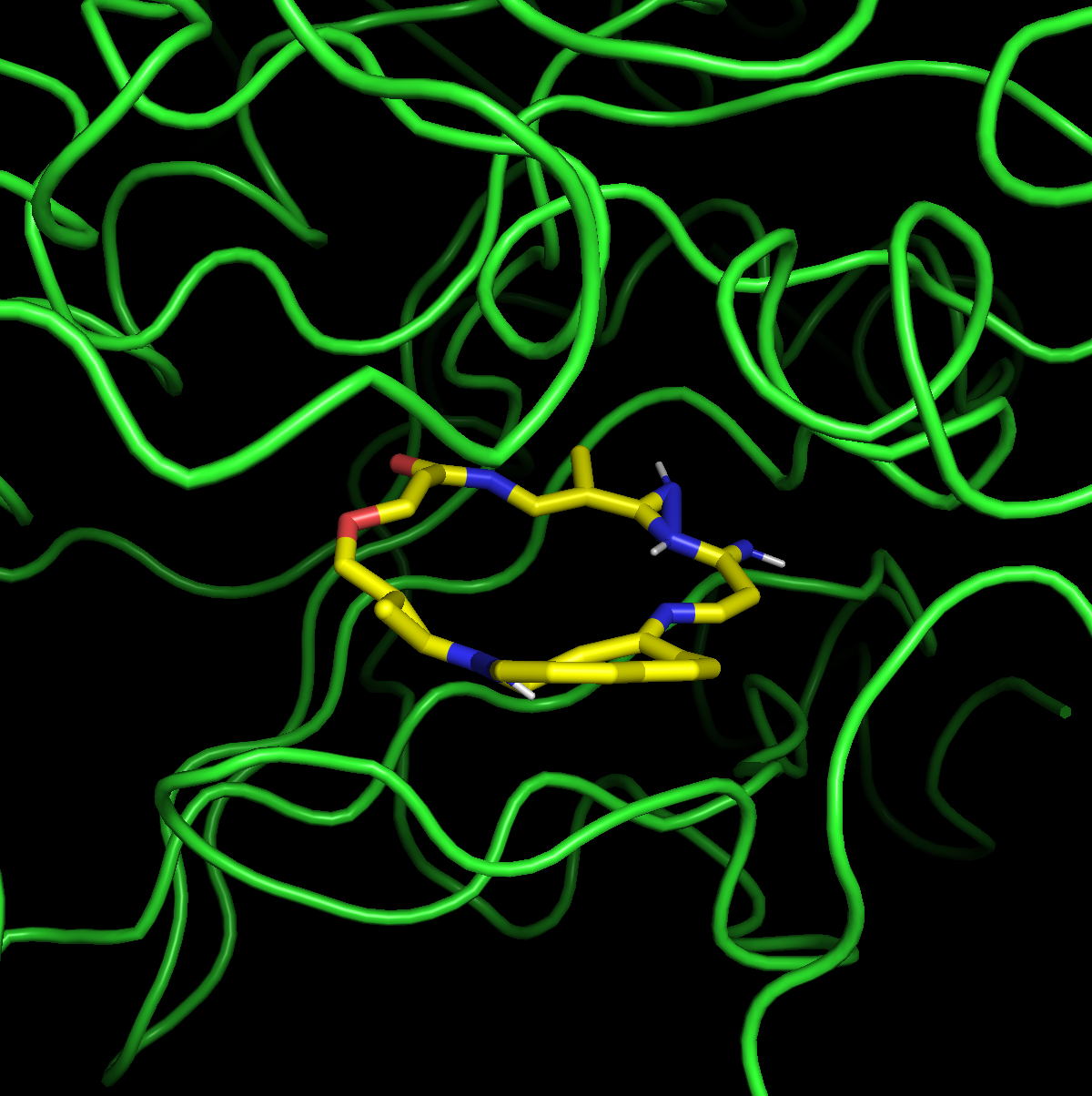}}
    \subfigure[\scriptsize{CMAES (Score = -10.579)}]{
    \label{cma2}
    \includegraphics[width=0.32\linewidth]{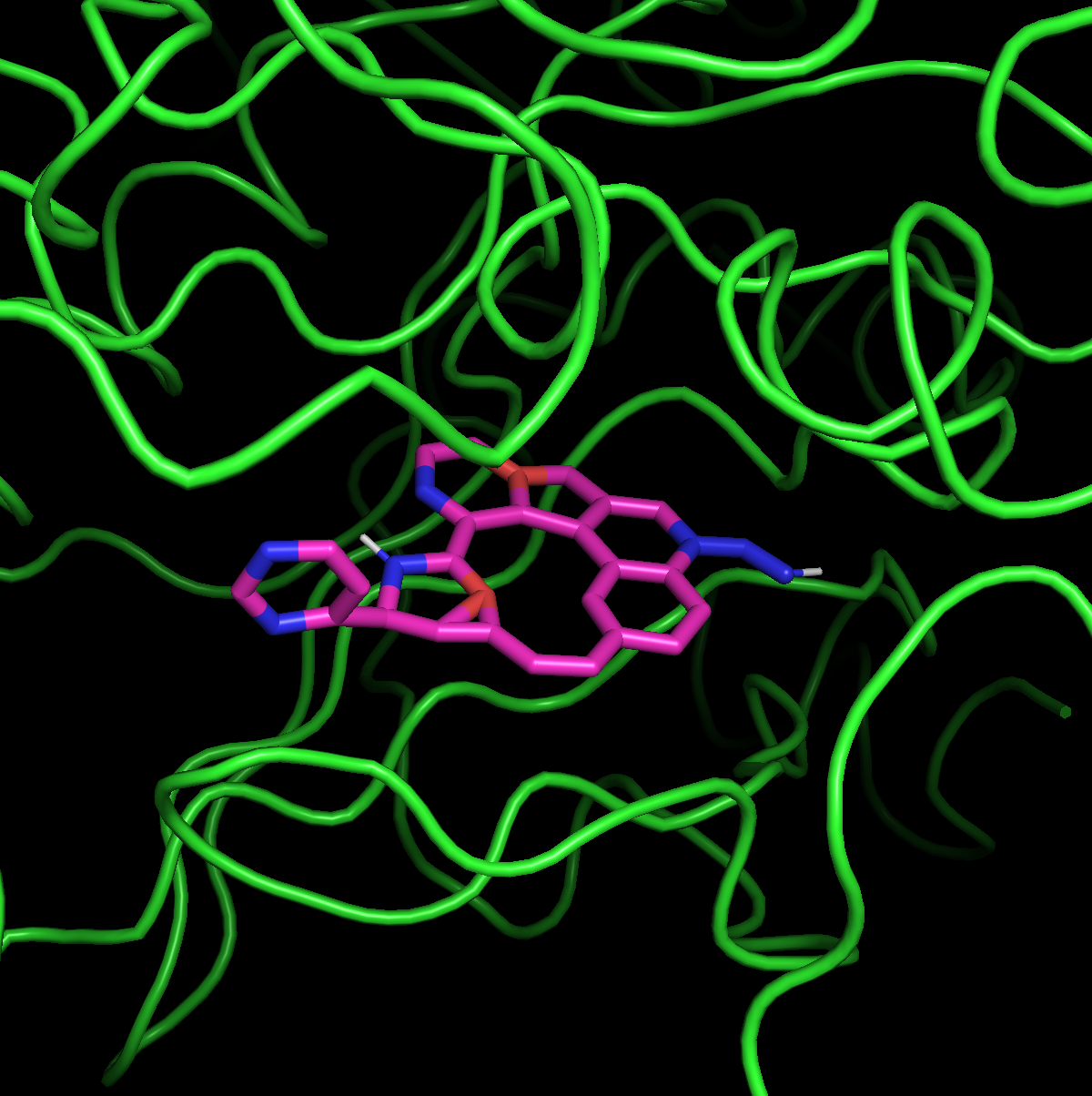}}
    \caption{Demonstration of the Generated 3D-molecule on Receptor-2}
    \label{Demo2}
\end{figure}

\begin{figure}[h]
    \centering
    \subfigure[\scriptsize{Pre-train Best (Score = -9.494)}]{
    \label{pretrainBest3}
    \includegraphics[width=0.32\linewidth]{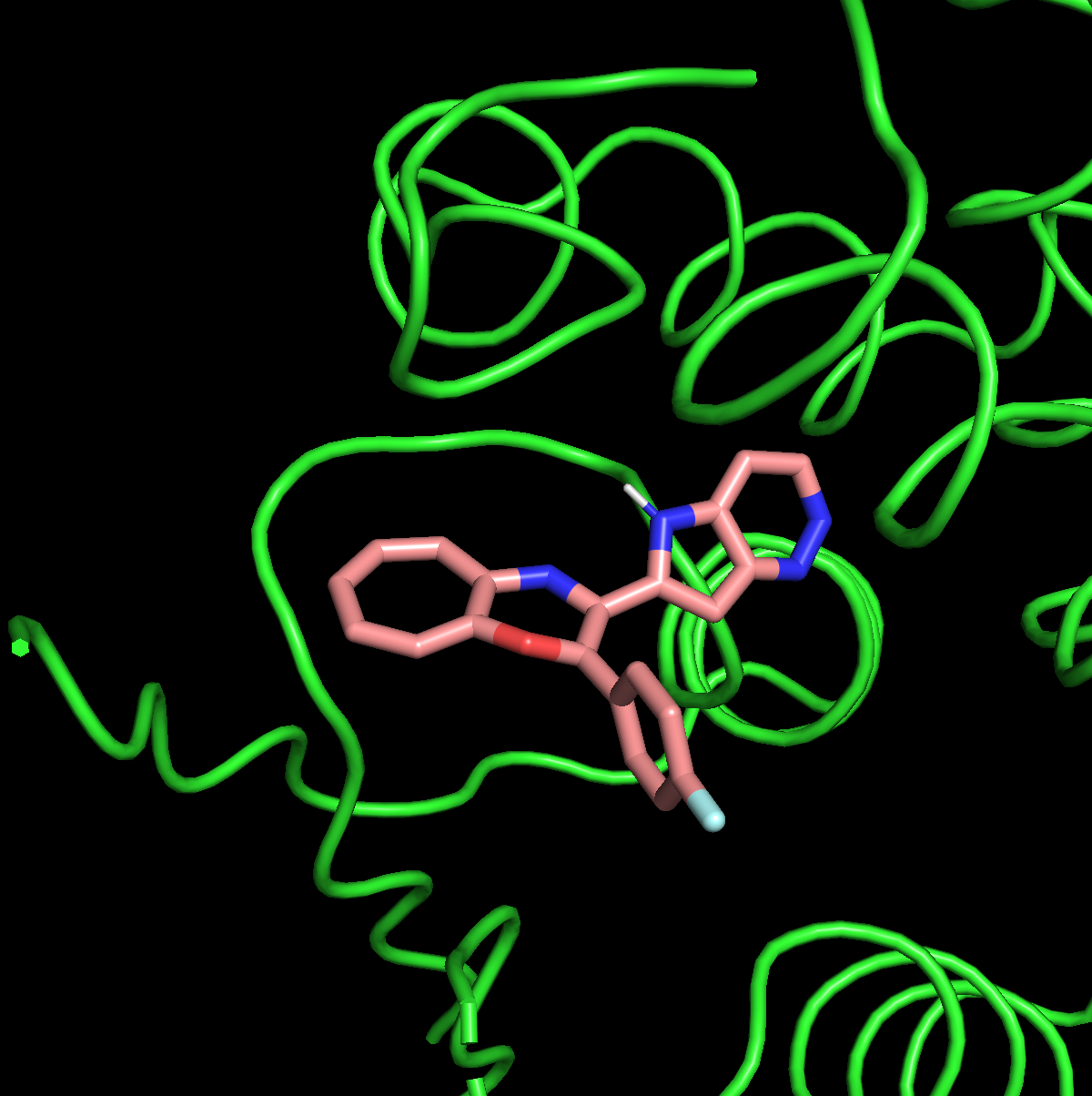}}
    \subfigure[\scriptsize{Before finetune (Score = -9.136)}]{
    \label{noFinetune3}
    \includegraphics[width=0.32\linewidth]{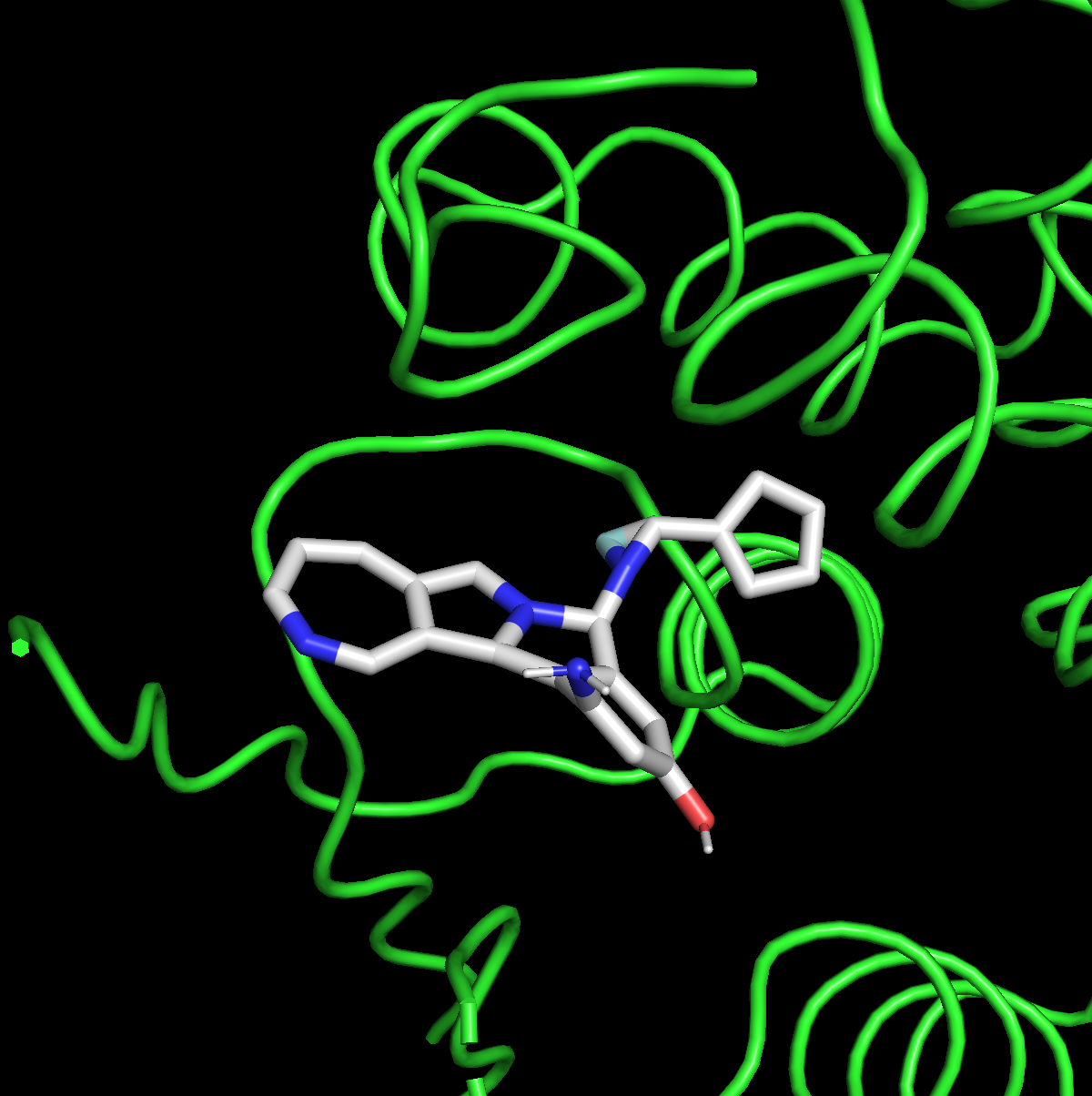}}
    \\
    \subfigure[\scriptsize{BDTG (ours) (Score = -12.291)}]{
    \label{ours3}
    \includegraphics[width=0.32\linewidth]{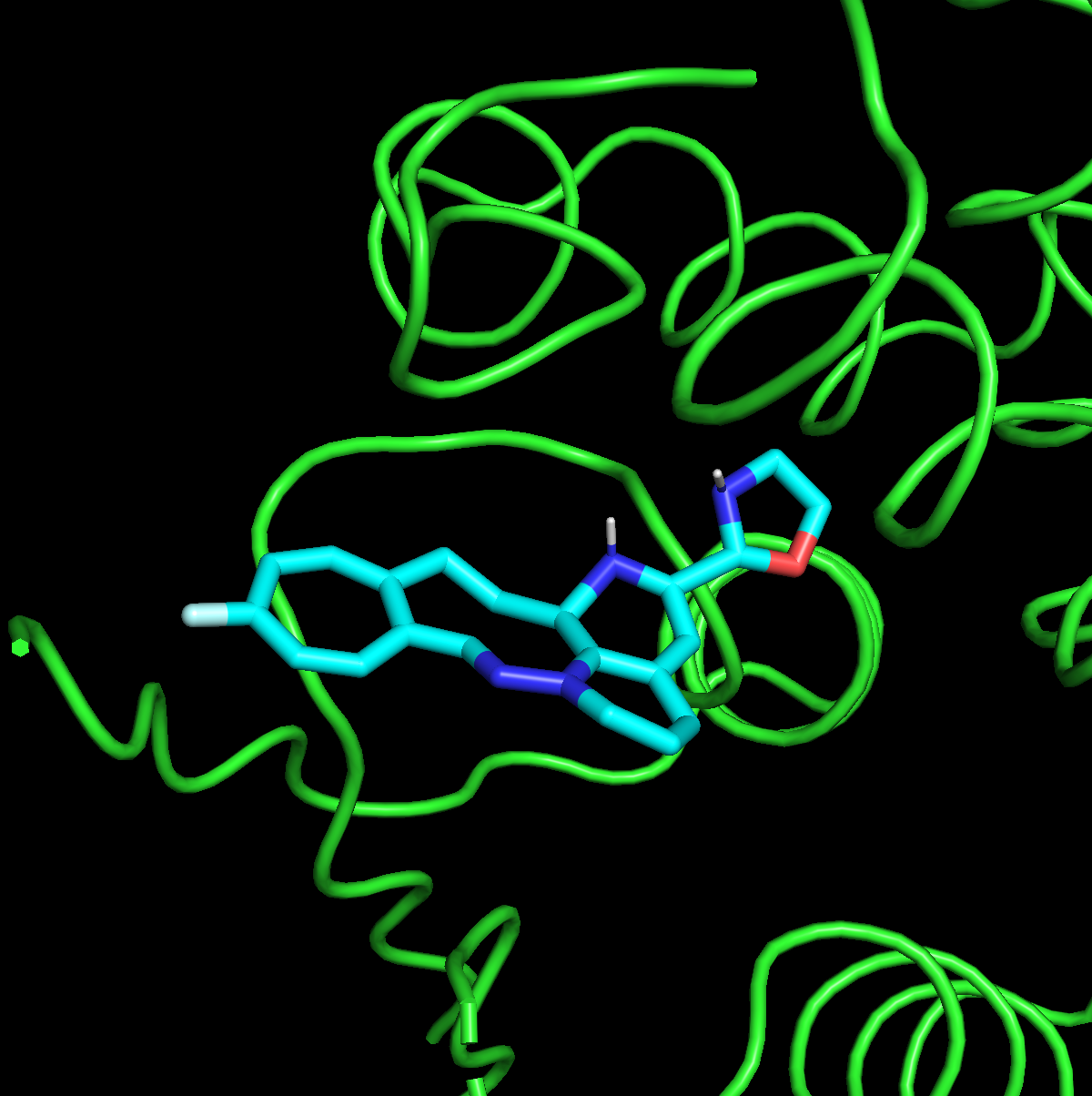}}
    \subfigure[\scriptsize{TuRBO (Score = -8.121)}]{
    \label{bo3}
    \includegraphics[width=0.32\linewidth]{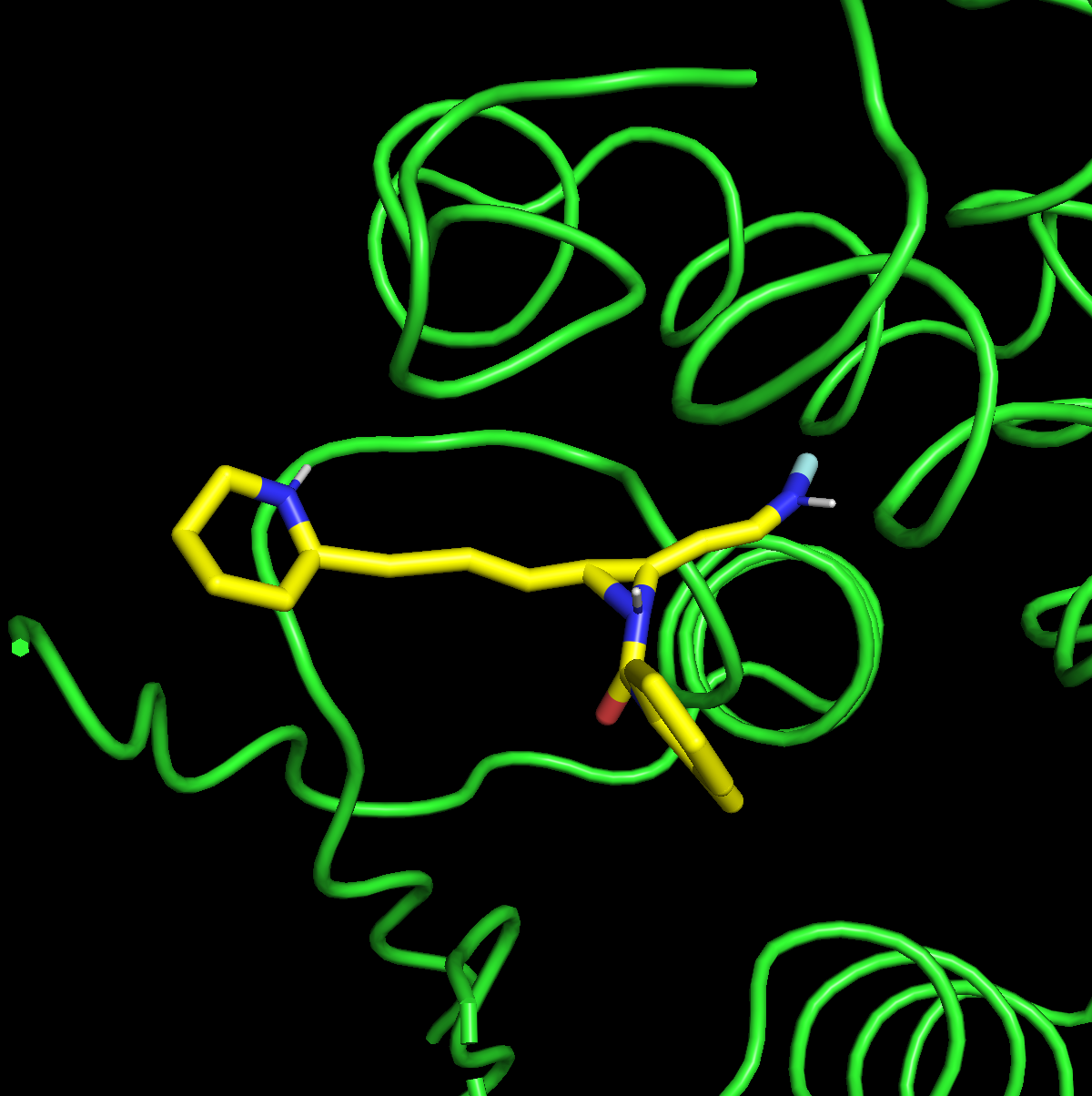}}
    \subfigure[\scriptsize{CMAES (Score = -9.838)}]{
    \label{cma3}
    \includegraphics[width=0.32\linewidth]{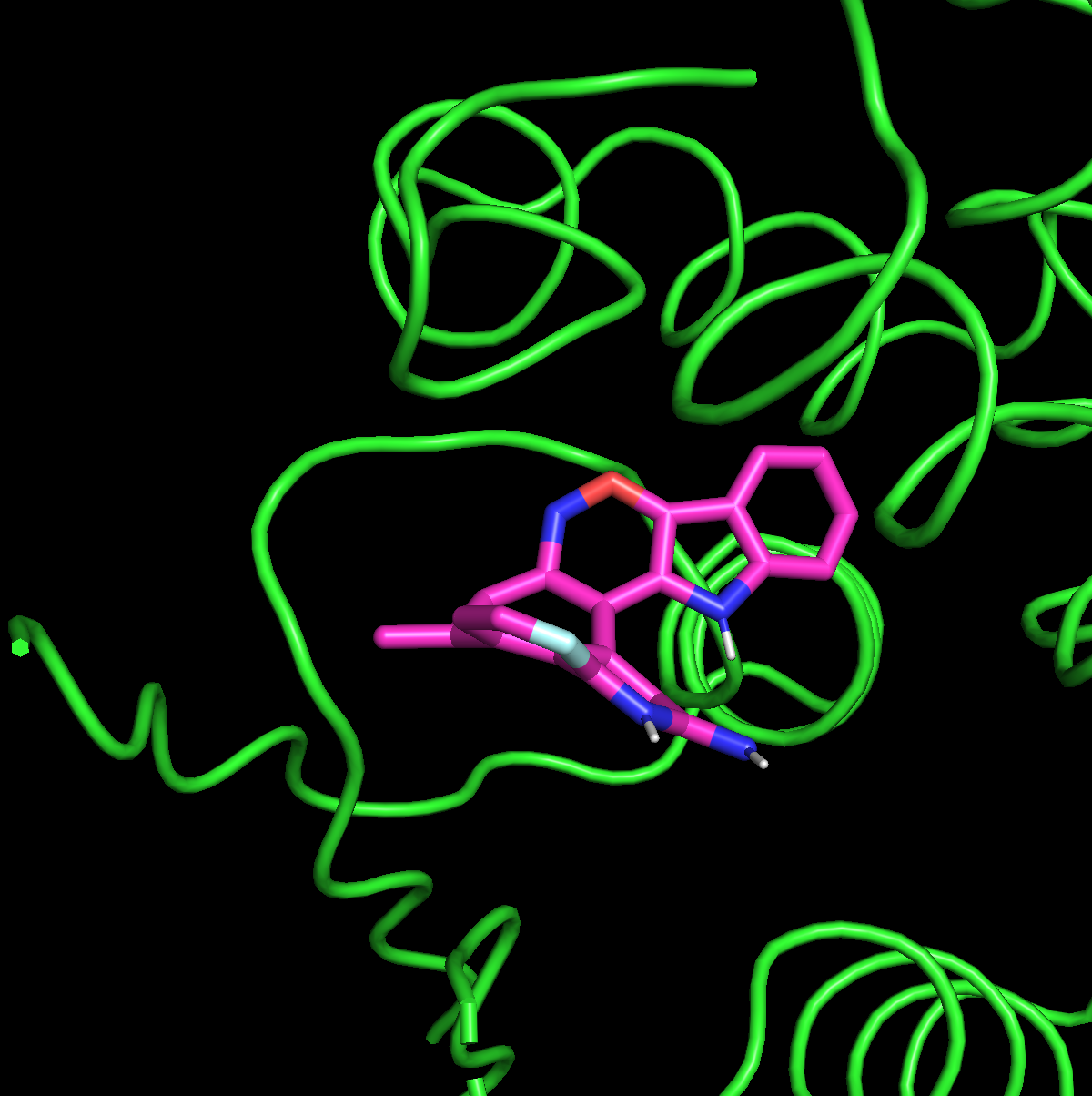}}
    \caption{Demonstration of the Generated 3D-molecule on Receptor-3}
    \label{Demo3}
\end{figure}

\begin{figure}[h]
    \centering
    \subfigure[\scriptsize{Pre-train Best (Score = -15.998)}]{
    \label{pretrainBest4}
    \includegraphics[width=0.32\linewidth]{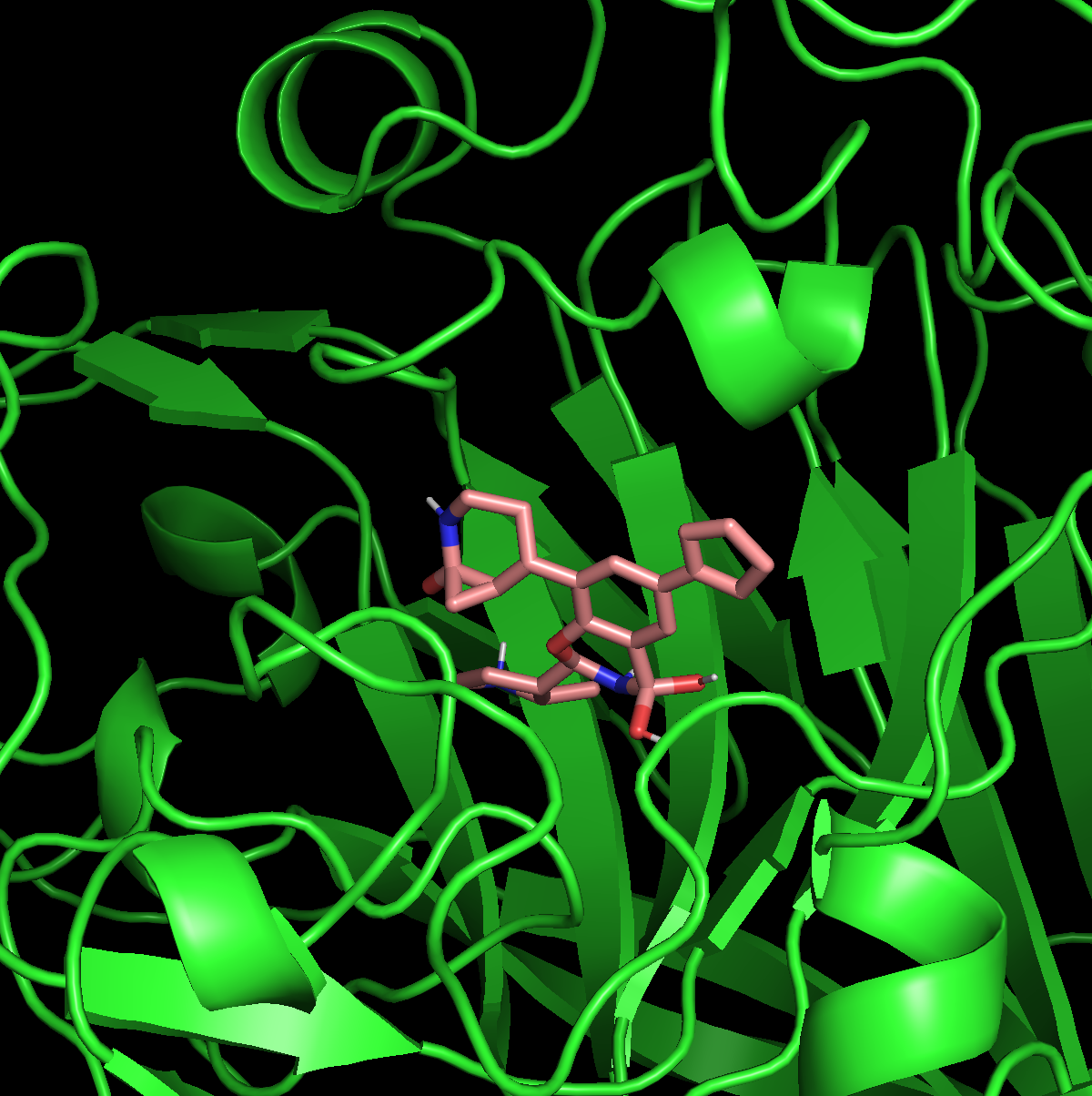}}
    \subfigure[\scriptsize{Before finetune (Score = -16.460)}]{
    \label{noFinetune4}
    \includegraphics[width=0.32\linewidth]{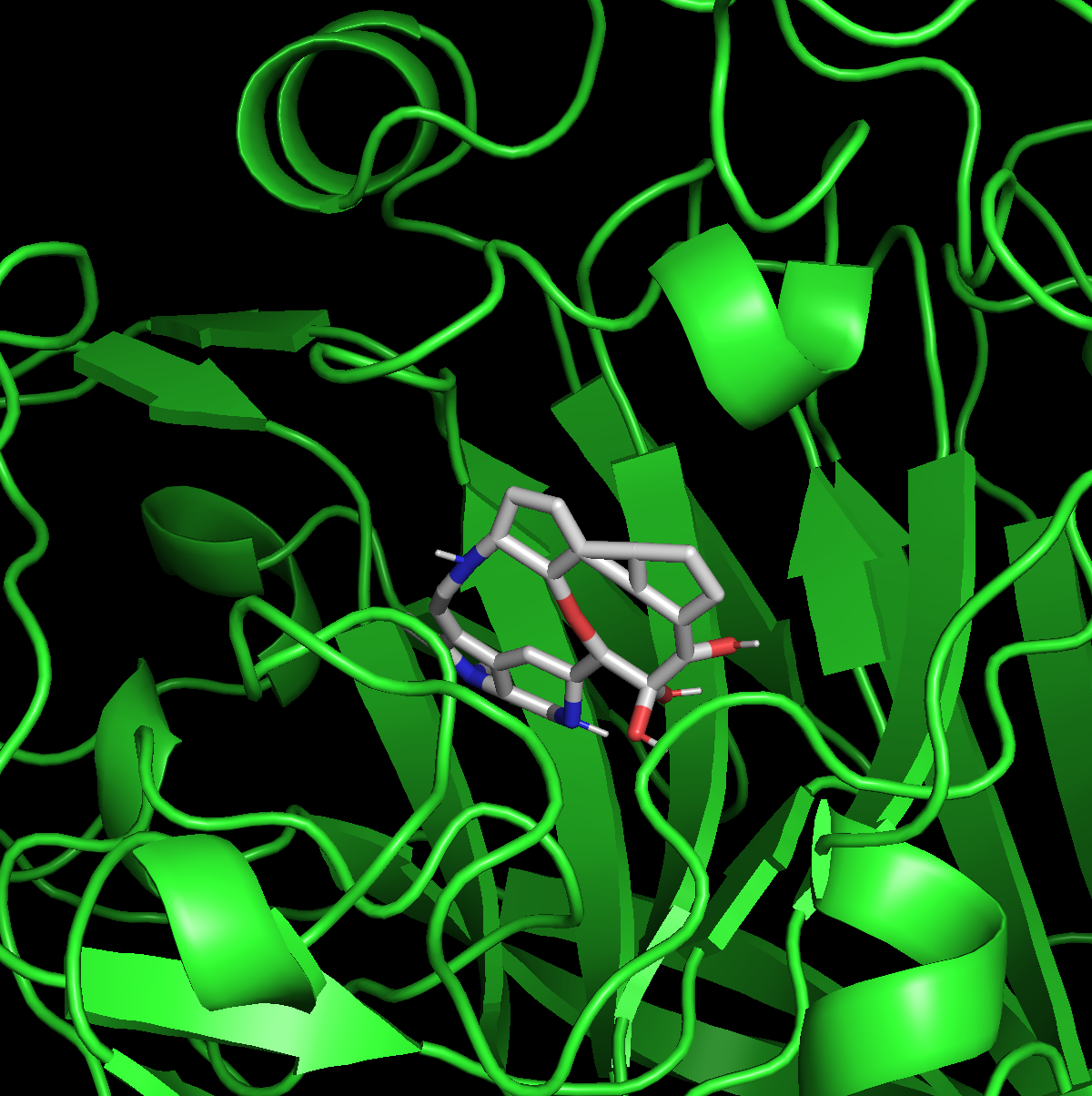}}
    \\
    \subfigure[\scriptsize{BDTG (ours) (Score = -18.908)}]{
    \label{ours4}
    \includegraphics[width=0.32\linewidth]{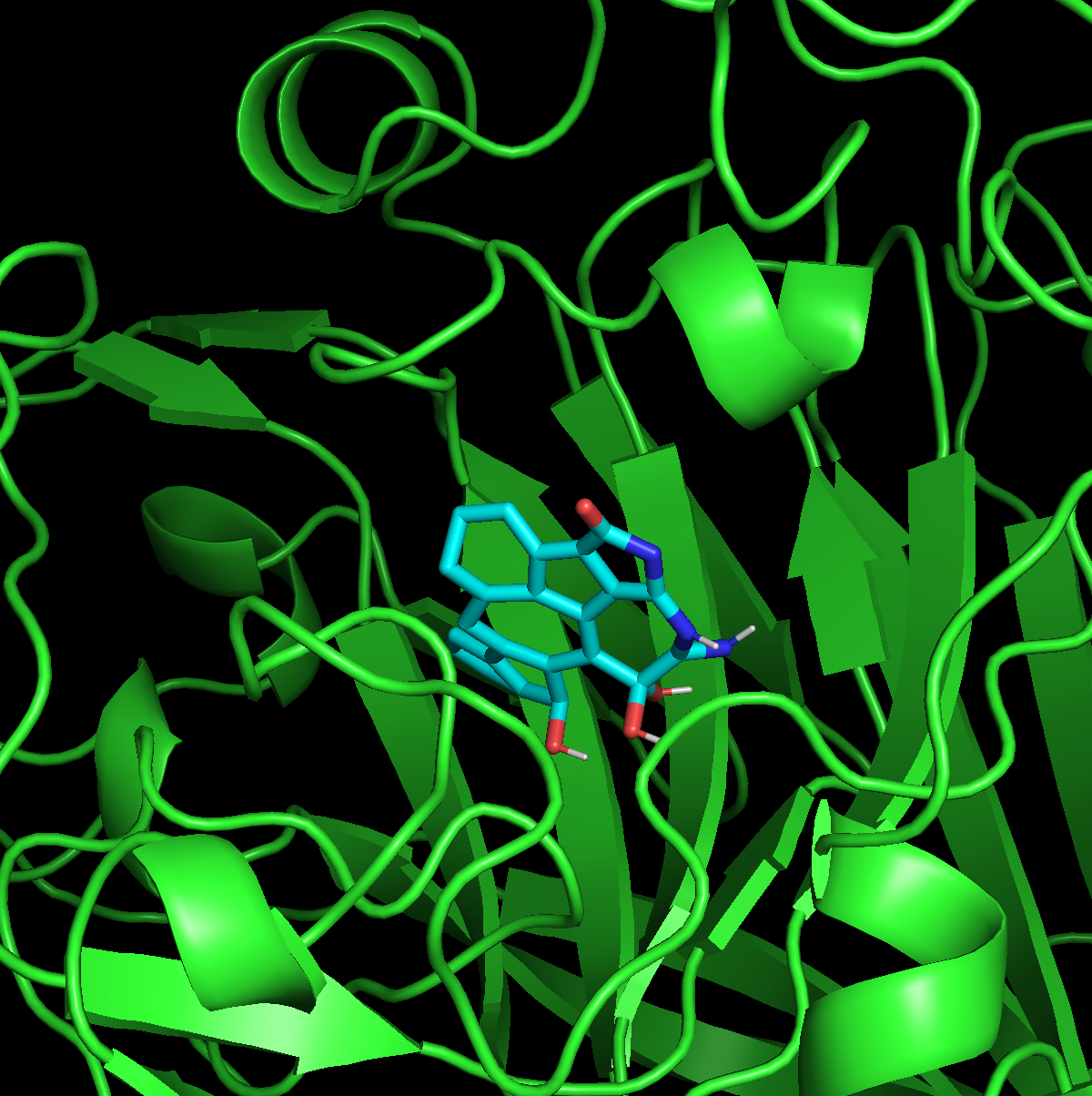}}
    \subfigure[\scriptsize{TuRBO (Score = -16.390)}]{
    \label{bo4}
    \includegraphics[width=0.32\linewidth]{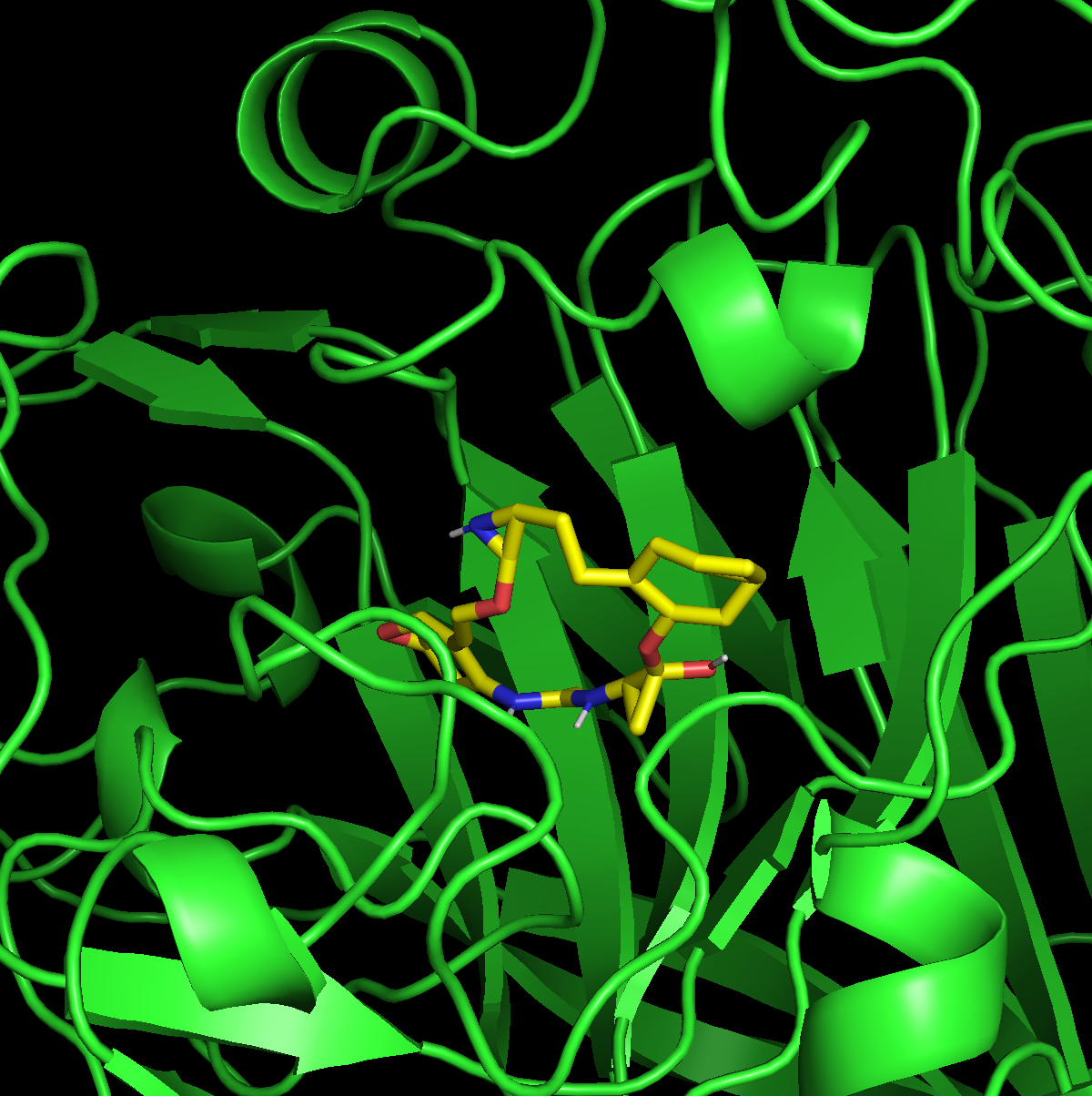}}
    \subfigure[\scriptsize{CMAES (Score = -14.734)}]{
    \label{cma4}
    \includegraphics[width=0.32\linewidth]{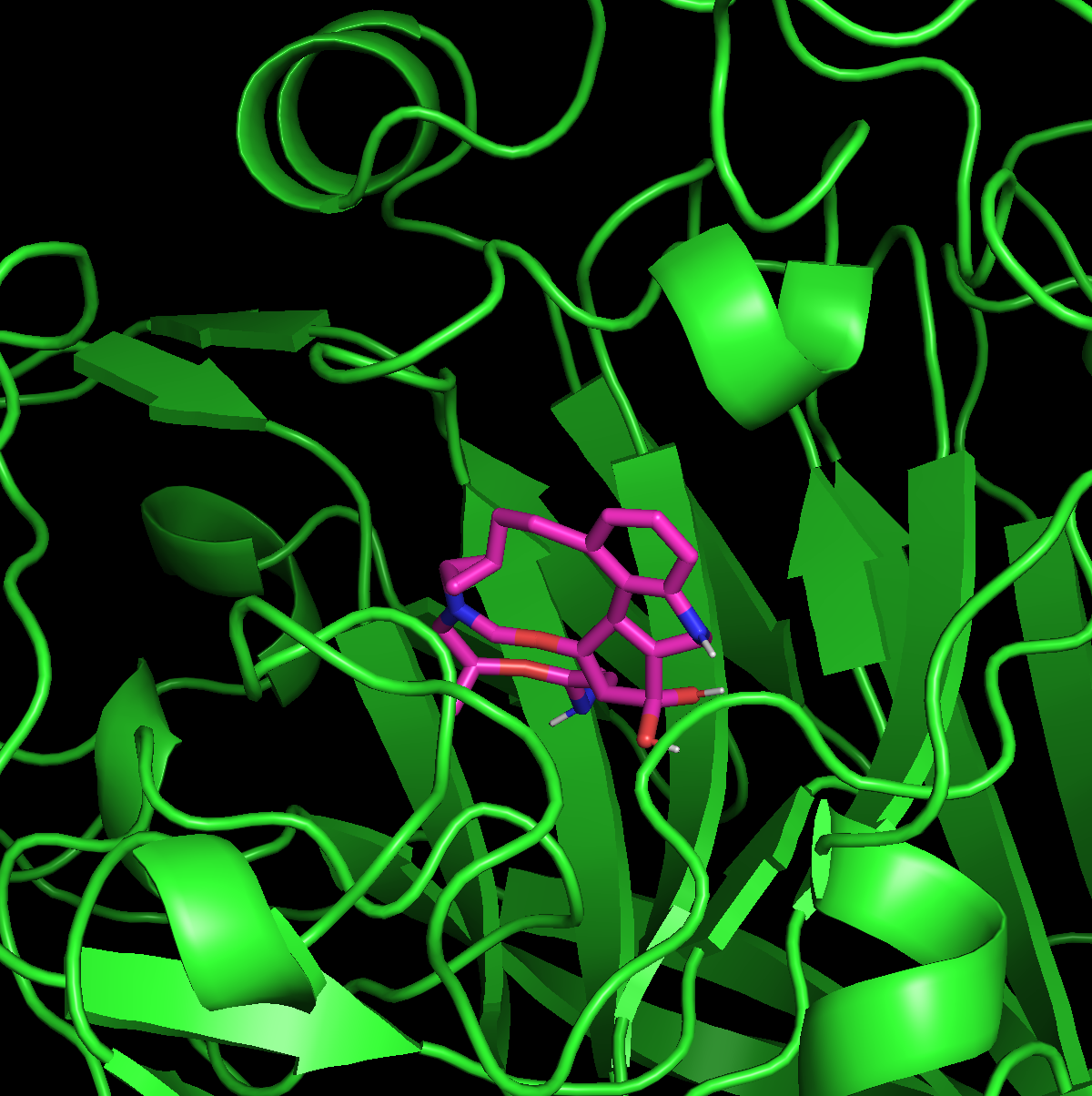}}
    \caption{Demonstration of the Generated 3D-molecule on Receptor-4}
    \label{Demo4}
\end{figure}

\begin{figure}[h]
    \centering
    \subfigure[\scriptsize{Pre-train Best (Score = -6.410)}]{
    \label{pretrainBest5}
    \includegraphics[width=0.32\linewidth]{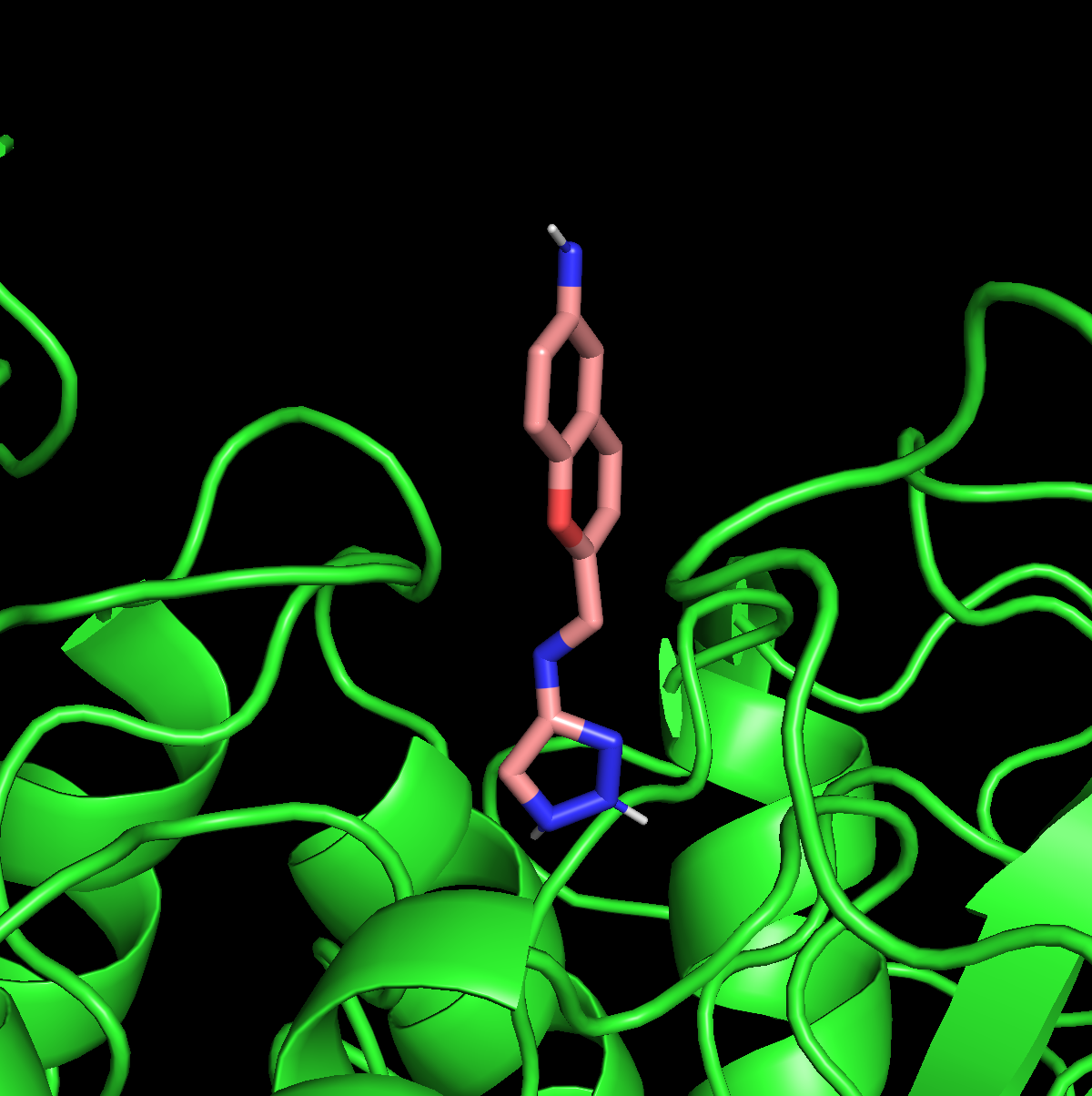}}
    \subfigure[\scriptsize{Before finetune (Score = -5.677)}]{
    \label{noFinetune5}
    \includegraphics[width=0.32\linewidth]{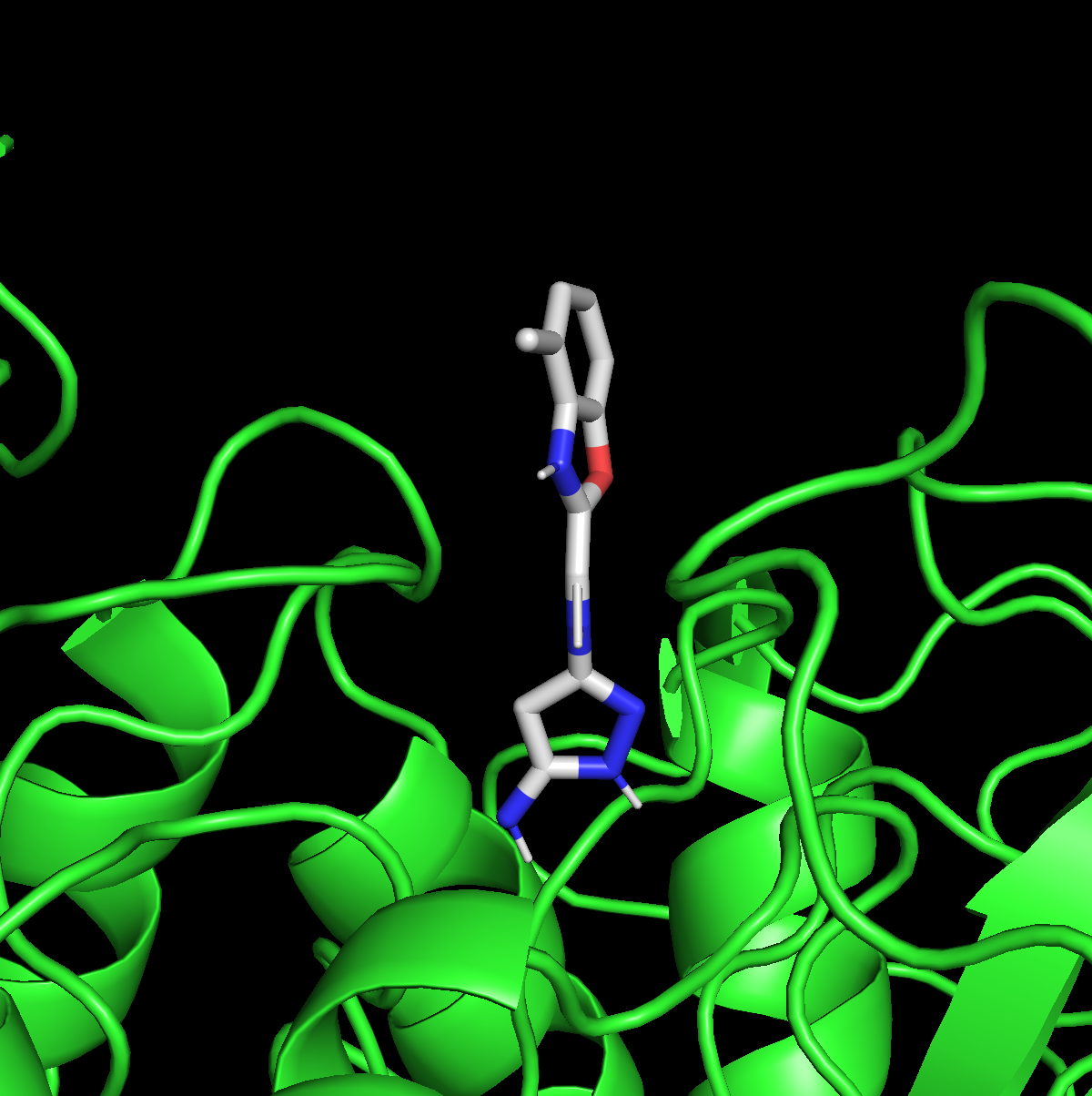}}
    \\
    \subfigure[\scriptsize{BDTG (ours) (Score = -9.589)}]{
    \label{ours5}
    \includegraphics[width=0.32\linewidth]{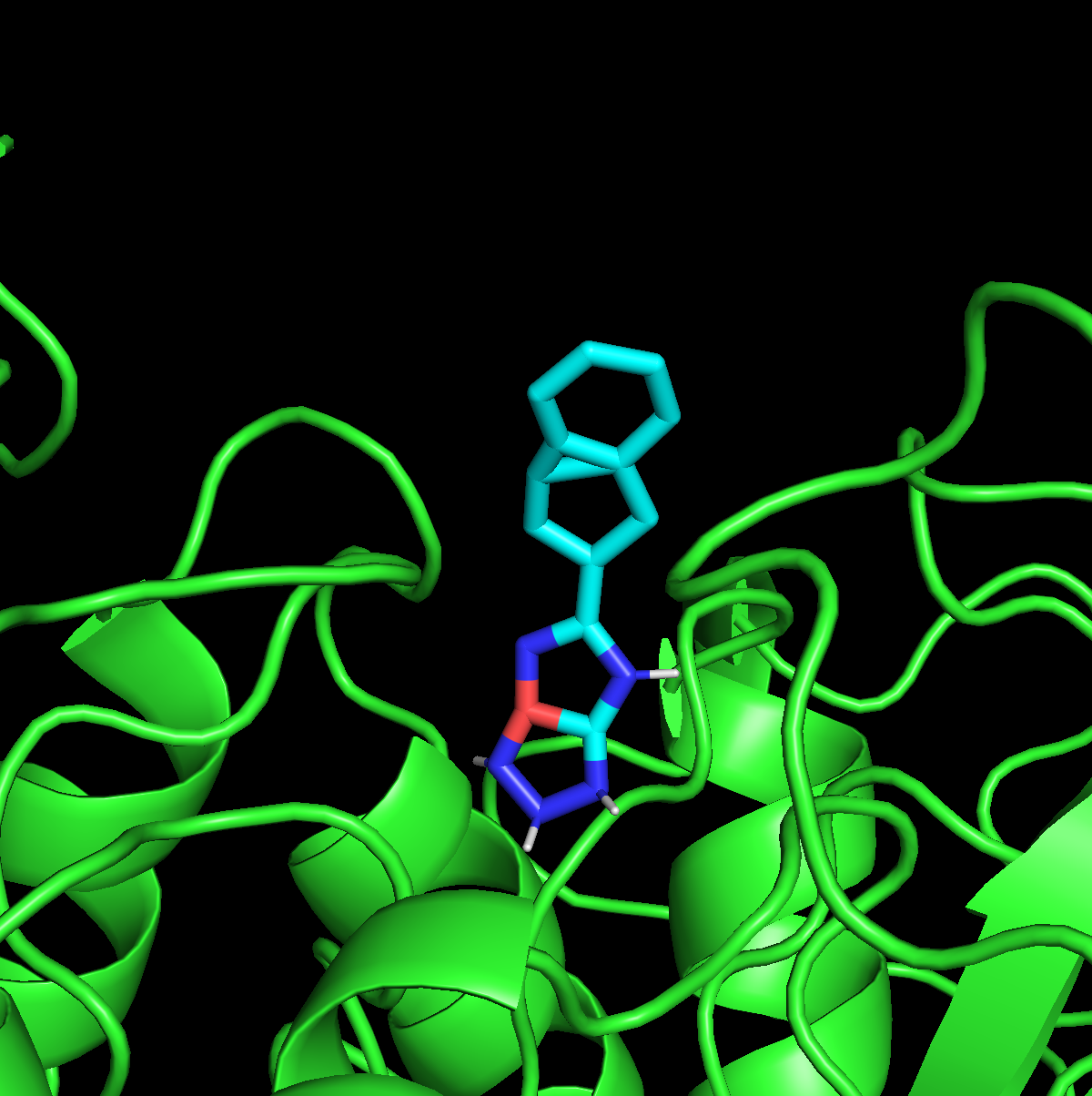}}
    \subfigure[\scriptsize{TuRBO (Score = -8.400)}]{
    \label{bo5}
    \includegraphics[width=0.32\linewidth]{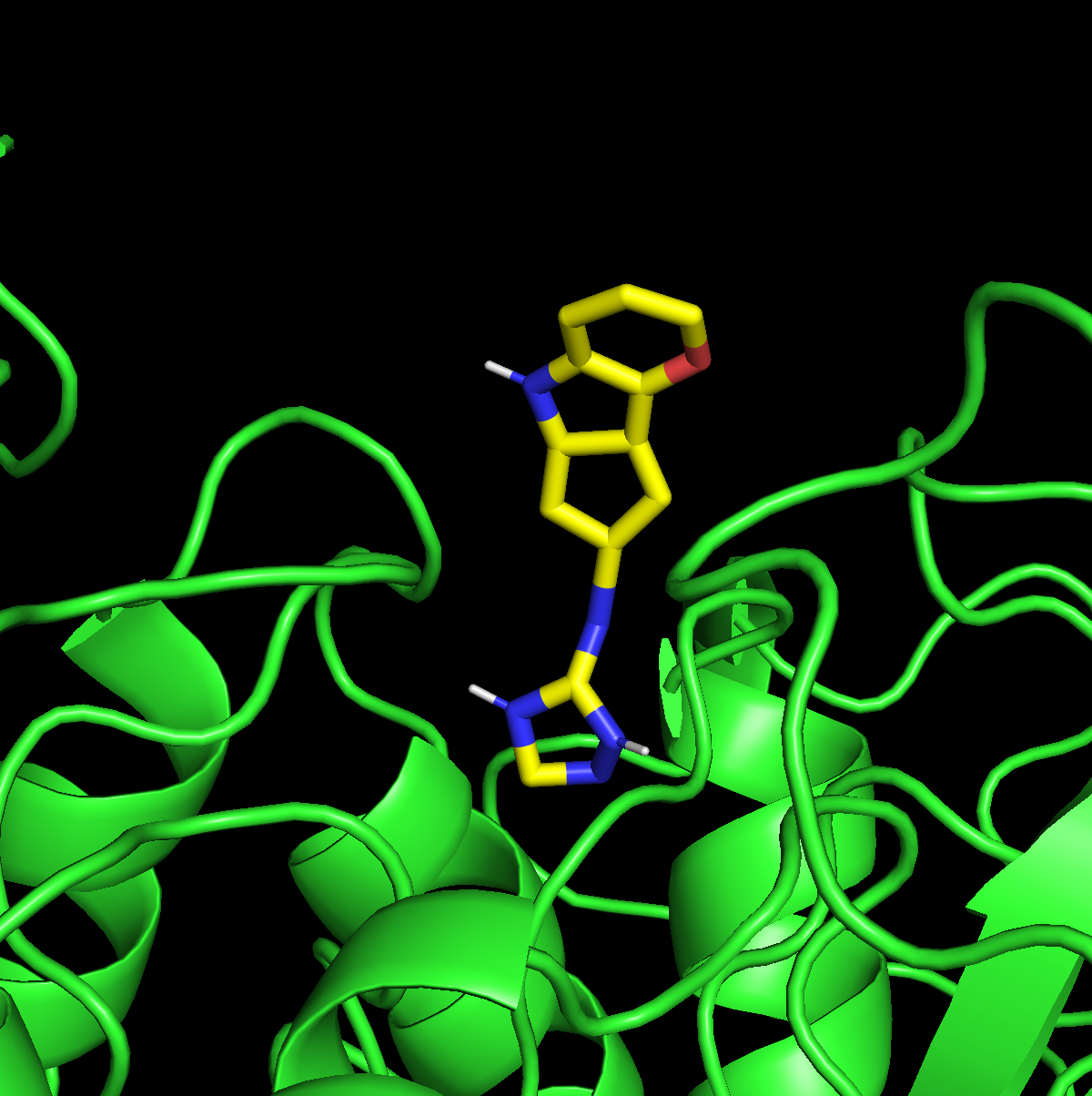}}
    \subfigure[\scriptsize{CMAES (Score = -6.440)}]{
    \label{cma5}
    \includegraphics[width=0.32\linewidth]{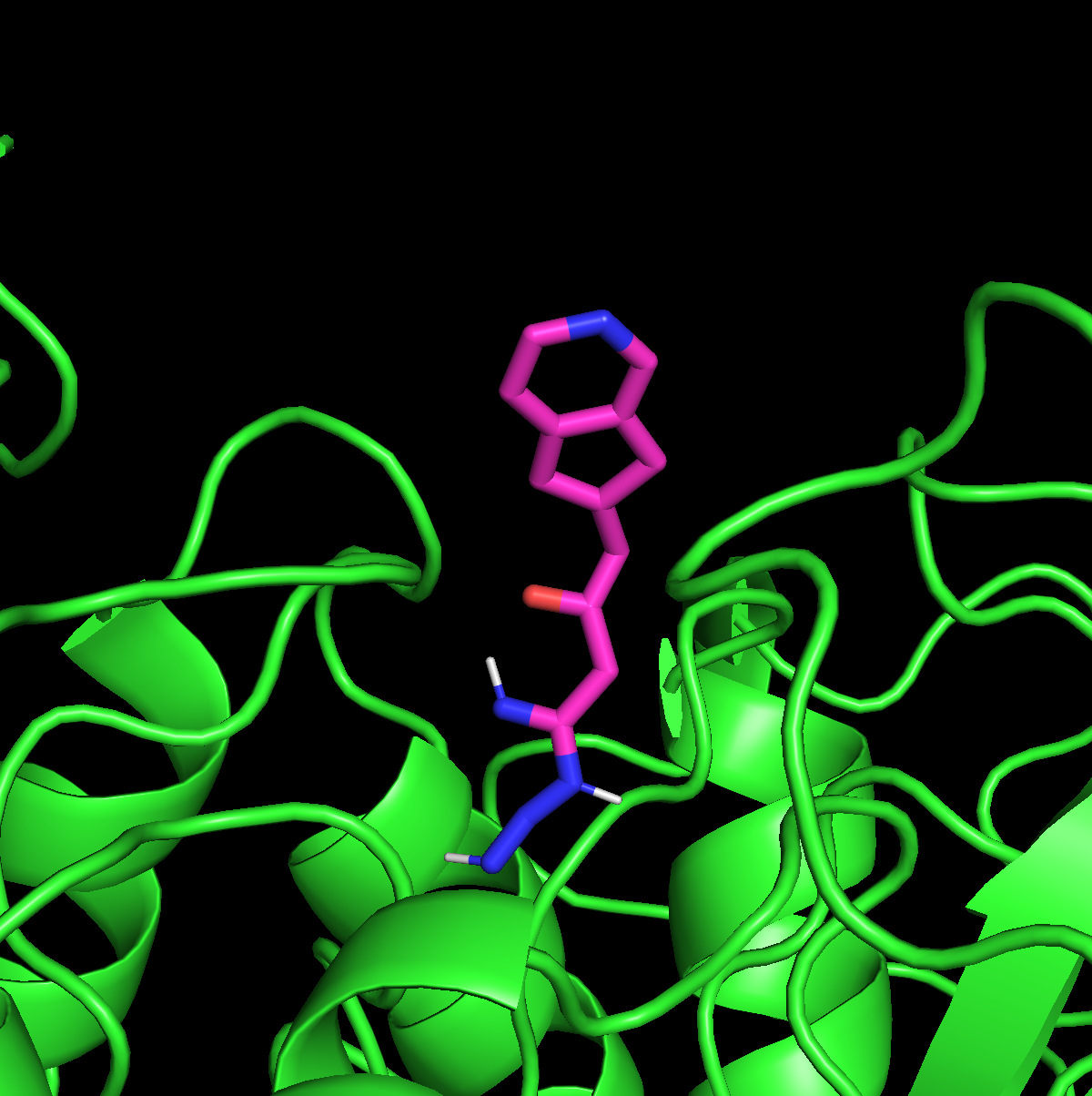}}
    \caption{Demonstration of the Generated 3D-molecule on Receptor-5}
    \label{Demo5}
\end{figure}

\clearpage

\section{Demonstration of Fine-Tuned Stable Diffusion for Targeted Image Generation}
\label{TargetedImage}

\begin{figure}[h]
\centering
\includegraphics[width=1.0\linewidth]{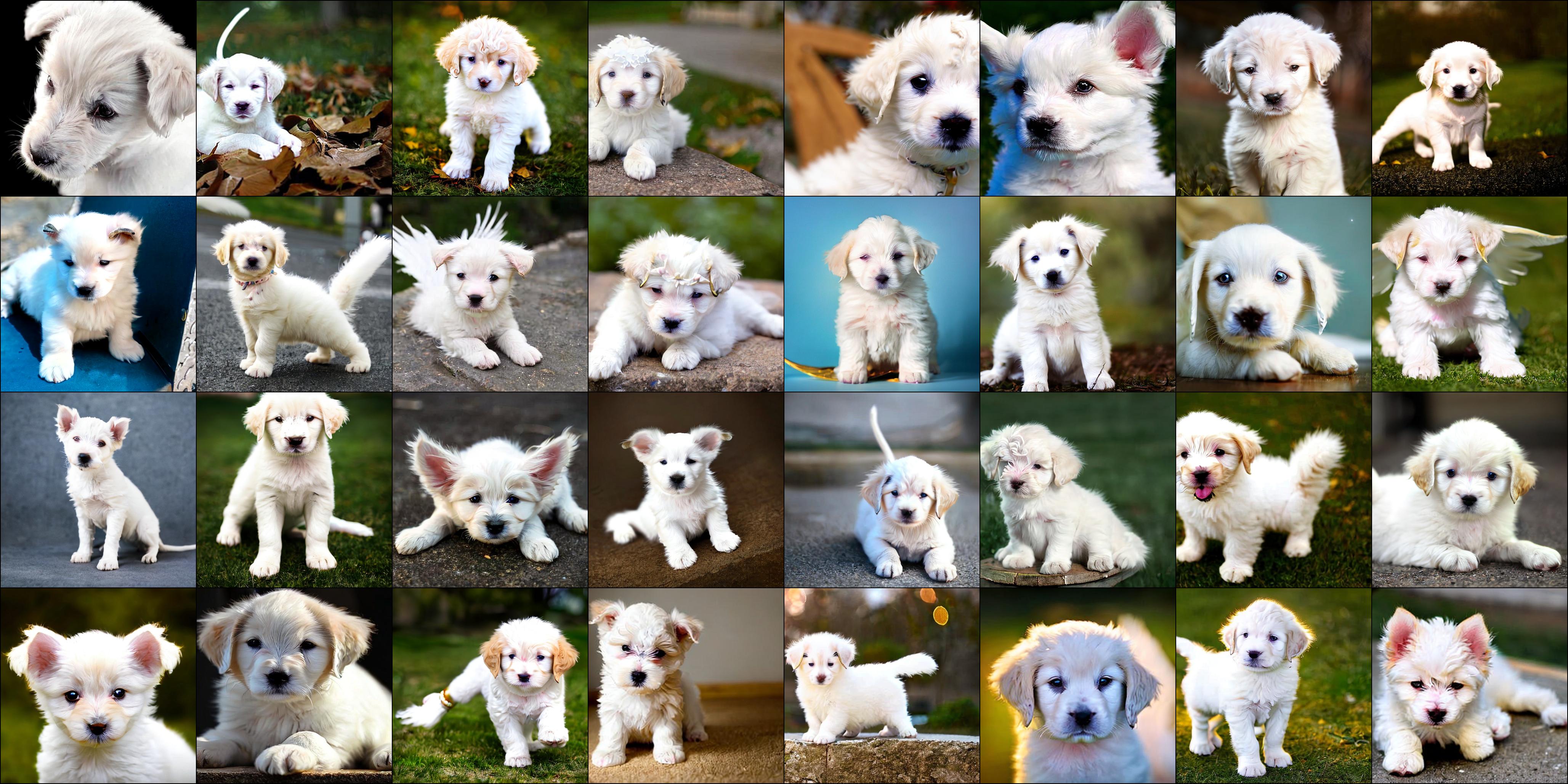}
\caption{Original Pre-trained Stable Diffusion, Average CLIP Score (Higher is Better) = 0.3084}
\label{dog_original}
\end{figure}

\textbf{Pre-trained Model.} We employ the Stable Diffusion v2.1 ~\citep{rombach2021highresolution} (\texttt{stable-diffusion-2-1-base}) as our pre-trained diffusion model for targeted image generation, with DPM-Solver++ as the SDE solver. All parameters are kept as their default settings, including the number of inference steps $K=50$. We randomly generate 32 images using text prompt \texttt{a white puppy with golden wings},  as shown in Figure~\ref{dog_original}. We can observe that the generated images are not semantically accurate; the puppy does not have wings. This demonstrates the limitation of the pre-trained diffusion model in generating unseen data.

\textbf{Objective.} We use OpenCLIP~\citep{ilharco_gabriel_2021_5143773,cherti2023reproducible,Radford2021LearningTV,schuhmann2022laionb} (\texttt{ViT-B-32}) model as our objective function. The CLIP model embeds text and images into a common space, allowing us to measure the cosine distance between the text and the images. In our experiment, we compute the cosine distance between the preset text prompt and the generated image as our \textit{CLIP score}. Our goal is to generate images with high CLIP scores.

To conserve computational resources, we avoid computing the $x_{k \rightarrow K}$ as described in \ref{molecules_generation}. Instead, we evaluate the CLIP score only on the final generated images and use this score for the intermediate sampling steps. This is equivalent as to setting $f(x_i):=f(x_K)$ for $i \in \{ 0,\cdots, K-1 \}$ in our formulation.

\textbf{Fine-Tuning.} The fine-tuning process is identical to our molecules generation experiment in \ref{molecules_generation} with the batch size $N=32$, learning rate $\alpha=50$, and fine-tune for $T=300$ optimization steps.

\begin{figure}[h]
\centering
\includegraphics[width=1.0\linewidth]{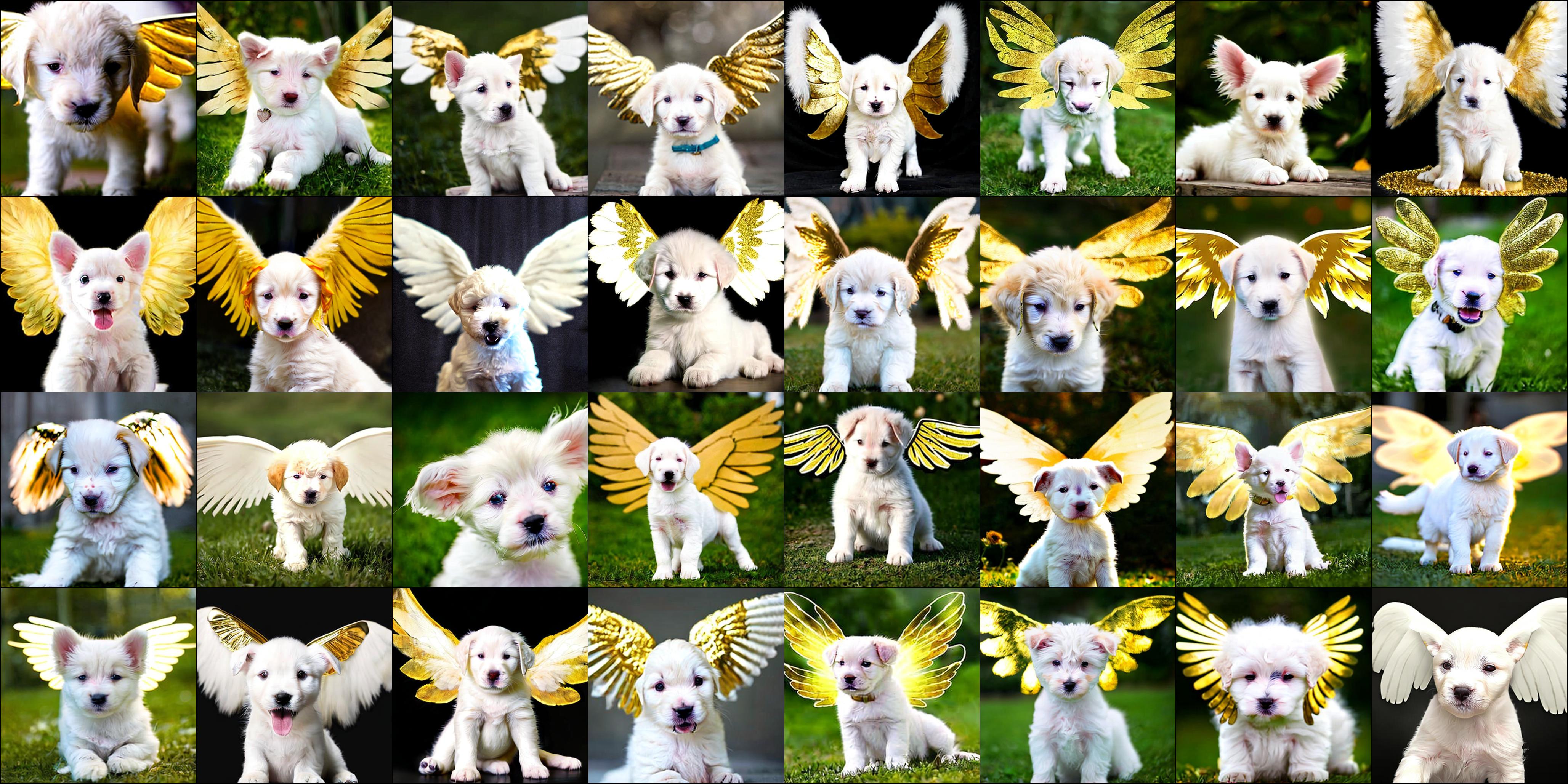}
\caption{Fine-tuned Stable Diffusion (Steps = 100), Average CLIP Score (Higher is Better) = 0.4037}
\label{dog_step=100}
\end{figure}

\begin{figure}[h]
\centering
\includegraphics[width=1.0\linewidth]{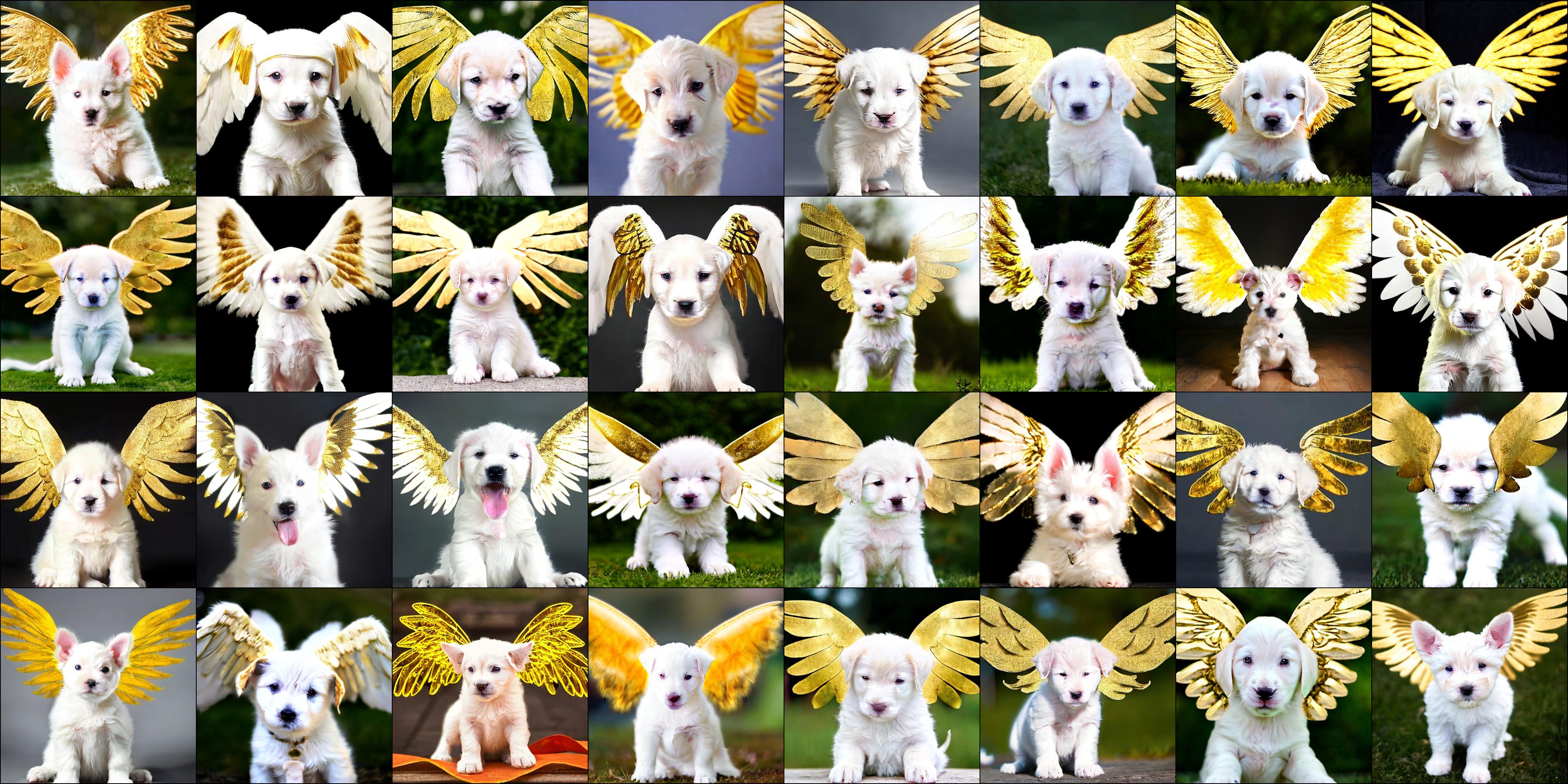}
\caption{Fine-tuned Stable Diffusion (Steps = 200), Average CLIP Score (Higher is Better) = 0.4224}
\label{dog_step=200}
\end{figure}

\begin{figure}[h]
\centering
\includegraphics[width=1.0\linewidth]{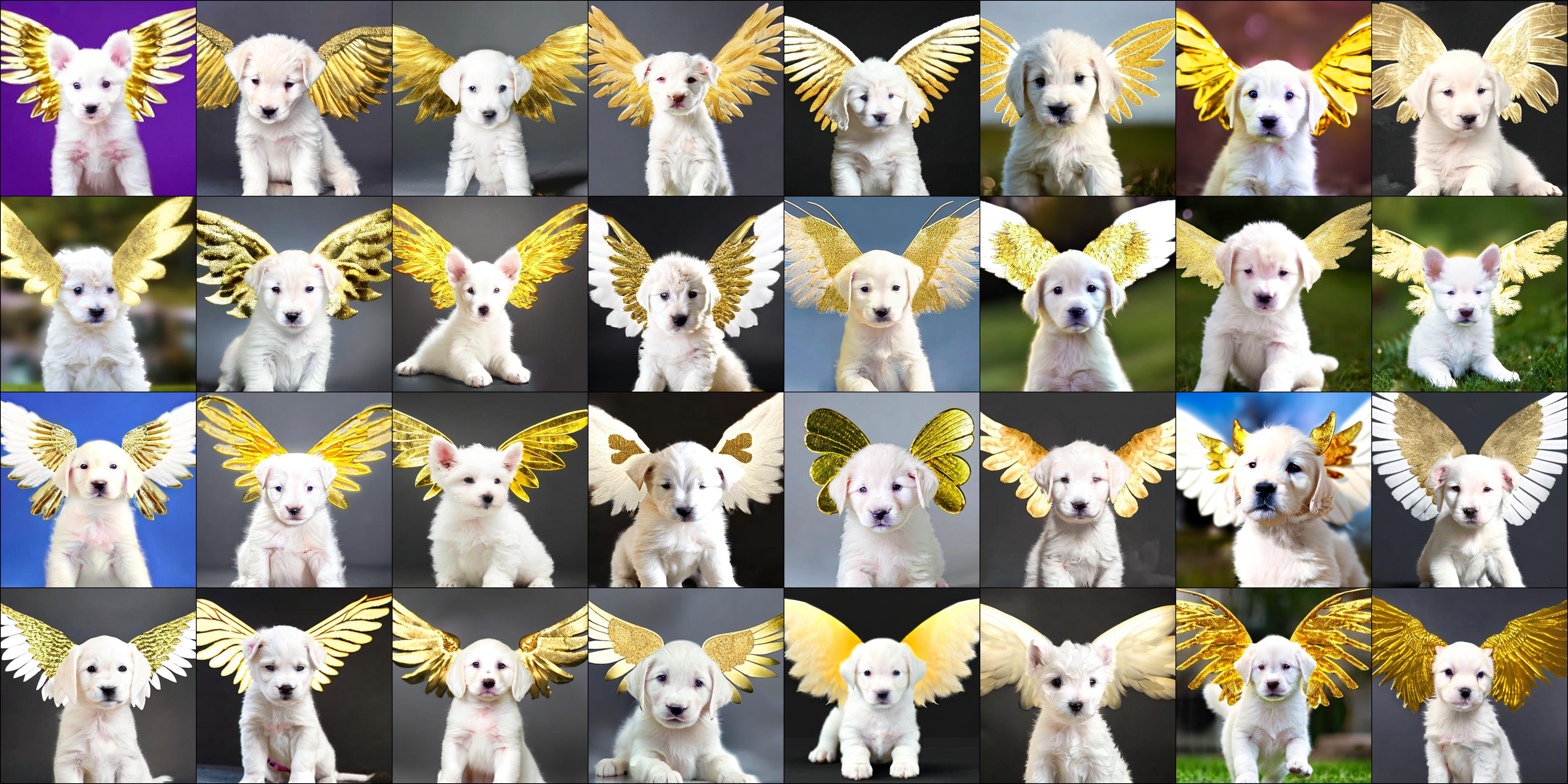}
\caption{Fine-tuned Stable Diffusion (Steps = 300), Average CLIP Score (Higher is Better) = 0.4274}
\label{dog_step=300}
\end{figure}

\textbf{Results.} We randomly sample 32 images at the fine-tuning steps $T=\{ 100,200,300\}$ and show the generated images in Figure \ref{dog_step=100}, \ref{dog_step=200}, \ref{dog_step=300}, respectively. We observe that the fine-tuned model begins to generate images with accurate visual appearances at the 100-th step (with most dogs having golden wings). By 300 steps, the images consistently show accurate visual appearances (with every dog having golden wings).

This experiment demonstrates that our method is generalizable and can be applied to various domains.

\end{document}

%% file: math_commands.tex
\usepackage{amsmath,amsfonts,bm}
\usepackage{makecell}

\def\eqref#1{equation~\ref{#1}}

\def\1{\bm{1}}

\DeclareMathAlphabet{\mathsfit}{\encodingdefault}{\sfdefault}{m}{sl}
\SetMathAlphabet{\mathsfit}{bold}{\encodingdefault}{\sfdefault}{bx}{n}

